\documentclass[sigconf]{acmart}
\preto{\abstractkeywords}{\nolinenumbers} 

\AtBeginDocument{%
  \providecommand\BibTeX{{%
    \normalfont B\kern-0.5em{\scshape i\kern-0.25em b}\kern-0.8em\TeX}}}
\usepackage{adjustbox}
\usepackage{tabularx} %
\usepackage{multirow}
\usepackage{float}
\usepackage{colortbl}
\usepackage{subfigure}
\usepackage{arydshln}
\numberwithin{table}{section}
\numberwithin{figure}{section}

\definecolor{mygreen}{RGB}{0, 128, 0}
\definecolor{cell1}{RGB}{230, 183, 69}

\definecolor{cell2}{RGB}{132, 186, 66}

\definecolor{cell3}{RGB}{168, 203, 223}

\copyrightyear{2025}
\acmYear{2025}

\setcopyright{acmlicensed}
\acmConference[MM '25] {Proceedings of the 33rd ACM International Conference on Multimedia}{October 27--31, 2025}{Dublin, Ireland.}
\acmBooktitle{Proceedings of the 33rd ACM International Conference on Multimedia (MM '25), October 27--31, 2025, Dublin, Ireland}
\acmISBN{979-8-4007-2035-2/2025/10}
\setcopyright{rightsretained}

\acmBooktitle{Proceedings of the 33rd ACM International Conference on Multimedia (MM '25), October 27--31, 2025, Dublin, Ireland}
\acmDOI{10.1145/3746027.3754871}

\begin{document}
\title{DepthDark: Robust Monocular Depth Estimation for Low-Light Environments}


\author{Longjian Zeng}
\email{245060073@hdu.edu.cn}
\affiliation{%
  \institution{Hangzhou Dianzi University}
  \city{Hangzhou}
  \country{China}}
  
\author{Zunjie Zhu}
\authornote{Corresponding Author.}
\email{zunjiezhu@hdu.edu.cn}
\affiliation{%
    \institution{Hangzhou Dianzi University}
  \city{Hangzhou}
  \country{China}}

\author{Rongfeng Lu}
\email{rongfeng-lu@hdu.edu.cn}
\affiliation{%
    \institution{Hangzhou Dianzi University}
  \city{Hangzhou}
  \country{China}}

\author{Ming Lu}
\email{lu199192@gmail.com}
\affiliation{%
  \institution{Intel Labs China}
  \city{Beijing}
  \country{China}}

\author{Bolun Zheng}
\email{blzheng@hdu.edu.cn}
\affiliation{%
    \institution{Hangzhou Dianzi University}
  \city{Hangzhou}
  \country{China}}

\author{Chenggang Yan}
\email{cgyan@hdu.edu.cn}
\affiliation{%
    \institution{Hangzhou Dianzi University}
  \city{Hangzhou}
  \country{China}}

\author{Anke Xue}
\email{akxue@hdu.edu.cn}
\affiliation{%
    \institution{Hangzhou Dianzi University}
  \city{Hangzhou}
  \country{China}}

\renewcommand{\shortauthors}{Longjian Zeng et al.}

\begin{abstract}
In recent years, foundation models for monocular depth estimation have received increasing attention. Current methods mainly address typical daylight conditions, but their effectiveness notably decreases in low-light environments. There is a lack of robust foundational models for monocular depth estimation specifically designed for low-light scenarios. This largely stems from the absence of large-scale, high-quality paired depth datasets for low-light conditions and the effective parameter-efficient fine-tuning (PEFT) strategy. To address these challenges, we propose DepthDark, a robust foundation model for low-light monocular depth estimation. We first introduce a flare-simulation module and a noise-simulation module to accurately simulate the imaging process under nighttime conditions, producing high-quality paired depth datasets for low-light conditions. Additionally, we present an effective low-light PEFT strategy that utilizes illumination guidance and multiscale feature fusion to enhance the model's capability in low-light environments. Our method achieves state-of-the-art depth estimation performance on the challenging nuScenes-Night and RobotCar-Night datasets, validating its effectiveness using limited training data and computing resources.
\end{abstract}


  
\maketitle

\section{Introduction}
In recent years, with the rapid advancement of convolutional neural networks \cite{Deep_learning} and vision transformers \cite{vit-transformers}, the fields of computer vision and natural language processing are undergoing a significant revolution. This revolution has also significantly boosted the development of monocular depth estimation, which predicts depth information from a single image, enabling applications in areas such as autonomous driving, augmented reality, and robotics. However, most existing methods perform well under sufficient illumination but degrade in low-light conditions because of substantial information loss and significant noise amplification.

Although low-light monocular depth estimation is a highly challenging task, it plays a pivotal role in various nighttime applications. However,
most existing studies on low-light monocular depth estimation, such as RNW \cite{RNW}, ADDS \cite{adds}, MonoVit \cite{Monovit}, MonoFormer \cite{MonoFormer}, and TDDC \cite{TDDC}, focus primarily on nighttime autonomous driving, which limits their applicability to other applications. Consequently, developing a foundation model capable of estimating depth information from a single nighttime image has become a critical goal in low-light monocular depth estimation. Although foundation models such as Depth Anything \cite{depthanything}, Depth Anything V2 \cite{depthanything_v2} and Marigold \cite{Marigold} demonstrate strong performance under typical daylight conditions, their effectiveness in complex low-light scenarios remains constrained due to limitations in training data coverage. The success of these models typically hinges on large-scale, high-quality training datasets. However, constructing a dataset that contains tens of millions of depth labels in low-light conditions is an exceedingly challenging task. Additionally, training a foundational model requires significant computational resources, which intensifies the challenges of research and practical application. Therefore, developing a foundation model based on parameter-efficient fine-tuning (PEFT) for low-light monocular depth estimation is urgent and essential.

In this study, we comprehensively address the challenges in low-light monocular depth estimation by introducing an innovative approach, DepthDark, to overcome the above challenges. Specifically, we propose a flare-simulation module and a noise-simulation module to synthesize highly realistic nighttime images from daytime images. The flare-simulation module technique effectively resolves photometric disparities caused by uneven light source distributions, while the noise-simulation module employs a physically decoupled framework to accurately simulate the significant noise distribution variations in low-light scenes. As a result, these innovations effectively overcome the difficulties associated with collecting large-scale, high-quality paired nighttime depth data.

We also propose an efficient parameter-efficient fine-tuning (PEFT) strategy to adapt pre-trained foundation models for low-light depth estimation. In this strategy, we introduce illumination guidance and multiscale feature fusion, which significantly enhance the model's performance in low-light environments. Using illumination guidance, the model can better adapt to varying light conditions, allowing it to focus on relevant features that may be obscured in low-light scenarios. Additionally, multiscale feature fusion enables the model to integrate information across different scales, improving depth perception and accuracy. This combination allows DepthDark to effectively tackle the challenges posed by low-light conditions, resulting in robust depth estimation performance. Moreover, our low-light PEFT strategy requires only a few hours of training on a single consumer-grade GPU and utilizes a minimal amount of synthetic RGB-D training data to achieve efficient adaptation of the foundation model. This efficiency significantly reduces the barrier to entry for researchers and developers working in low-light monocular depth estimation, making it more accessible. By leveraging these advancements, DepthDark not only performs well in challenging scenarios but also democratizes the capabilities of depth estimation in low-light environments, encouraging further exploration and innovation in this area. The main contributions of our work are summarized as follows:

\begin{itemize}
  \item We propose a novel approach to synthesizing low-light images by realistically simulating light sources and noise characteristics. This method addresses the challenge of collecting large-scale paired depth data for low-light scenarios.
  \item We design an efficient parameter-efficient fine-tuning (PEFT) strategy that significantly enhances the model's robustness in low-light environments by incorporating illumination guidance and multiscale feature fusion.
  \item DepthDark achieves state-of-the-art performance on the nuScenes-Night and RobotCar-Night datasets with a single consumer-grade GPU, short training times, and minimal training data.
\end{itemize}

\begin{figure*}[htbp]
    \hsize=\textwidth
    \centering
    \includegraphics[width=0.7\linewidth]{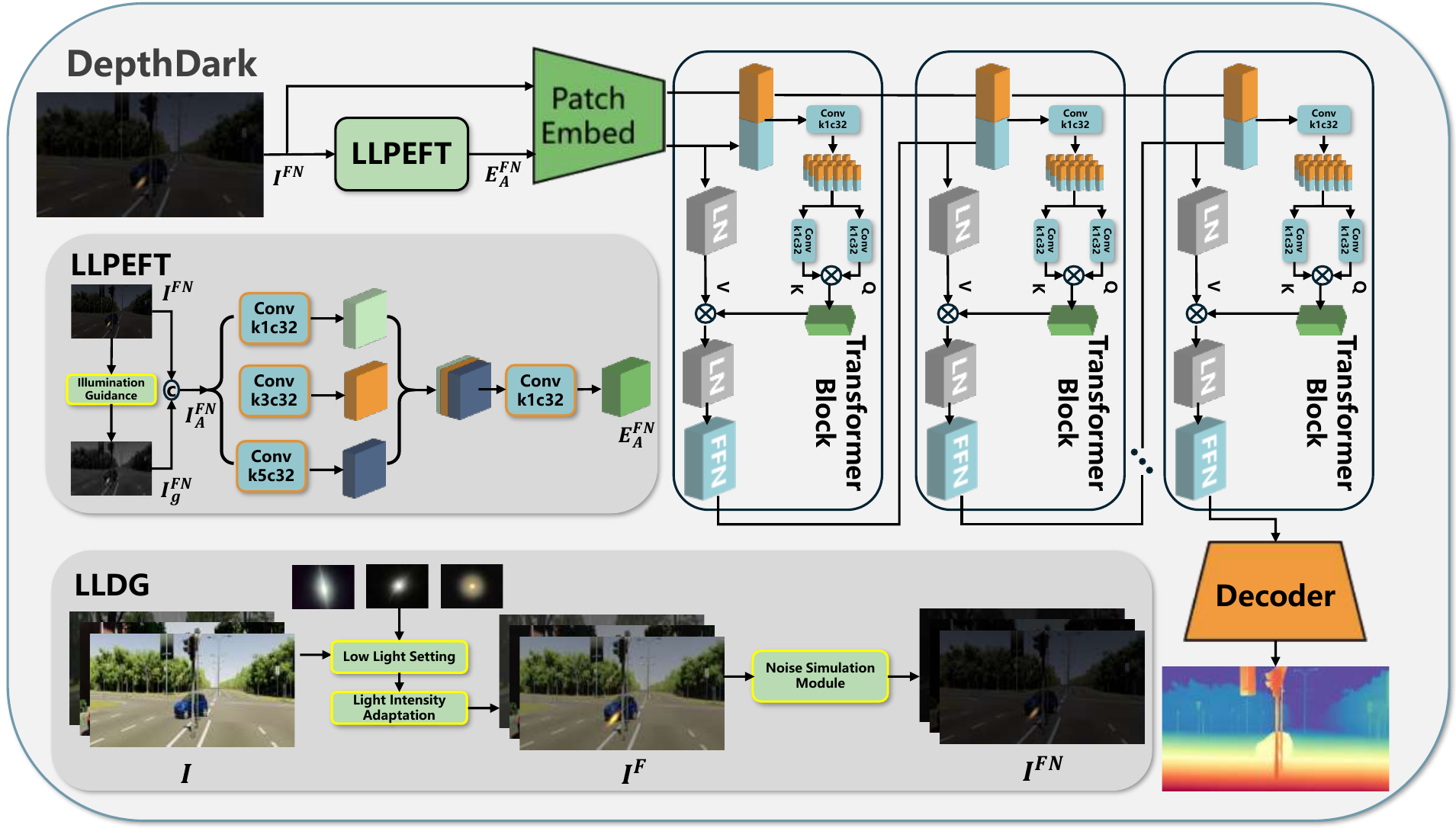}
    \caption{
    {\textbf{Overview of our DepthDark Training Framework}.} The framework consists of LLDG and LLPEFT. The Patch Embed module is inspired by vision transformers\protect\cite{vit-transformers} for efficient feature extraction.
    }
    \label{Fig: framework}
    \vspace{-2mm}
    \Description{A block diagram showing the architecture of the proposed neural network method.}
\end{figure*}

\section{Related Work}

Early studies on monocular depth estimation \cite{hoiem2007recovering,liu2008sift,huang2022neighbor2neighbor} and other methods \cite{huang2021neighbor2neighbor, ACM-5, ACM-6, ACM-7, ACM-4, lu_1, lu_3} relied mainly on hand-made features and traditional computer vision techniques, which struggled to handle complex scenes with occlusions and textureless regions. However, The advent of deep learning revolutionized monocular depth estimation, with Eigen \cite{eigen2014depth} pioneering multiscale networks and demonstrating that depth results could be converted into metric depths for datasets recorded with a single sensor. Subsequent research has focused on improving accuracy by incorporating additional priors \cite{yang2023gedepth,shao2023nddepth,li2015depth, buff} and optimizing objective functions \cite{xian2020structure,yin2019enforcing}.

\subsection{Low Light Monocular depth estimation}

Although these methods have provided satisfactory depth estimation results, they often fail in nighttime scenarios due to significant differences between day and night distributions. To address the challenges of low-light depth estimation, various methods employing domain adaptation or photometric loss have been proposed. For example, ADFA \cite{ADFA} introduces a new encoder that simulates a daytime depth estimation model, enabling the generation of daytime features from nighttime images through domain adaptation. ITDFA \cite{ITDFA} employs migrated image pairs to constrain the training process in both feature and output spaces. ADDS \cite{adds} utilizes independent encoders to decompose day and night images into invariant and variant feature maps, estimating depth using invariant information. RNW \cite{RNW} regularizes nighttime photometric outliers through a prior-based output space domain adaptation approach. However, the performance of the aforementioned methods is constrained by the challenges posed by the brightness and illumination inconsistencies in low-light images. WSGD \cite{WSGD} is the first one-stage method that directly trains its proposed framework for nighttime image segmentation. 
TDDC \cite{TDDC} presents a self-supervised low-light monocular depth estimation method by applying physical priors to compensate for key day-night discrepancies. 

Although existing low-light monocular depth estimation methods show potential in nighttime autonomous driving scenarios, their performance is often limited to specific low-light environments, making it challenging to generalize to other nighttime applications. Therefore, training a foundational model for low-light monocular depth estimation becomes crucial. Such a model can improve the generalization ability of depth estimation across diverse low-light scenarios, providing more stable and accurate depth information for various practical applications.

\subsection{Foundation Models}
Vision Foundation Models (VFMs) are large-scale neural networks trained on extensive datasets. The remarkable scaling of VFMs has significantly advanced visual understanding, allowing efficient fine-tuning across a broad spectrum of downstream tasks with minimal effort. Prompt-tuning techniques \cite{zhang2023prompt,yao2023visual,bahng2022exploring,jie_2, lu_2} have shown that by designing appropriate prompts, VFMs can be effectively tailored to specific scenarios. Feature adaptation approaches \cite{hu2021lora,zhao2023unleashing,bahng2022exploring,du2023generative,gao2024clip,jie_1} further enhance the applicability of VFMs across different tasks. VPD \cite{zhao2023unleashing} has demonstrated the potential of extracting features from pre-trained text-to-image models for domain-specific depth estimation. Meanwhile, I-LoRA \cite{du2023generative} demonstrated the multimodal capabilities of pre-trained image generators. Depth Anything \cite{depthanything} pioneered the use of teacher model training to construct large-scale datasets, establishing itself as the first foundation model for monocular depth estimation. Marigold \cite{Marigold} employed Stable Diffusion \cite{diffusion_models} to generate synthetic images with complete depth information, further enhancing the quality of synthetic image labels. Depth Anything V2 \cite{depthanything_v2} replaces labeled real-world images with synthetic images and enhances the capacity of the teacher model, leading to the development of an advanced depth estimation model. However, Depth Anything V2 performs poorly under low-light conditions, indicating the model's limited generalization ability in challenging low-light environments.

To address the aforementioned challenges, we propose an efficient foundational model for low-light depth estimation, DepthDark. This approach constructs a high-quality synthetic training set comprising 74k pairs of daytime and nighttime depth data by incorporating both a flare-simulation module and a noise-simulation module.  Additionally, we introduce an efficient parameter-efficient fine-tuning strategy.

\section{Method}

Monocular depth estimation foundation models, such as Depth Anything \cite{depthanything} and Depth Anything V2 \cite{depthanything_v2}, utilize a data engine to collect and automatically annotate large-scale unlabeled data. However, these methods lack effective modeling of nighttime scenes, limiting their performance in downstream tasks under low-light conditions. To overcome this limitation, we aim to develop a robust foundation model specifically designed for low-light monocular depth estimation. For the first time, we introduce Low-Light Dataset Generation (LLDG), which integrates a flare-simulation module and a noise-simulation module techniques to precisely simulate the imaging process under low-light conditions, thereby producing a high-quality paired depth dataset for low-light scenes, as detailed in Sec.  \ref{sec:method_noise}. Additionally, to further enhance robustness and generalization in low-light scenarios, we propose an efficient Low-Light Parameter-Efficient Fine-Tuning (LLPEFT) strategy tailored for low-light scenarios. This strategy leverages Illumination Guidance and Multiscale Feature Fusion to significantly enhance the model's adaptability and robustness under low-light conditions, as detailed in Sec.  \ref{sec:method_PEFT}. The overall framework of DepthDark is illustrated in Fig. \ref{Fig: framework}.

\subsection{Low Light Dataset Generation}

\label{sec:method_noise}

To address the significant discrepancies between daytime and nighttime image distributions, we propose the Low-Light Dataset Generation (LLDG). This framework synthesizes realistic nighttime image distributions from daytime images by incorporating two key components: Flare Simulation Module (FSM) and Noise Simulation Module (NSM). These components utilize physical priors to simulate the unique noise and photometric characteristics of nighttime scenes, enabling the creation of high-quality paired training datasets for low-light scenes. 

\subsubsection{Flare Simulation Module}
The uneven distribution of light sources and significant optical aberrations in nighttime scenes pose challenges to depth estimation models. To address this, we introduce the Flare Simulation Module (FSM) as the core component of the Low-Light Dataset Generation, which accurately simulates the photometric inconsistencies caused by flares, glare, and brightness peaks in nighttime images, generating a more realistic synthetic dataset. This helps bridge the gap left by existing models that neglect optical defects, enhancing the robustness and generalization of depth estimation models in low-light environments.

\paragraph{Low Light Setting:} In the field of computer graphics, some methods approximate this optical phenomenon using 2D Fourier transforms \cite{lipson2010optical}. However, these methods often produce uncontrollable outputs and are computationally expensive. Therefore, instead of directly simulating light source images, we construct a light source library based on the Flare7K dataset \cite{dai2022flare7k}. Flare7K is currently the only large-scale dataset containing 7,000 light source samples. A light source can be randomly sampled from this dataset and resized to match the dimensions of the input image, ensuring compatibility. Additionally, we apply simple image augmentation operations, such as resizing and cropping, to diversify the light source patterns, with the enhanced light source denoted as $L_S$.

The sampled light sources are randomly positioned in the 3D scene to simulate realistic depth relationships. A depth constraint is applied to limit the effective range of the light source, ensuring that distant light sources do not produce excessive flare artifacts. The 3D position of the light source is given by:
\begin{equation}
P_i = z_i \cdot K_I^{-1} \begin{bmatrix} u_i \\ v_i \\ 1 \end{bmatrix},
\end{equation}
where $z_i$ represents the depth of the light source, and its upper limit is 20m, $K_I$ is the intrinsic camera matrix, and $(u_i, v_i)$ are the 2D coordinates of the light source.

\paragraph{Light Intensity Adaptation:} Since daytime images typically exhibit sufficient illumination and high overall brightness, directly applying light source imaging $L_S$ tends to fail in generating prominent brightness peaks. 

To address this issue, we apply a simple random darkening operation to the input image, where the brightness scaling factor $s_b$ is sampled from a uniform distribution. To enrich the diversity of light sources while avoiding excessively glaring effects, the peak brightness intensity $F$ is defined as the product of the number of light sources $N_F$ and the scaling factor $s_F$, both $s_F$ and $F$ are sampled from a log-uniform distribution:
\begin{equation}
\begin{split}
    s_F &\sim \exp\left(U(\log s_F^{min}, \log s_F^{max})\right), \\
    F &\sim \exp\left(U(\log F^{min}, \log F^{max})\right).
\end{split}
\end{equation}
where $s_F$ is the scale factor and $F^{min}, F^{max}$ represents the intensity range. The total number of light sources $N_F$ is computed as:
    \begin{equation}
    N_F = \max\left(\left\lfloor \frac{F}{s_F} + 0.5 \right\rfloor, 1\right).
    \end{equation}
Additionally, the brightness scale $s_b$ is sampled as $s_b \sim U(0.4, 1)$, and the gamma correction factor is sampled as $g_F \sim U(1.8, 2.2)$. Using the Phong illumination model \cite{phong1998illumination,TDDC}, the final flare image at the current stage, $I^{F}$, is calculated by the following equation:
    \begin{equation}
    I^{F} = \left(s_b \cdot I\right)^{g_F} + \sum_{i=1}^{N_F} \left(ss(L_S, s_F, P_i)\right)^{g_F},
    \end{equation} 

    
where $I$ is the daytime image and $ss(.)$ represents scaling and shifting the sampled $L_S$ by the scale rate $s_F$ and the 2D coordinate $P_i$.

Ultimately, FSM generates more realistic nighttime flare images by incorporating existing synthetic flare images, applying random darkening operations, and adopting an adaptive light source intensity strategy. This approach effectively mitigates the photometric inconsistency issues in nighttime images and significantly enhances the diversity and quality of the training dataset.


\subsubsection{Noise Simulation Module}

In the previous section, we derived the calculation process of $I^{F}$. However, low-light images often suffer from amplified noise due to low photon counts and high system gain in camera sensors. Inspired by the physical priors of noise formation, we introduce the Noise Simulation Module (NSM) technique to simulate the noise distribution in nighttime scenes. Based on the shot-read noise model \cite{eld,feng2022learnability, TDDC}, we take the flare image as input and precisely simulate noise under low-light conditions in a physically decoupled manner. This approach ensures that the synthesized images closely resemble those captured in real low-light scenarios. Therefore, the final low-light image $I^{FN}$ can be expressed by the following formula:
\begin{equation}
I^{FN} = I^{F} + N ,
\end{equation}
where $I^{F}$ represents the flare image of the real scene, and $N $ refers to the summation of all physical noise. According to the physical noise models used in base image denoising \cite{rethinking}, the overall noise $N$ can be decomposed into four components: photon shot noise $N_{p}$, read noise $N_{read}$, row noise $N_{r}$, and quantization noise $N_{q}$:
\begin{equation}
N = KN_{p} + N_{read} + N_{r} + N_{q}.
\end{equation}

\begin{figure}[tbp]
    \centering
    \newcommand{\turnheightnew}{0.32\columnwidth}
    \hspace*{-1.8mm}  
    \begin{tabular}{@{}c@{\hskip 0.4mm}c@{\hskip 0.4mm}c@{\hskip 0.4mm}c@{}}
    \vspace{-0.5mm}
    {\rotatebox{90}{\hspace{6mm}\footnotesize{$I$}}} &
    \includegraphics[width=\turnheightnew]{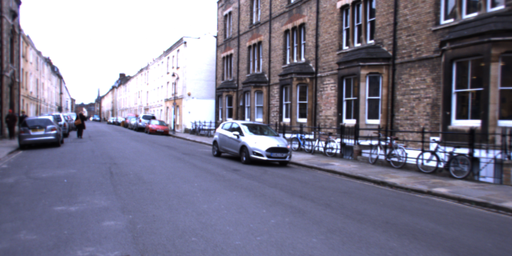} &
    \includegraphics[width=\turnheightnew]{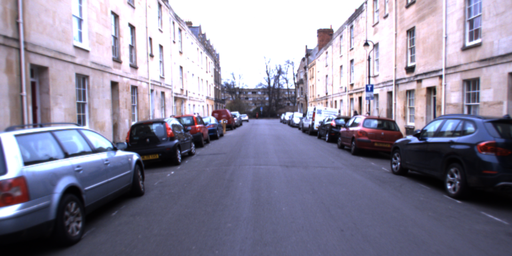} &
    \includegraphics[width=\turnheightnew]{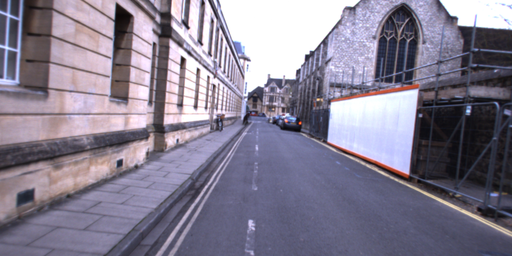} \\
    \vspace{-0.5mm}
    {\rotatebox{90}{\hspace{4.8mm}\footnotesize{$I^{F}$}}} &
    \includegraphics[width=\turnheightnew]{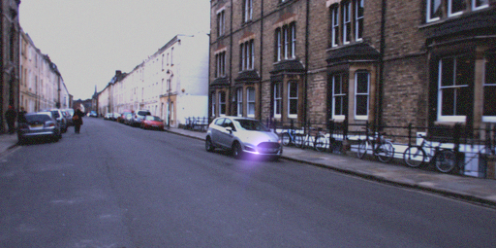} &
    \includegraphics[width=\turnheightnew]{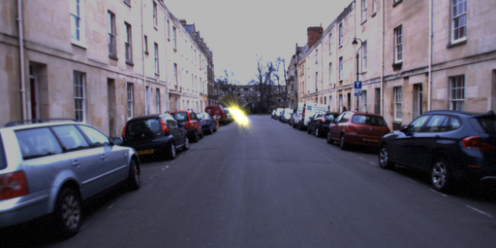} &
    \includegraphics[width=\turnheightnew]{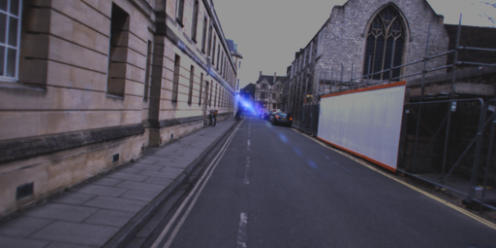} \\
    \vspace{-0.5mm}
    {\rotatebox{90}{\hspace{4mm}\footnotesize{$I^{FN}$}}} &
    \includegraphics[width=\turnheightnew]{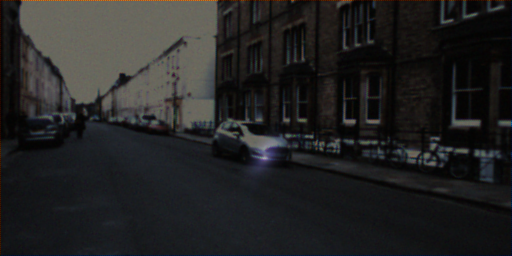} &
    \includegraphics[width=\turnheightnew]{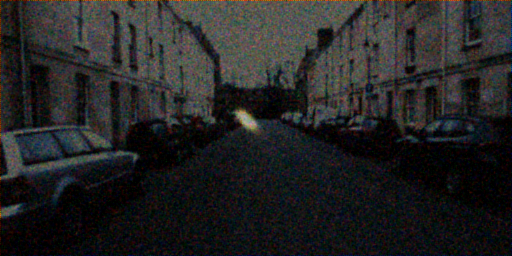} &
    \includegraphics[width=\turnheightnew]{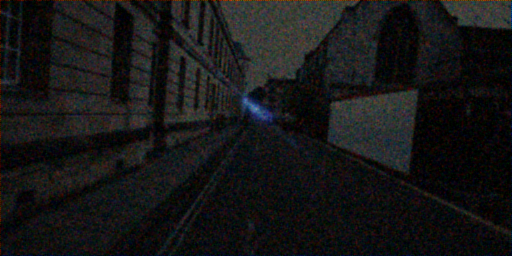} \\
     
    \end{tabular}
    \vspace{-2mm}
     \caption{\textbf{Paired visual examples for different scenes}, $I$ represents the image under normal lighting conditions, $I^{F}$ represents the visualization result of adding FSM to the normal lighting image, and $I^{FN}$ represents the visualization result of adding both FSM and NSM to the normal lighting image. 
     \vspace{-2mm}}
    \label{Fig: FN_samples}
    \vspace{-1mm}
    \Description{A block diagram showing the architecture of the proposed neural network method.}
\end{figure}

Where $K$ denotes the total system gain, whose value is determined by the camera's ISO setting, $N_{p}$ represents the photon shot noise, which depends on the incident light intensity and follows a Poisson distribution. $N_{read}$ is a unified term encompassing multiple noise sources, including dark current noise, thermal noise, and source follower noise. $N_{r}$ typically manifests as horizontal or vertical lines in the image and is modeled as a fixed offset applied to each column or row, sampled from a zero-mean Gaussian distribution with variance $\lambda_{row}$. $N_{q}$ is introduced by the finite bit depth of the sensor and is modeled as a uniform sampling of pixel values, with the variance parameterized by $\lambda_{quant}$. All of the above parameters follow the design principles outlined in ELD \cite{eld} and we give some paired visual examples of $ I^{FN} $ in Fig. \ref{Fig: FN_samples}.

\subsection{Low-Light Parameter-Efficient Fine-Tuning}
\label{sec:method_PEFT}

Although we synthesized large-scale paired depth data for daytime and low-light scenarios through LLDG, we observed that the directly fine-tuned foundation model exhibited unstable loss convergence during training, while full model optimization required substantial computational resources. To address these challenges, we introduced a Low-Light Parameter-Efficient Fine-Tuning strategy to optimize the foundation model for low-light depth estimation. The LLPEFT strategy integrates illumination guidance and multiscale feature fusion to guide and enhance the model's performance under challenging low-light conditions.

\subsubsection{Illumination Guidance}

To address the aforementioned issues, we analyzed that the main factors \cite{illu-transformer, AUX-self} stem from two aspects. First, high-ISO and long-exposure settings in dark scenes inevitably introduce noise artifacts. Second, uneven brightness in the images amplifies noise artifacts, leading to underexposure/overexposure and color distortion. To mitigate the challenges arising from training foundation models under low-light conditions, our approach injects illumination guidance into the training pipeline to guide the model in addressing noise artifacts and uneven photometric distribution in low-light images. Specifically, we introduce illumination guidance term $I_g^{FN} $ in the LLPEFT strategy, which is defined as follows:  
\begin{equation}
I_g^{FN} = mean_c \big( LLDG(I) \big),
\end{equation}

The operation $mean_c$ computes the mean value of each pixel along the channel dimension. As shown in Fig. \ref{Fig: framework}, the low-light image $ I^{FN} $ is synthesized from the daytime image $ I $ using LLDG, while $ I_g^{FN} $ is generated from $ I^{FN} $ using illumination guidance. This process effectively simplifies the illumination guidance procedure by converting the low-light image into a grayscale representation using a simple but efficient approach. This transformation helps to reduce noise, enhance the brightness distribution, and simplify the information structure. Although this approach results in the loss of color information, it does not significantly affect the final outcome, as the color information is already preserved in the low-light image $ I^{FN} $. 

Ultimately, by introducing this illumination guidance, the model can focus on learning robust feature representations, thereby mitigating noise amplification, color distortion, and artifacts caused by long exposure and high ISO settings.

\subsubsection{Multiscale Feature Fusion}

To effectively transform the input low-light image $I^{FN}$ and its illumination guidance image $I_g^{FN} $ from the image space to the feature space, we propose an augmented feature fusion method that incorporates illumination guidance. Specifically, we first concatenate the low-light image and illumination guidance image along the channel dimension to form a new augmented low-light image, denoted as the low-light auxiliary image $I_A^{FN}$. Subsequently, we design and employ a Multiscale Feature Fusion. This method captures multiscale contextual information and dynamically adjusts feature weights, ensuring that the extracted features comprehensively represent the multi-level information embedded in the low-light image and its illumination guidance. The detailed process is as follows:

Firstly, let the low-light image $I^{FN} \in \mathbb{R}^{H \times W \times C_1} $ and the illumination guidance $I_g^{FN} \in \mathbb{R}^{H \times W \times C_2} $. We concatenate them along the channel dimension to produce the low-light auxiliary image $I_A^{FN} $, as defined by:
\begin{align}
I_A^{FN} = Concat(I^{FN}, I_g^{FN}),
\end{align}

where $I_A^{FN} \in \mathbb{R}^{H \times W \times (C_1 + C_2)} $. This image integrates the visual information of the low-light image with the statistical information of the illumination guidance, providing a more comprehensive input for subsequent feature extraction.

Subsequently, the low-light auxiliary image $I_A^{FN} $ is processed by multiscale feature fusion. This method consists of three parallel convolutional layers with kernel sizes of $1 \times 1 $, $3 \times 3 $, and $5 \times 5 $, each with $C_3 $ channels. The corresponding output features are denoted as:
\begin{align}
\mathbf{E}_1 &= Conv_{1\times1}(I_A^{FN}), \\
\mathbf{E}_2 &= Conv_{3\times3}(I_A^{FN}), \\
\mathbf{E}_3 &= Conv_{5\times5}(I_A^{FN}),
\end{align}

where $\mathbf{E}_1, \mathbf{E}_2, \mathbf{E}_3 \in \mathbb{R}^{H \times W \times C_3} $ represent the multiscale features extracted with different receptive fields.

To dynamically integrate multiscale features, we introduce the Softmax function to process features from different receptive fields. Specifically, we compute the attention weights $\alpha_i $ for each scale feature as:
\begin{align}
\alpha_i = Softmax(\mathbf{W}_i \cdot \mathbf{E}_i + \mathbf{b}_i), \quad i \in \{1, 2, 3\},
\end{align}

where $\mathbf{W}_i $ and $\mathbf{b}_i $ are learnable parameters, and $\alpha_i $ measures the importance of the $i $-th scale feature.

Subsequently, the fused feature representation is given by:
\begin{align}
\mathbf{E}_{fused} = \sum_{i=1}^3 \alpha_i \cdot \mathbf{E}_i,
\end{align}

where $\mathbf{E}_{fused} \in \mathbb{R}^{H \times W \times C_3} $. Finally, a $1 \times 1 $ convolutional layer is applied to reduce the channel dimension, producing the final feature representation $\mathbf{E}_A^{FN} $:
\begin{align}
\mathbf{E}_A^{FN} = Conv_{1\times1}(\mathbf{E}_{fused}),
\end{align}

The extracted low-light feature map $\mathbf{E}_A^{FN} \in \mathbb{R}^{H \times W \times C_1} $ integrates multi-level information from both the low-light input image and its illumination guidance image, forming a robust representation that provides reliable feature support for subsequent modules. Additionally, both the extracted feature map and the low-light image $I_{FN}$ are processed through convolutional layers before being fed into the patch embed module, followed by vision transformers \cite{vit-transformers} to generate the final depth map.

Ultimately, multiscale feature fusion transforms the low-light image $I_{FN}$ and its illumination guidance $I_{FN}^g$ from the image space to the feature space, extracting a piece of comprehensive multi-level information in the form of a low-light feature map $\mathbf{E}_A^{FN}$. This significantly enhances the robustness of the depth estimation model under challenging low-light conditions. 

\section{Experiments}
\label{sec:experiments}
\subsection{Implementation Details} 

\begin{table*}[!t]
	\begin{center}

	\scriptsize
    \resizebox{1\linewidth}{!}{
        \begin{tabular}{c| l | c | c | c | c c c c | ccc}
            \hline
            Type & Method & Train on & Train Res.  & Max depth & \cellcolor{cell1} ABS rel$\downarrow$ & \cellcolor{cell1} Sq rel$\downarrow$  & \cellcolor{cell1} RMSE$\downarrow$ & \cellcolor{cell1} RMSE log$\downarrow$  & \cellcolor{cell2} $\delta_{1} \uparrow$ & \cellcolor{cell2} $\delta_{2} \uparrow$ & \cellcolor{cell2} $\delta_{3} \uparrow$\\
            \hline
            \multicolumn{12}{c}{\cellcolor{cell3} Test on nuScenes-Night} \\

            \multirow{2}{*}{DT}
            & MonoViT\cite{Monovit}           & N $ d \& n $ &  640 $\times$ 320   & 60     & 1.726     	& 93.031    	& 30.321    	& 2.183     	& 0.143     	& 0.291     	& 0.437  \\
            & WSGD\cite{WSGD}& N $ d \& n $ &  640 $\times$ 320   & 60     & 0.663     	& 9.573     	& 15.200    	& 0.755     	& 0.199     	& 0.388     	& 0.567  \\
            \hline
            \multirow{4}{*}{DA}
            & ITDFA\cite{ITDFA}               & N $ d \& n $ &  640 $\times$ 320   & 60       & 0.337     	 & 4.511     	& 10.118    	& 0.403     	& 0.515     	& 0.767     	& 0.890  \\        
            & RNW\cite{RNW}                   & N $ d \& n $ &  640 $\times$ 320   & 60       & 0.341          & 5.516     	& 11.152    	& 0.406     	& 0.531     	& 0.789     	& 0.902 \\
            & ADDS\cite{adds}                 & N $ d \& n $ &  640 $\times$ 320   & 60       & 0.299     	& 4.790     	& 10.372    	& 0.371     	& 0.620  	& 
            0.814     	& 0.907 \\
            \hline
            \multirow{8}{*}{G}
            & ITDFA\cite{ITDFA}               & R $ d \& n $ &  640 $\times$ 320   & 60       & 0.362           & 3.760     	& 10.252    	& 0.441     	& 0.418     	& 0.702     	& 0.867  \\
            & RNW\cite{RNW}                   & R $ d \& n $ &  640 $\times$ 320   & 60       & 0.376           & 4.732     	& 11.193    	& 0.506     	& 0.451     	& 0.712     	& 0.835 \\
            & ADDS\cite{adds}                 & R $ d \& n $ &  640 $\times$ 320   & 60       & 0.322           & 4.401     	& 10.584    	& 0.397     	& 0.527     	& 0.786     	& 0.892 \\
            & MonoFormer\cite{MonoFormer} & K            &  768 $\times$ 256   & 60       & 0.307     	    & 3.591     	& 10.162    	& 0.413     	& 0.521     	& 0.762     	& 0.872  \\
            
            
            & TDDC\cite{TDDC}           & K            &  768 $\times$ 256   & 60       & \underline{0.259} & 3.147& \underline{8.547}& 0.344& \textbf{0.641}& \underline{0.850}& 0.928  \\
            
         & Depth Anything\cite{depthanything}        & UD \& LD        &  -   & 60   &   0.302  &   3.051  &   9.233  &   0.321  &   0.487  &   0.784  &   \underline{0.972}  \\  

         & Depth Anything V2\cite{depthanything_v2}          &  RD  \& SD            &  -   & 60    &   0.272  &   \underline{2.551}  &   8.576  &   \underline{0.304}  &   0.518  &   0.843  &   0.971  \\
    
             & \textbf{Ours DepthDark}                   & H, VK            & -  & 60  &   \textbf{0.210}  &   \textbf{1.910} &   \textbf{7.764}  &  \textbf{0.260}   &   \underline{0.630}  &  \textbf{0.914}   &  \textbf{0.976}   \\


            \multicolumn{12}{c}{\cellcolor{cell3} Test on RobotCar-Night} \\
            \multirow{2}{*}{DT}
            & MonoViT\cite{Monovit}           & R $ d \& n $ &  640 $\times$ 320  & 40       & 0.513     	    & 13.558    	& 9.867     	& 0.479     	& 0.588     	& 0.846     	& 0.918  \\
            & WSGD\cite{WSGD}& R $ d \& n $ &  640 $\times$ 320  & 40       & \underline{0.202}& 1.835     	& 5.985     	& 0.231& 0.737& 0.934& 0.977   \\
            \hline
            \multirow{4}{*}{DA}
            & ITDFA\cite{ITDFA}               & R $ d \& n $ &  640 $\times$ 320  & 40       & 0.266     	    & 3.010     	& 8.293     	& 0.287     	& 0.567     	& 0.888     	& 0.962  \\
            & ADDS\cite{adds}                 & R $ d \& n $ &  640 $\times$ 320  & 40       & 0.209     	& 2.179     	& 6.808     	& 0.254     	& 0.704     	& 0.918     	& 0.965 \\
            & RNW\cite{RNW}                   & R $ d \& n $ &  640 $\times$ 320  & 40       & 0.197 & 1.789     	& 5.896& 0.234& 0.742 & 0.930     	& 0.972 \\
            \hline   

            \multirow{8}{*}{G}     
            & ITDFA\cite{ITDFA}               & N $ d \& n $ &  640 $\times$ 320  & 40       & 0.302     	    & 3.692     	& 8.642     	& 0.327     	& 0.548     	& 0.852     	& 0.938   \\
            & ADDS\cite{adds}                 & N $ d \& n $ &  640 $\times$ 320  & 40       & 0.265     	    & 3.651     	& 8.700     	& 0.309     	& 0.640     	& 0.870     	& 0.945 \\
            & RNW\cite{RNW}                   & N $ d \& n $ &  640 $\times$ 320  & 40       & 0.237     	    & 2.958     	& 8.187     	& 0.298     	& 0.683     	& 0.885     	& 0.948 \\
            & MonoFormer\cite{MonoFormer} & K            &  768 $\times$ 256  & 40       & 0.289     	    & 2.893     	& 7.468     	& 0.302     	& 0.543     	& 0.873     	& 0.964  \\
            & TDDC\cite{TDDC}           & K            &  768 $\times$ 256   & 40       & 0.210     	    & \underline{1.515}& \underline{5.386}& \underline{0.238}     	& 0.676     	& \underline{0.936} & \underline{0.980 } \\

            & Depth Anything\cite{depthanything}     & UD \& LD        &  -  & 40  &   0.302  &   3.331  &   6.622  &   0.314  &   0.635  &   0.822  &   0.918  \\
                     
             & Depth Anything V2\cite{depthanything_v2}     &  RD  \& SD            &  -   & 40   &   0.235  &   2.474  &   6.239  &   0.268  &  \underline{0.697}   &   0.868  &   0.946  \\          
    
             & \textbf{Ours DepthDark }       & H, VK            & -   & 40   &    \textbf{0.157}  &  \textbf{1.063}   &   \textbf{4.284}   & \textbf{0.202}    &  \textbf{0.760}  &  \textbf{0.941}   &  \textbf{0.985}    \\
            \hline
        \end{tabular}}
        \caption{ The best results are in \textbf{bold}, while the second-best scores are \underline{underline}. H, VK, K, N, and R denote the Hypersim, Virtual KITTI, KITTI, nuScenes-Night, and RobotCar-Night datasets. UD and LD refer to Depth Anything being trained on 62M unlabeled datasets and 1.5M labeled datasets. RD and SD denote that Depth Anything V2 is trained on 62M real datasets and 0.5M synthetic datasets. $d$ and $n$ refer to the daytime and nighttime training splits proposed by RNW~\protect\cite{RNW} and ADDS~\protect\cite{adds}. Max depth refers to the upper limit of the ground truth depth values. 
    \vspace{-2mm}
    }
    \label{Table:all_result}
    \end{center}
    \vspace{-2mm}
\end{table*}

To train foundation models for monocular depth estimation under low-light conditions, we implemented DepthDark using the PyTorch framework. During training, we adopted the training settings from Depth Anything V2 and used the DPT decoder \cite{dpt-vision} in the ZoeDepth \cite{zoedepth} pipeline, built on the DINOv2 encoder. All images in the training dataset were resized to a resolution of 518 $\times$ 518 for training, and our model was eventually fine-tuned for downstream monocular depth estimation tasks.

Additionally, data augmentation techniques, including random cropping and horizontal flipping, were employed to enhance the training dataset, ensuring efficient use of memory resources on a single Nvidia RTX 3090 GPU. 

\subsection{Datasets} 
\subsubsection{Training Datasets} 

\begin{figure*}[ht]
    \centering
    \newcommand{\turnheightnew}{0.33\columnwidth}
    \begin{tabular}{@{\hskip 0mm}c@{\hskip 1mm}c@{\hskip 1mm}c@{\hskip 1mm}c@{\hskip 1mm}c@{\hskip 1mm}@{\hskip 0mm}c@{\hskip 1mm}c@{\hskip 1mm}c@{\hskip 1mm}c@{\hskip 1mm}c@{\hskip 1mm}c@{}}
    {
 
    \rotatebox{90}{\hspace{3mm}\fontsize{5pt}{5pt}\selectfont{N-N \& R-N}}} &
    \includegraphics[width=\turnheightnew]{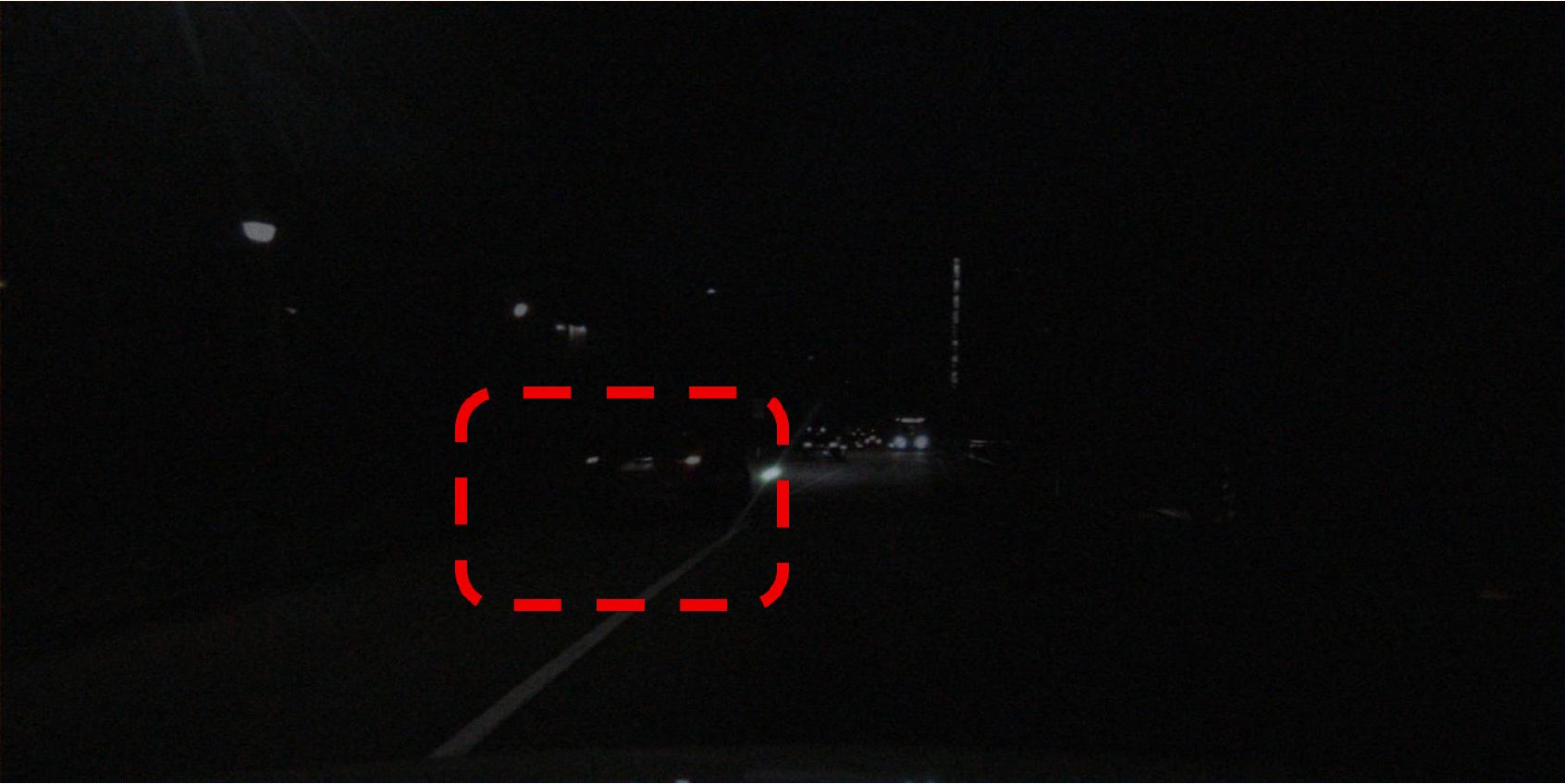} &
    \includegraphics[width=\turnheightnew]{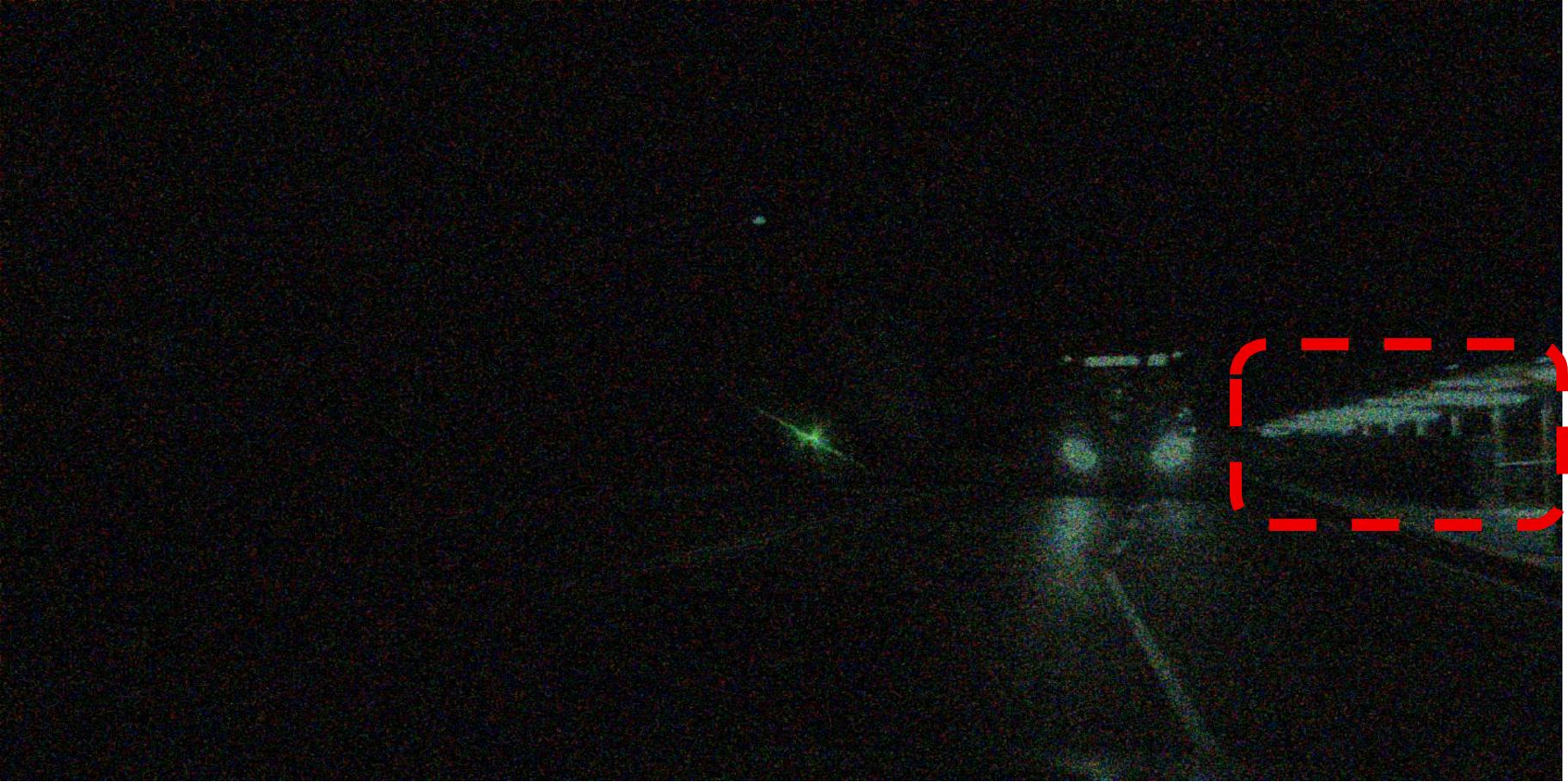} &
    \includegraphics[width=\turnheightnew]{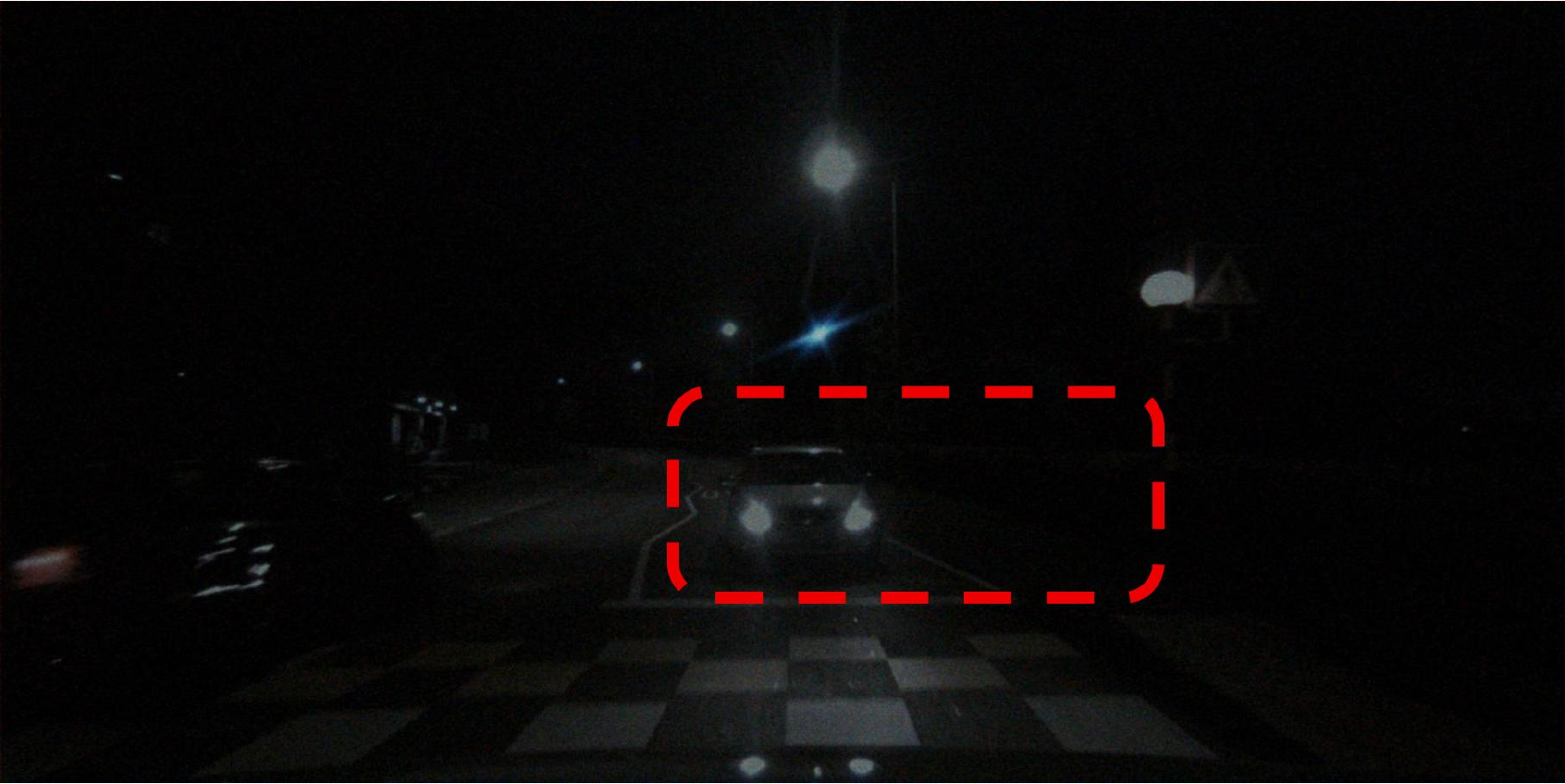} &
    
    \includegraphics[width=\turnheightnew]{img/Fig_visible/nuscenes-robot-night_pdf/nuscen_night_3_good/69_11.pdf} &
    \includegraphics[width=\turnheightnew]{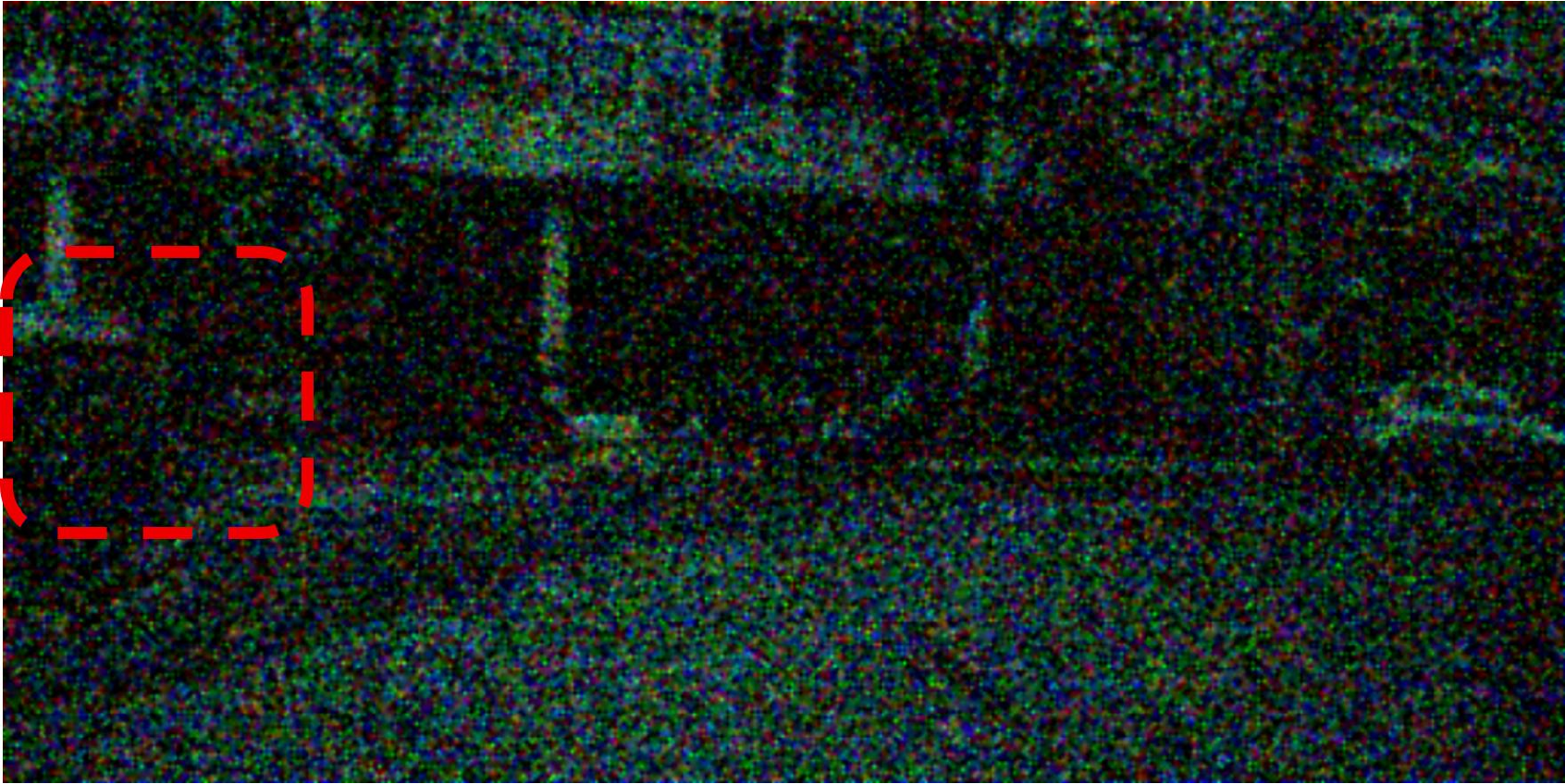} &
    \includegraphics[width=\turnheightnew]{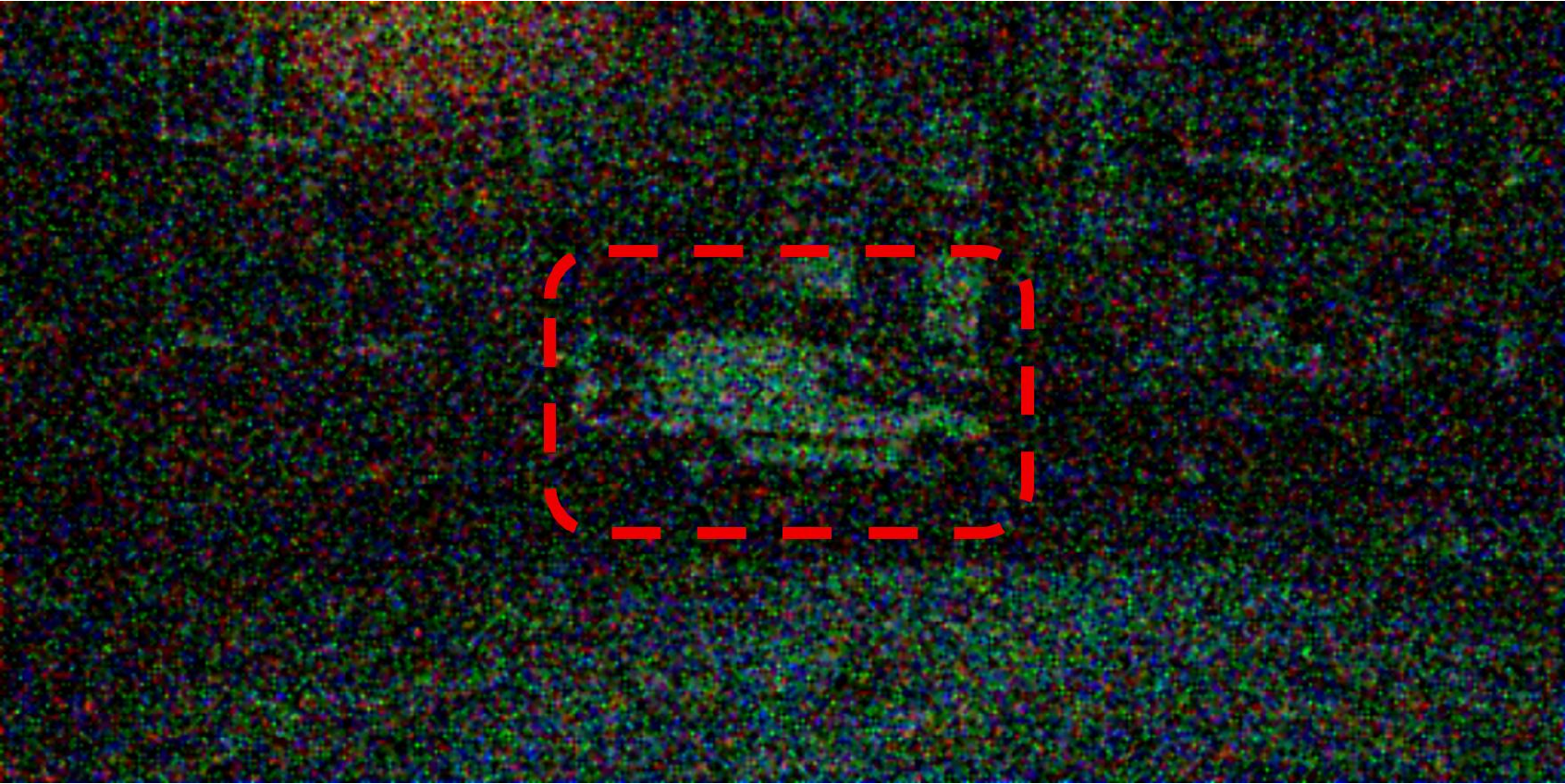} 
    \\
 
    \rotatebox{90}{\hspace{5mm}\fontsize{5pt}{5pt}\selectfont{ADDS}
    }&
   \includegraphics[width=\turnheightnew]{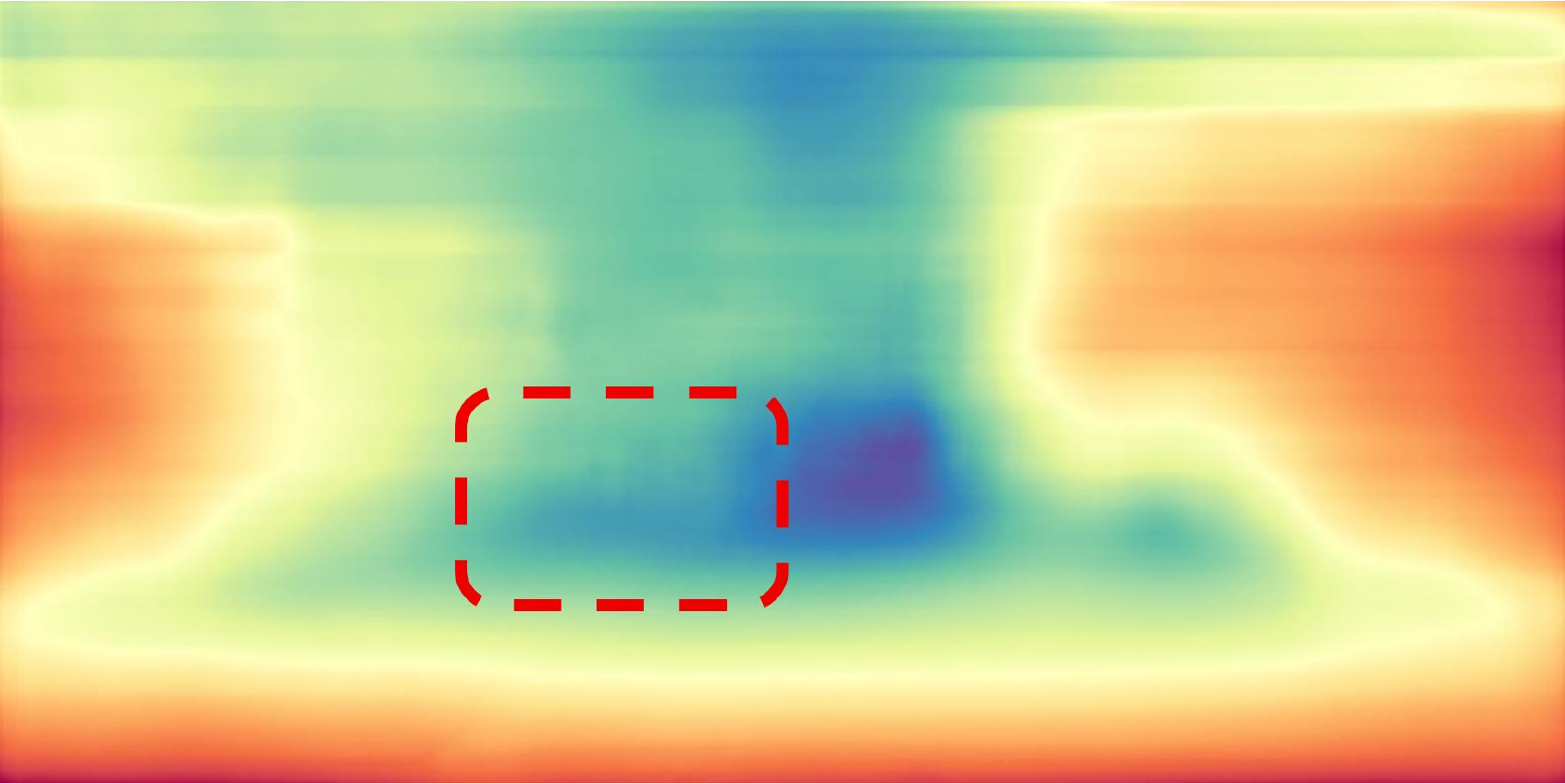} &
    \includegraphics[width=\turnheightnew]{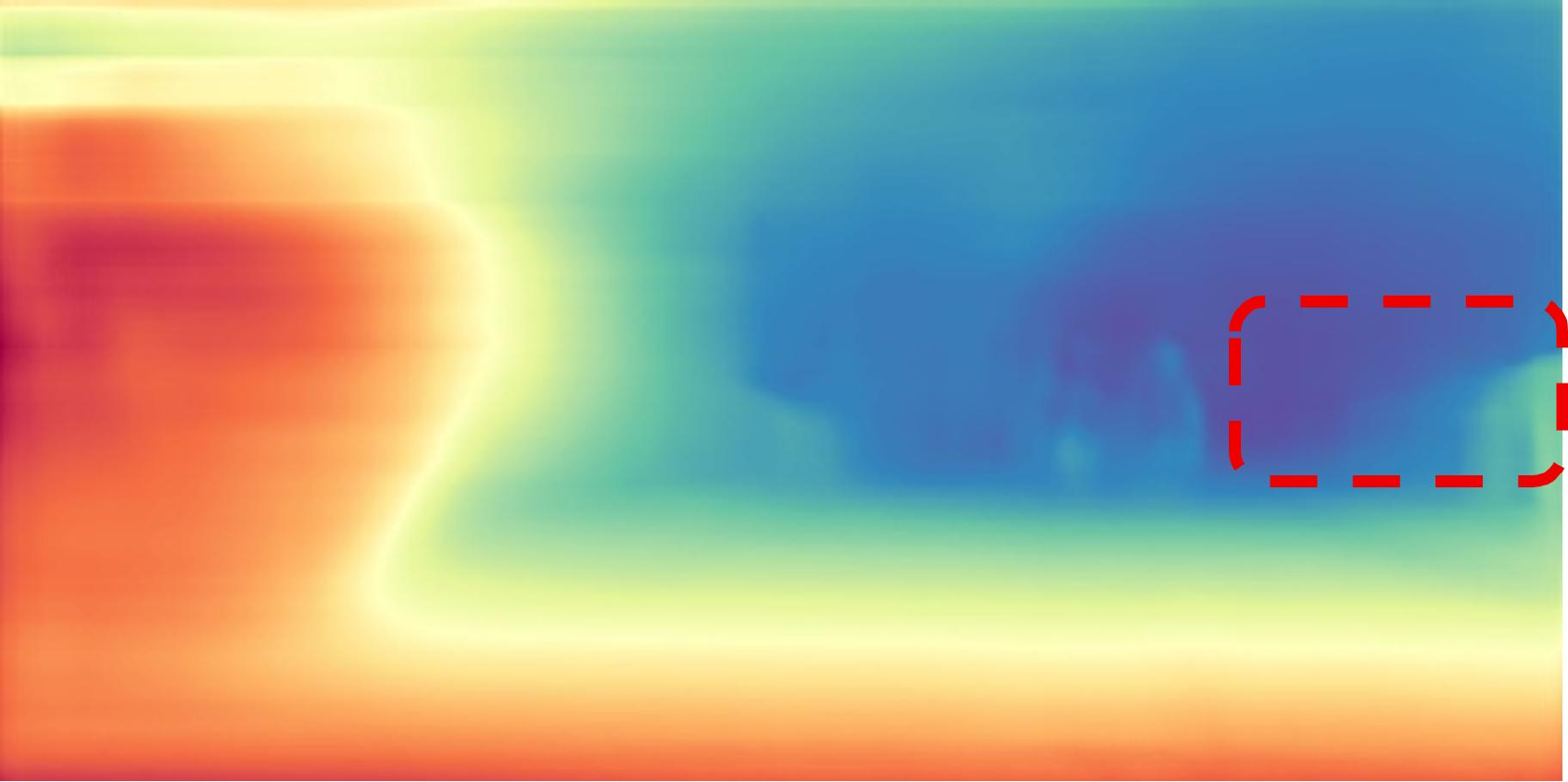} &
    \includegraphics[width=\turnheightnew]{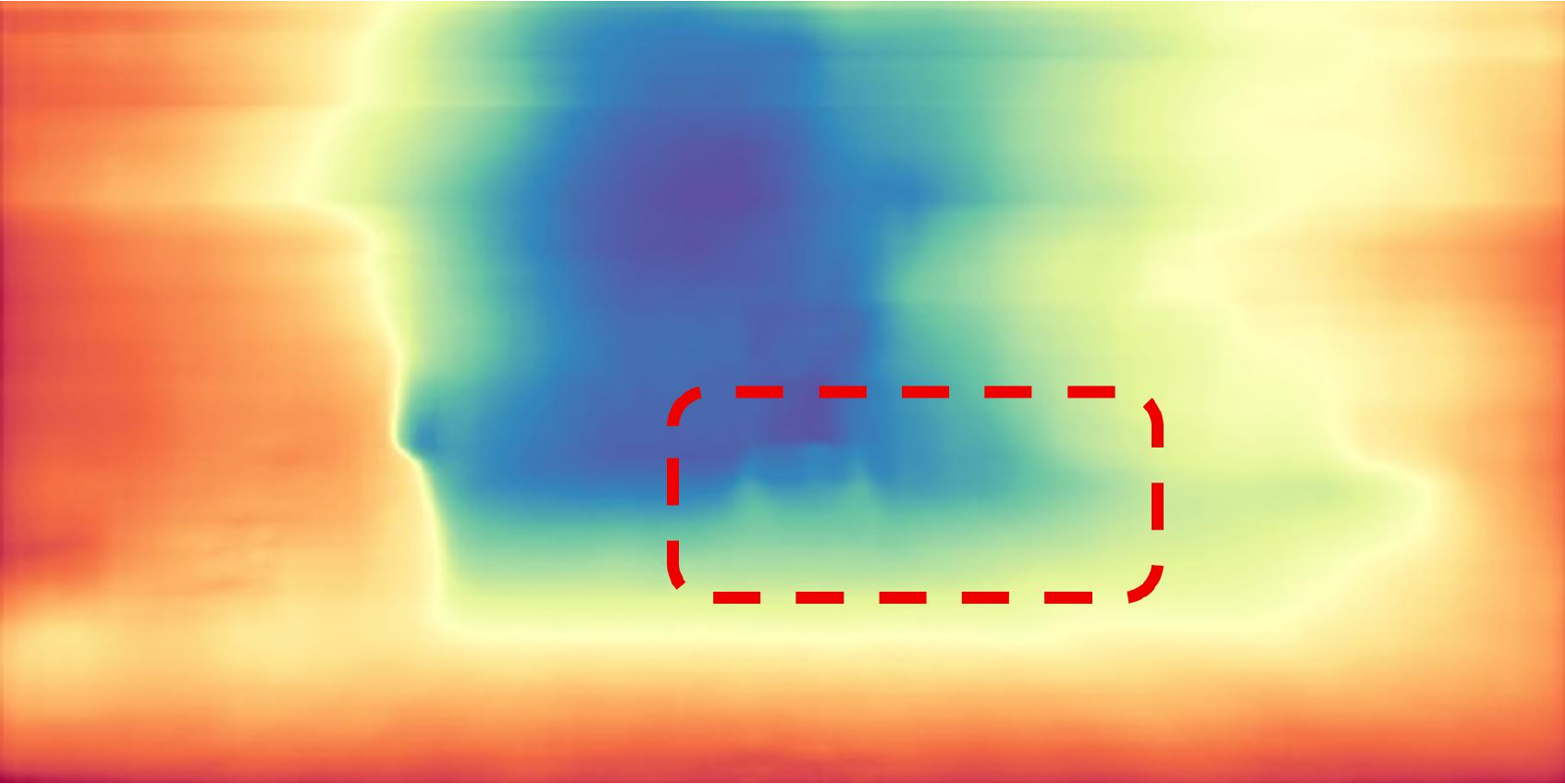} &
    
    \includegraphics[width=\turnheightnew]{img/Fig_visible/nuscenes-robot-night_pdf/nuscen_night_3_good/69_12.pdf} &
    \includegraphics[width=\turnheightnew]{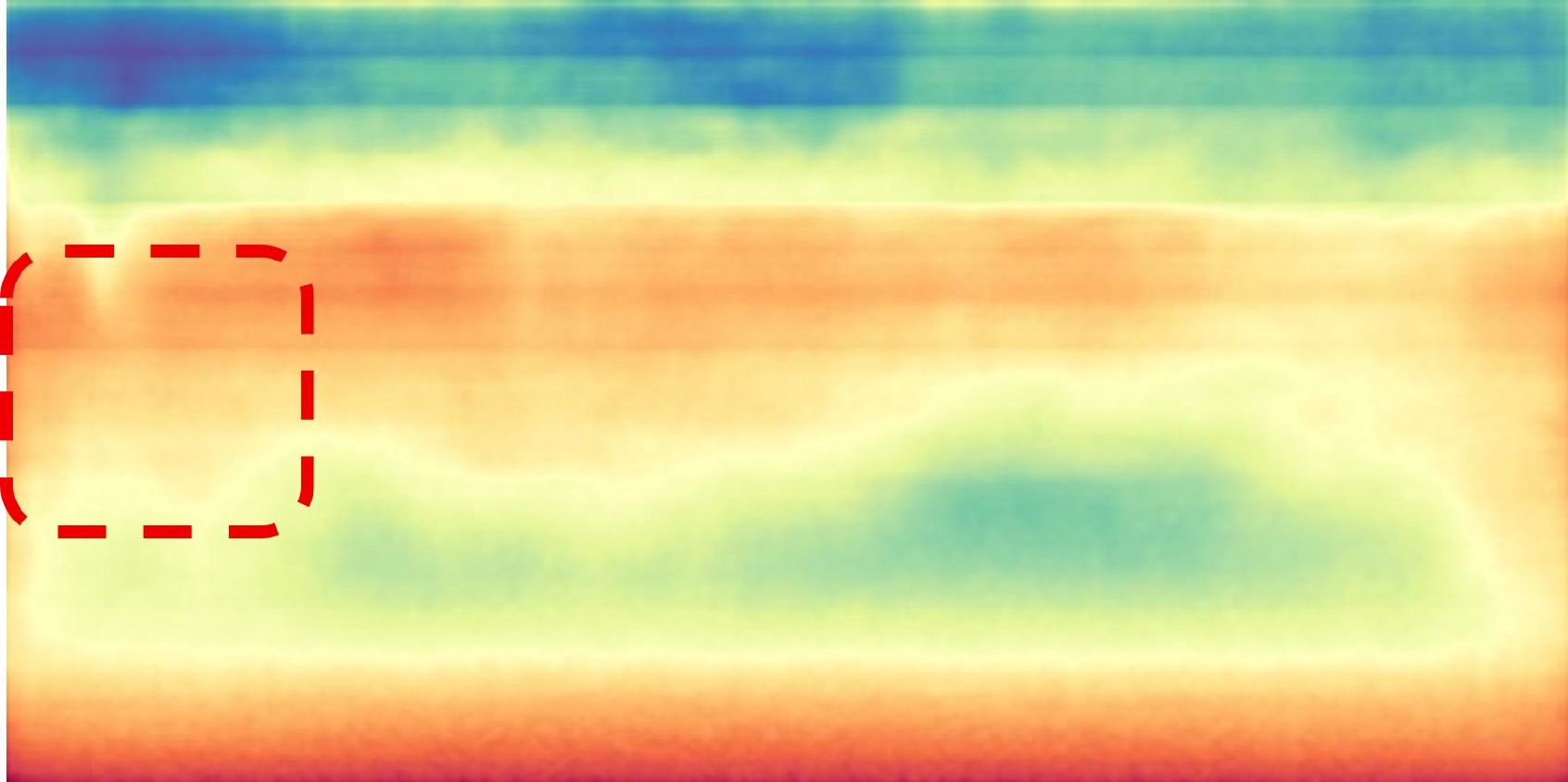} &
    \includegraphics[width=\turnheightnew]{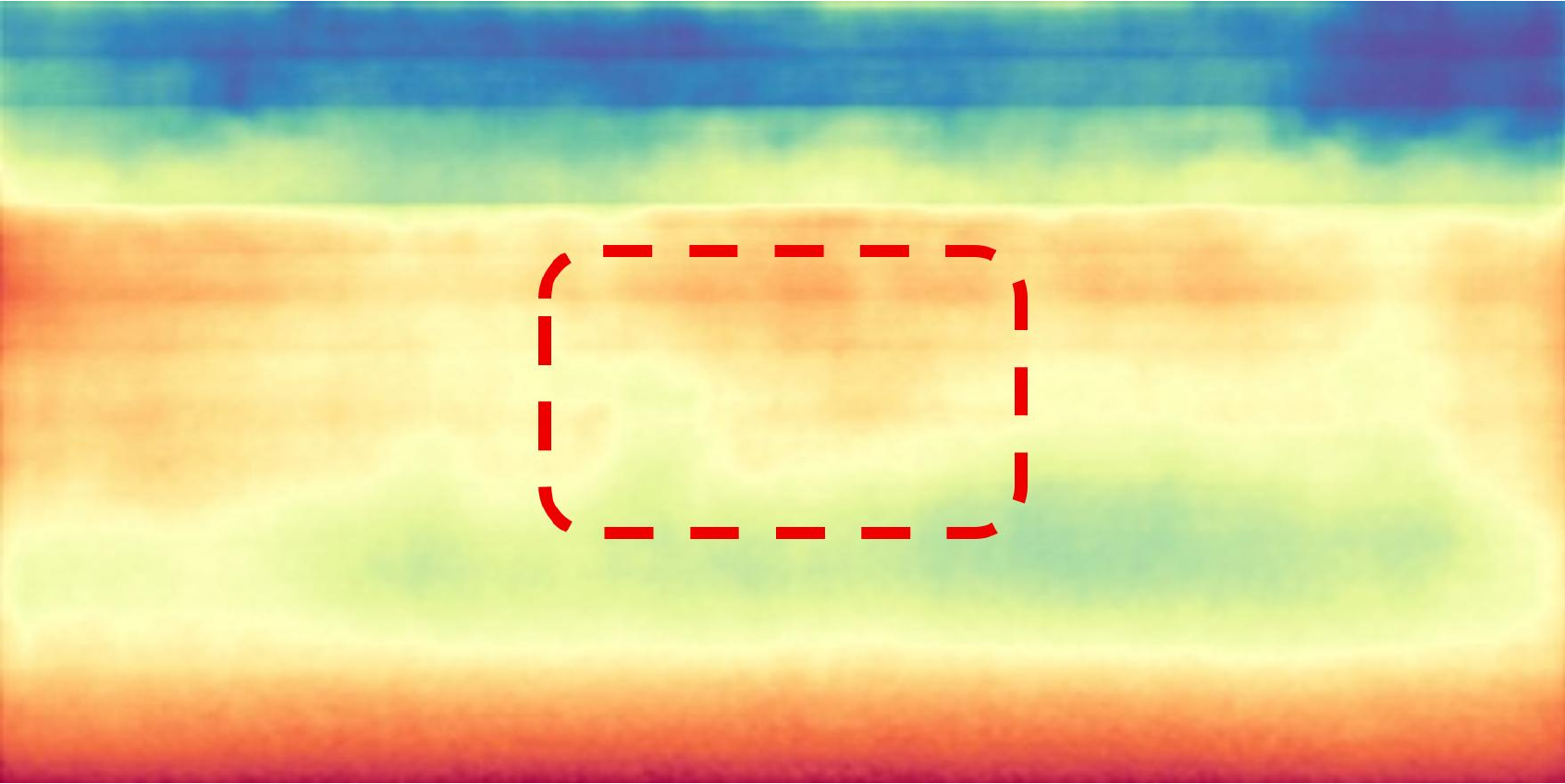} 
    \\
  
    \rotatebox{90}{\hspace{1mm}\fontsize{5pt}{5pt}\selectfont{Depth Anything}
    }&
    \includegraphics[width=\turnheightnew]{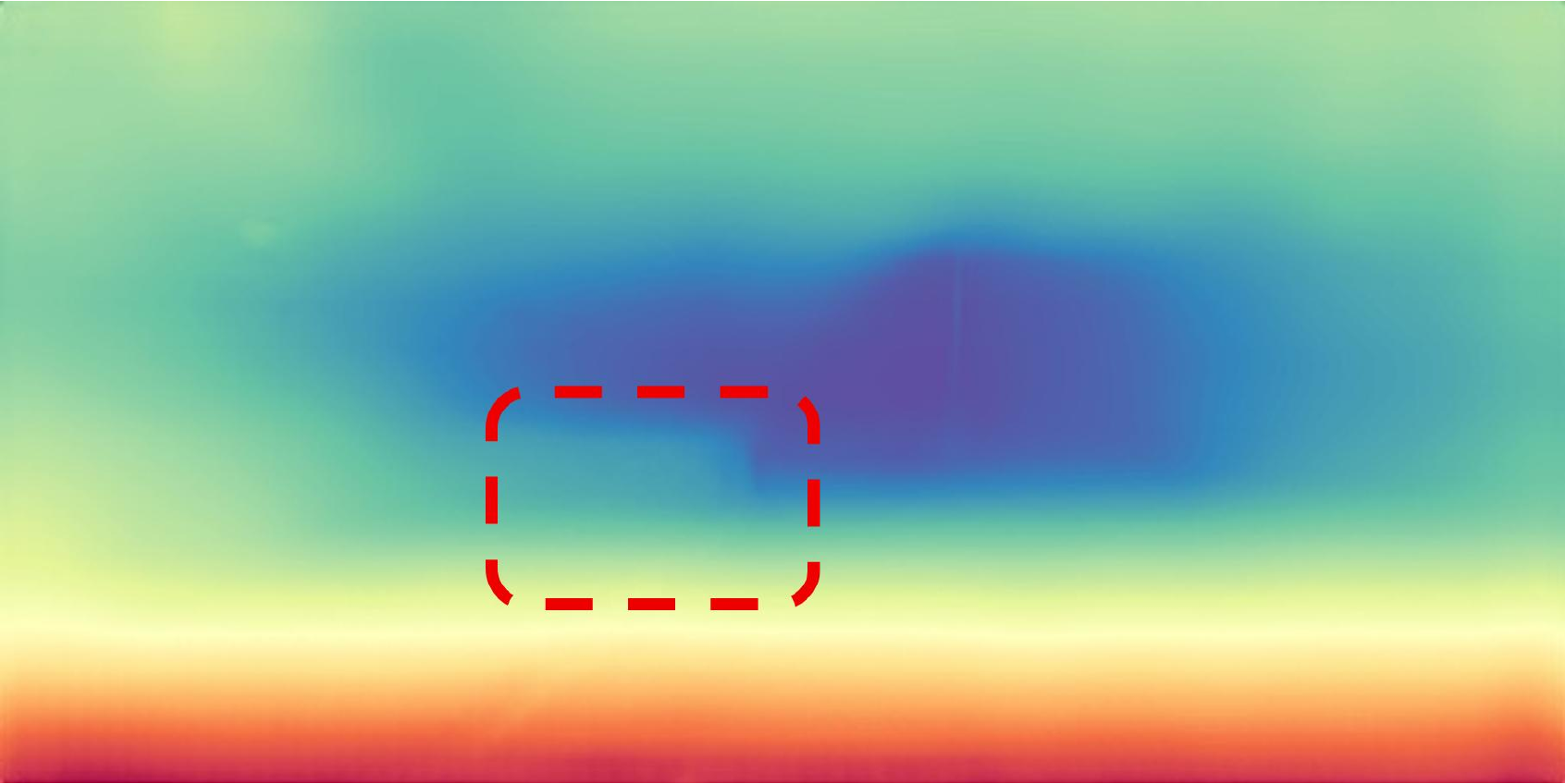} &
    \includegraphics[width=\turnheightnew]{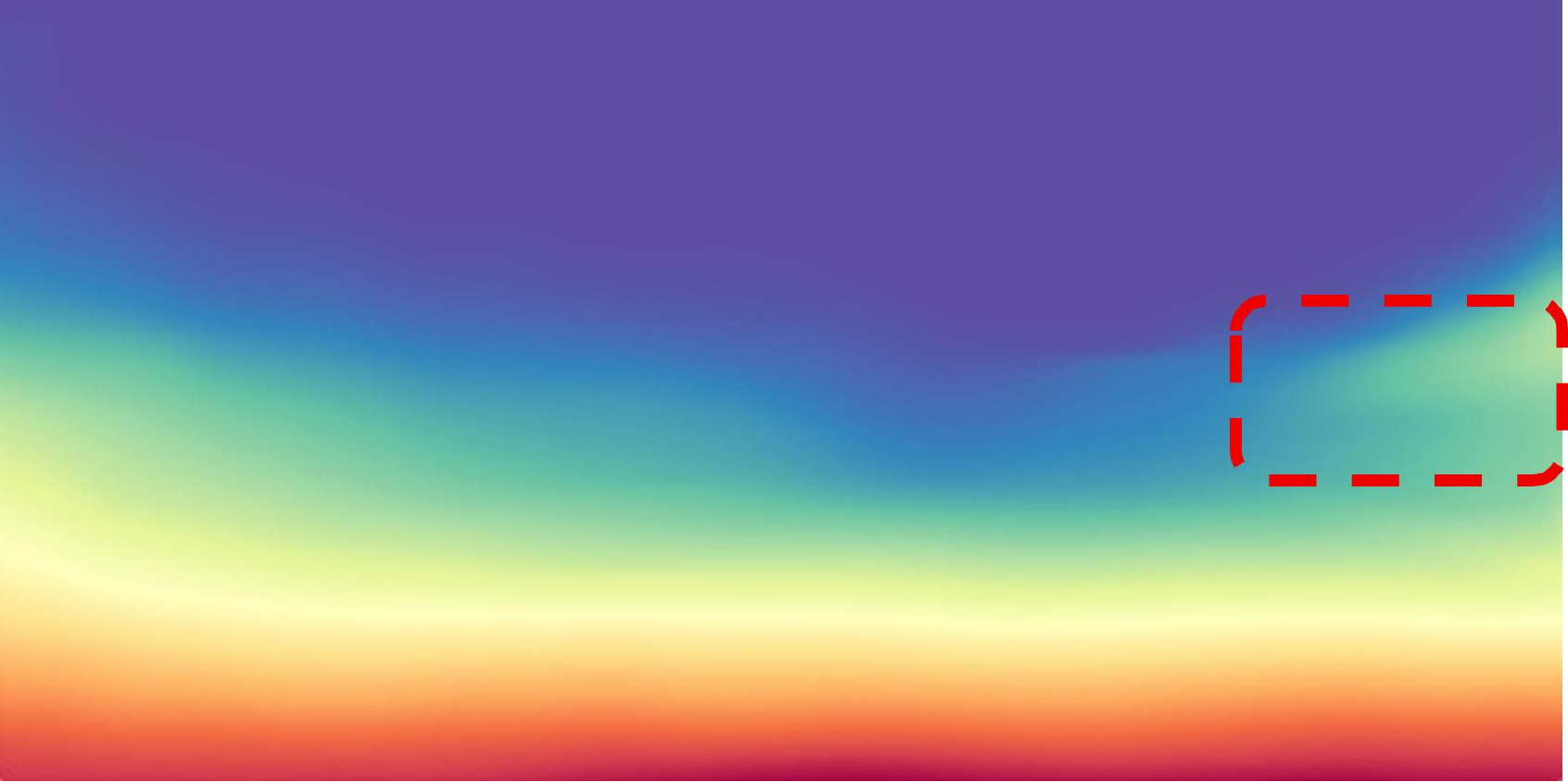} &
    \includegraphics[width=\turnheightnew]{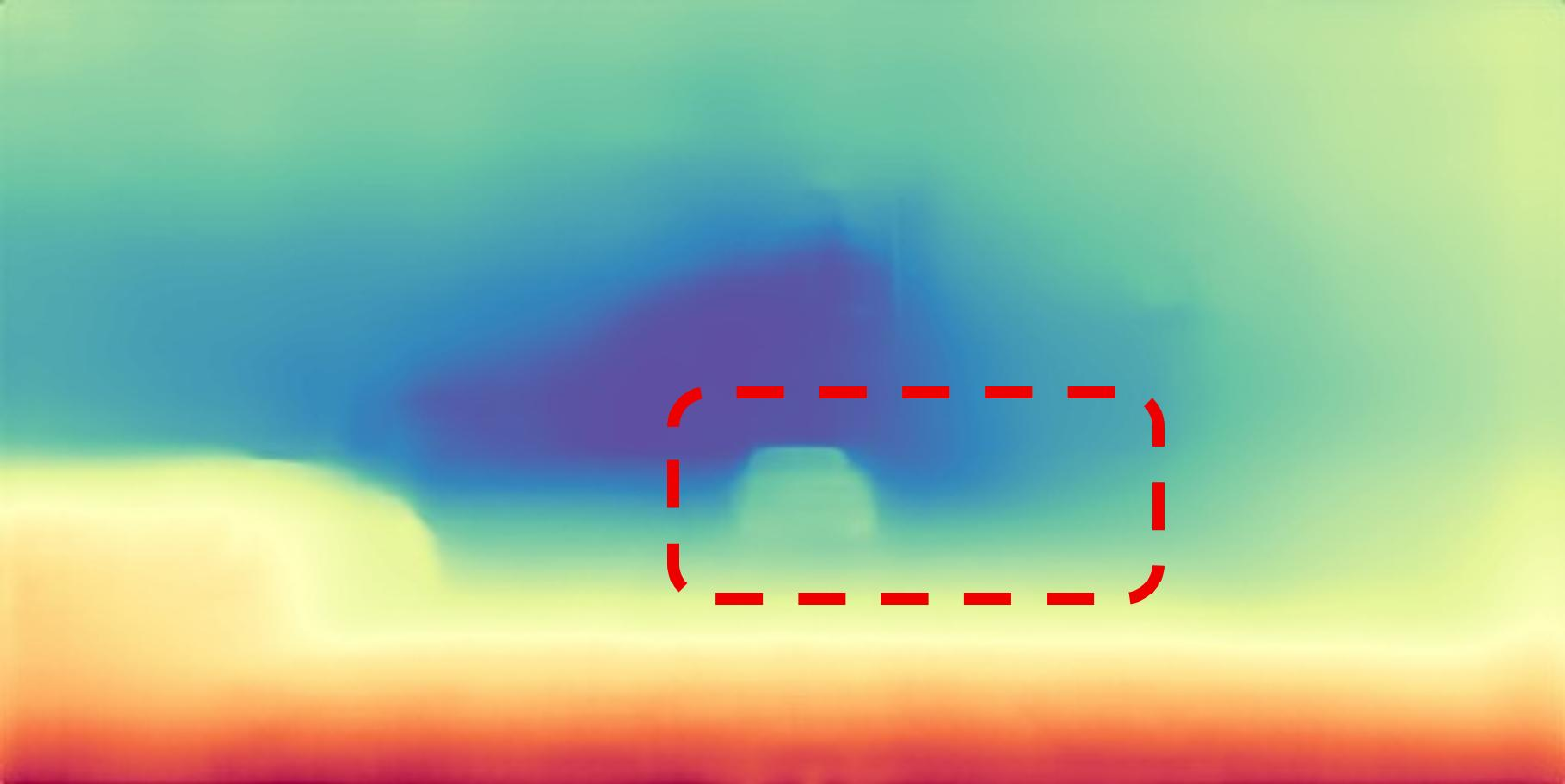} &
    
    \includegraphics[width=\turnheightnew]{img/Fig_visible/nuscenes-robot-night_pdf/nuscen_night_3_good/69_13.pdf} &
    \includegraphics[width=\turnheightnew]{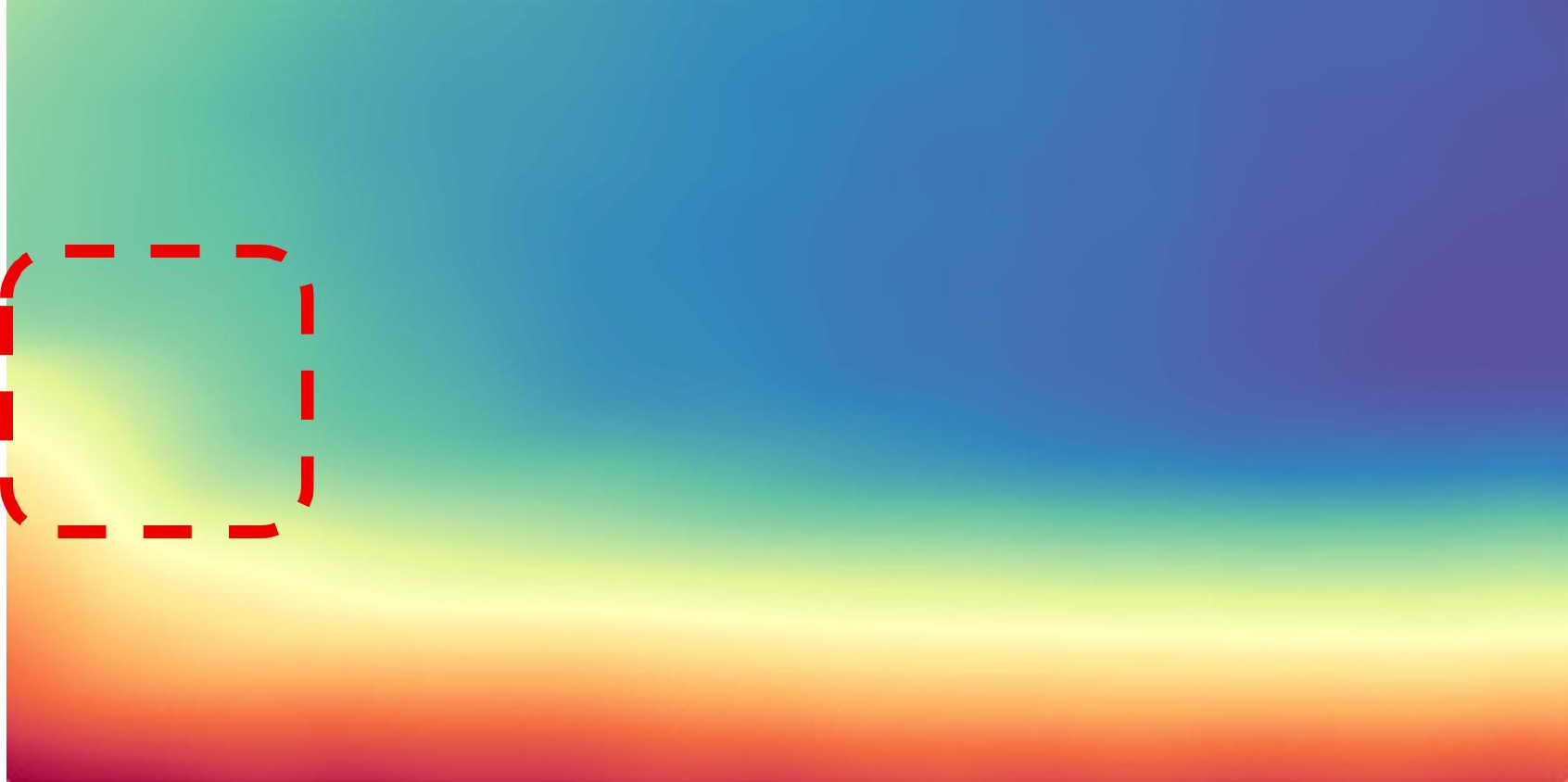} &
    \includegraphics[width=\turnheightnew]{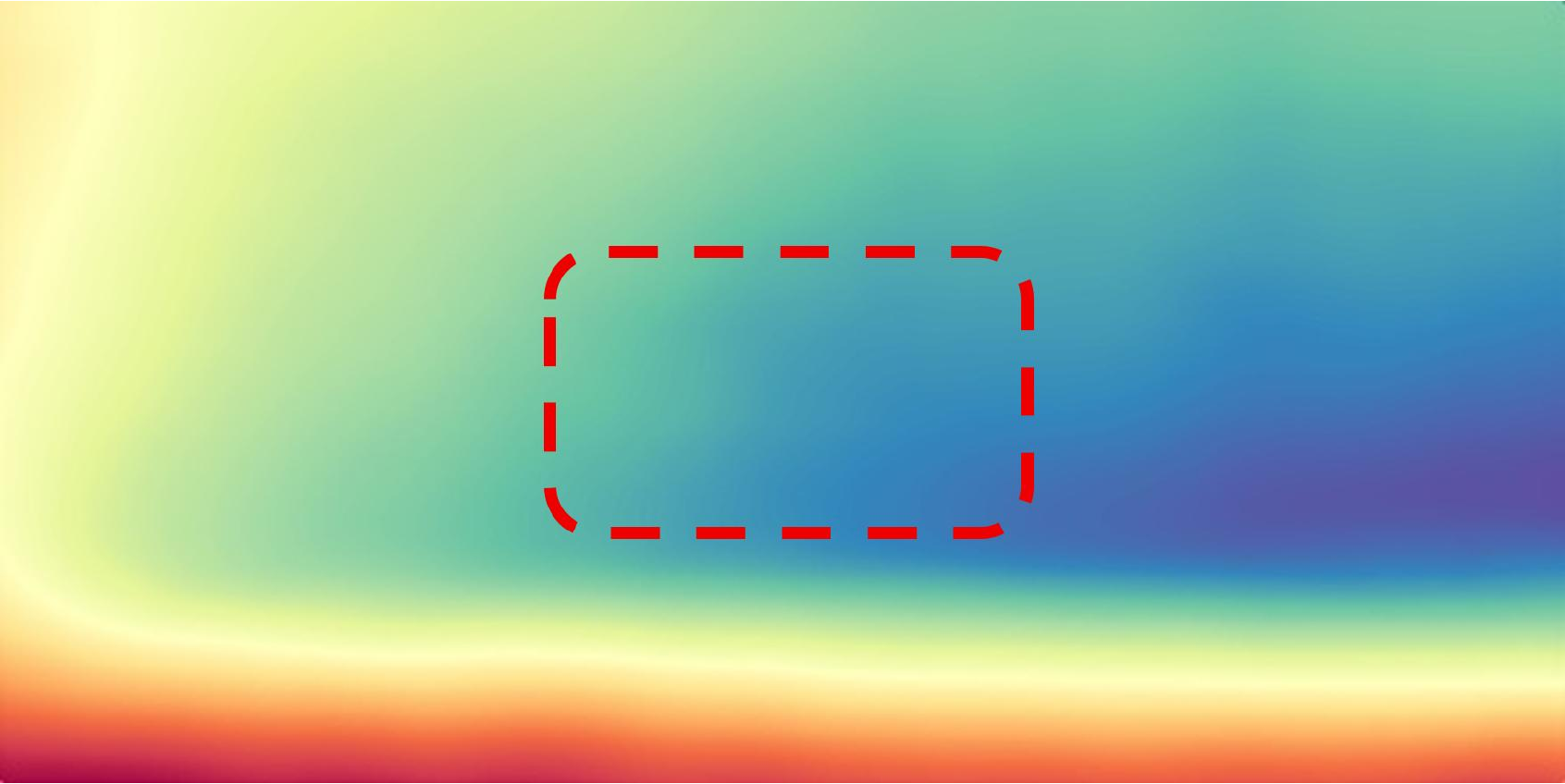} 
    \\

    \rotatebox{90}{\hspace{0mm}\fontsize{5pt}{5pt}\selectfont{Depth Anything V2}
    }&
    \includegraphics[width=\turnheightnew]{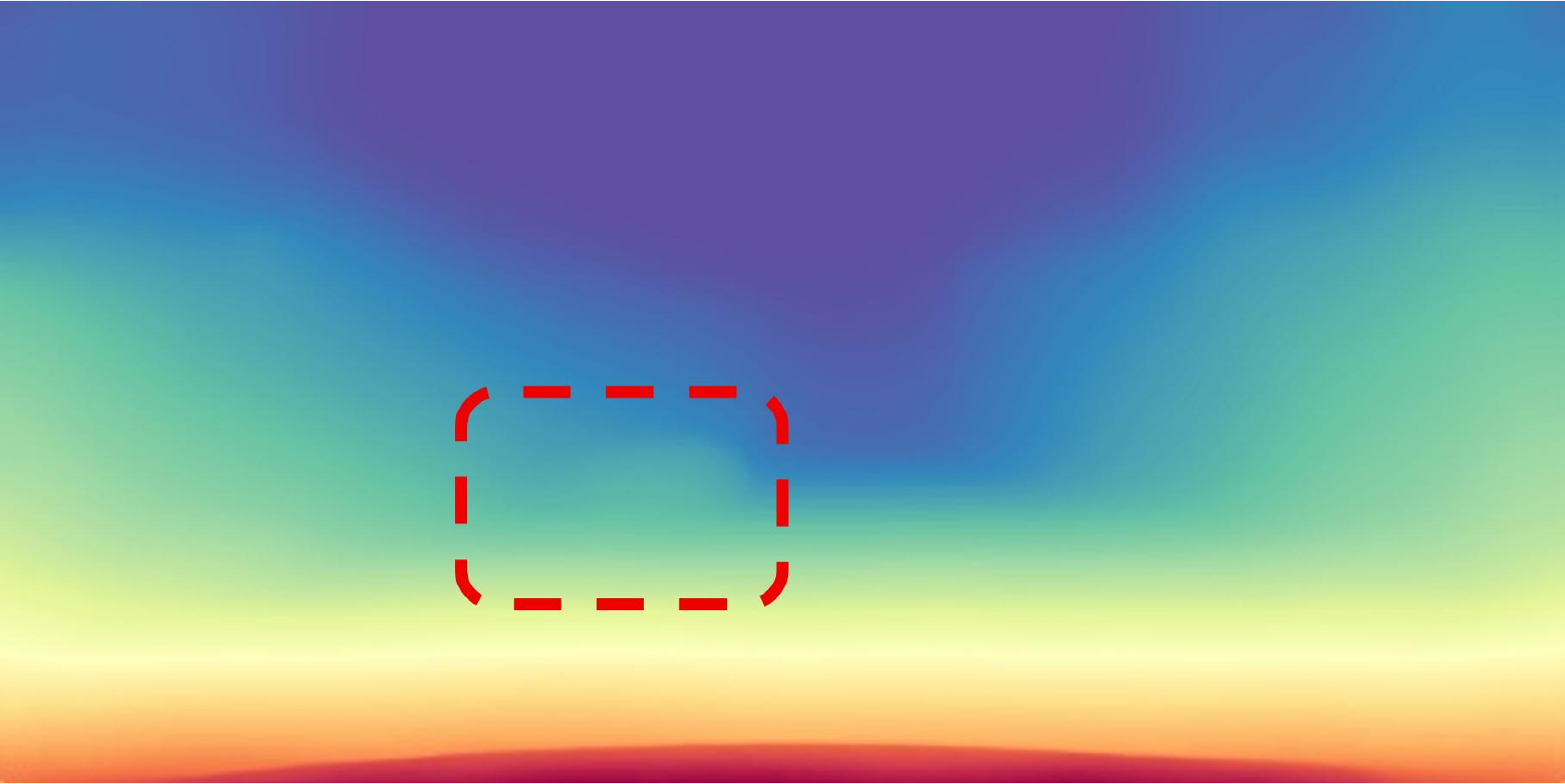} &
    \includegraphics[width=\turnheightnew]{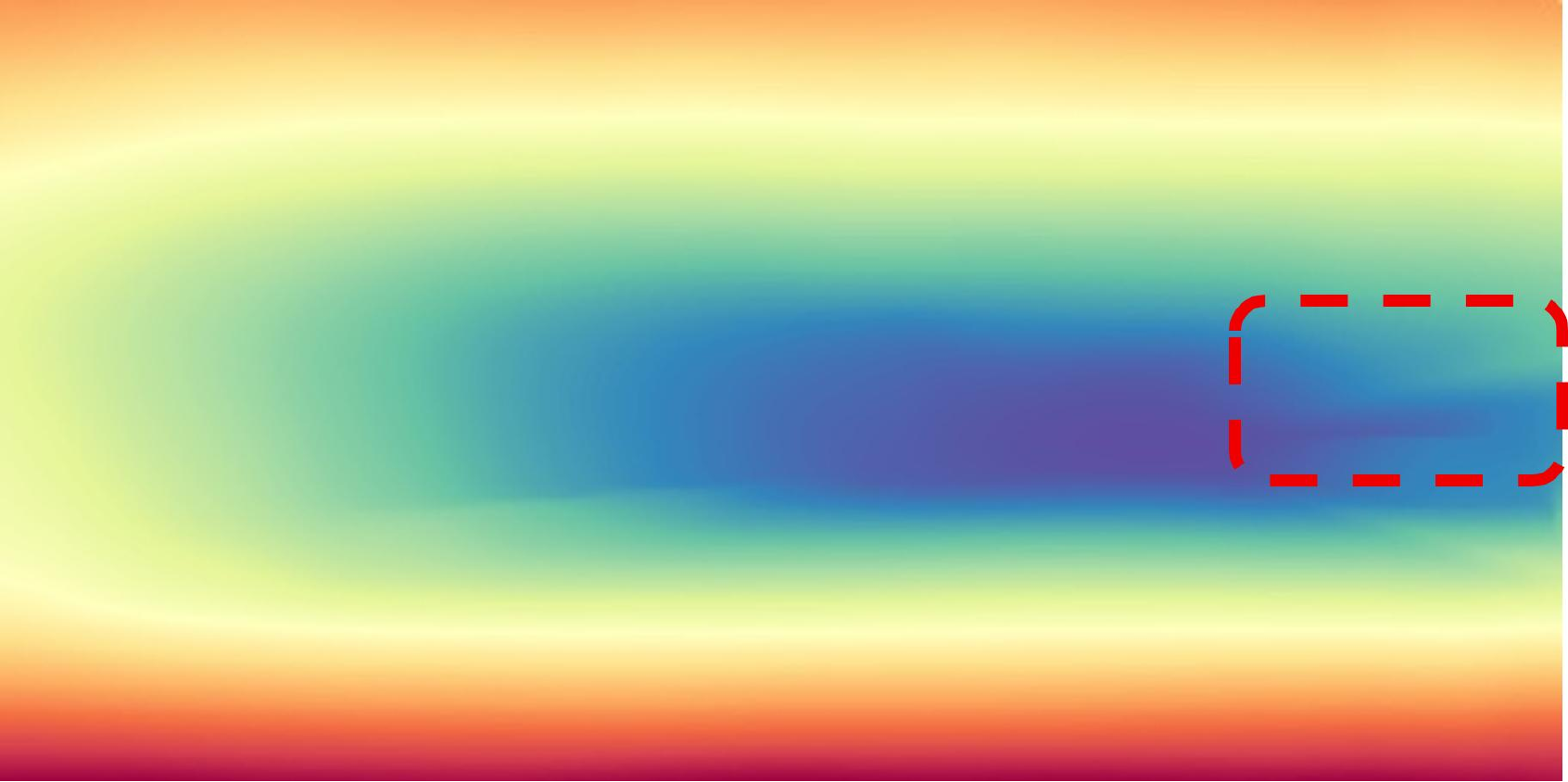} &
    \includegraphics[width=\turnheightnew]{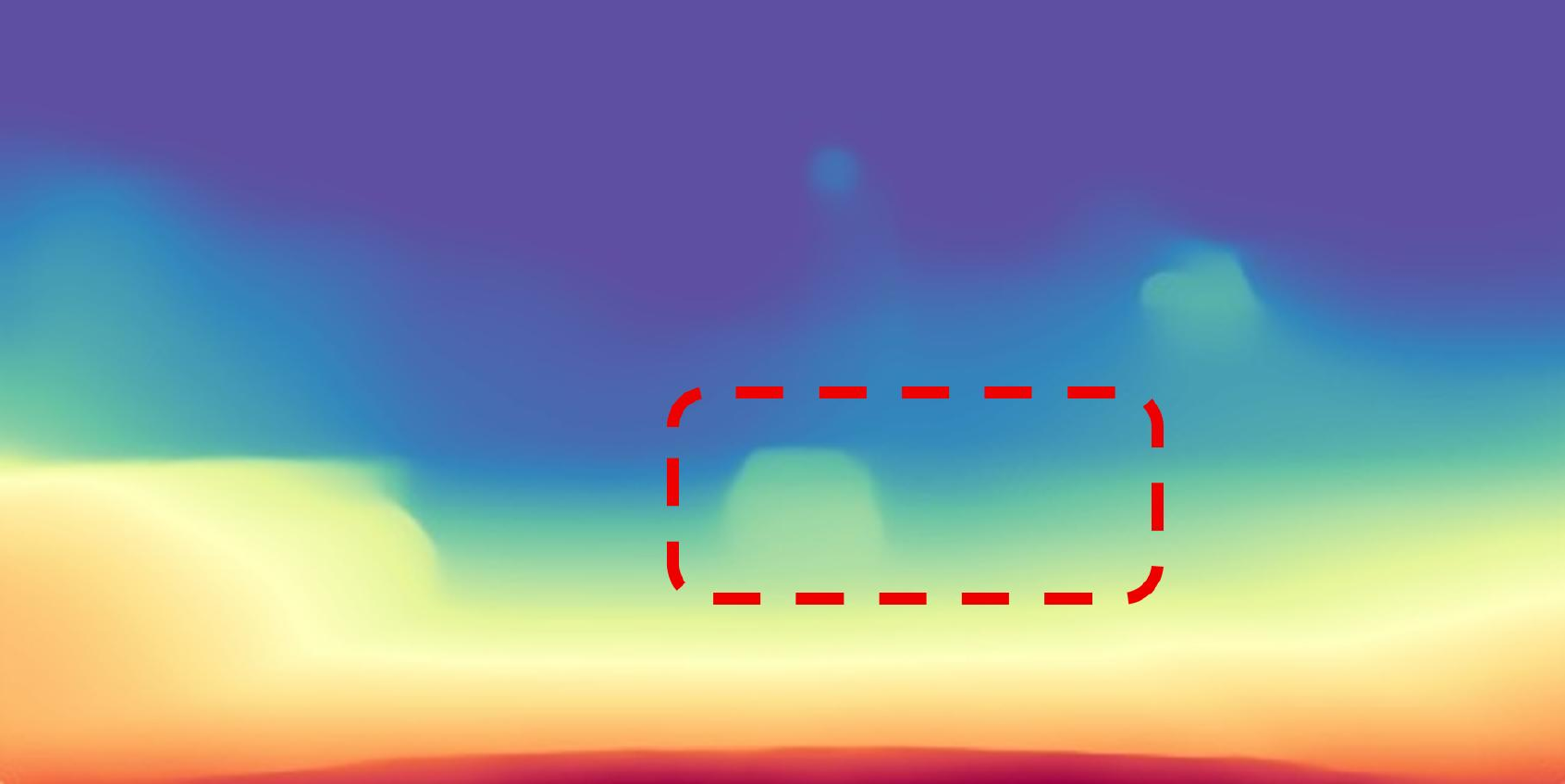} &
    
    \includegraphics[width=\turnheightnew]{img/Fig_visible/nuscenes-robot-night_pdf/nuscen_night_3_good/69_14.pdf} &
    \includegraphics[width=\turnheightnew]{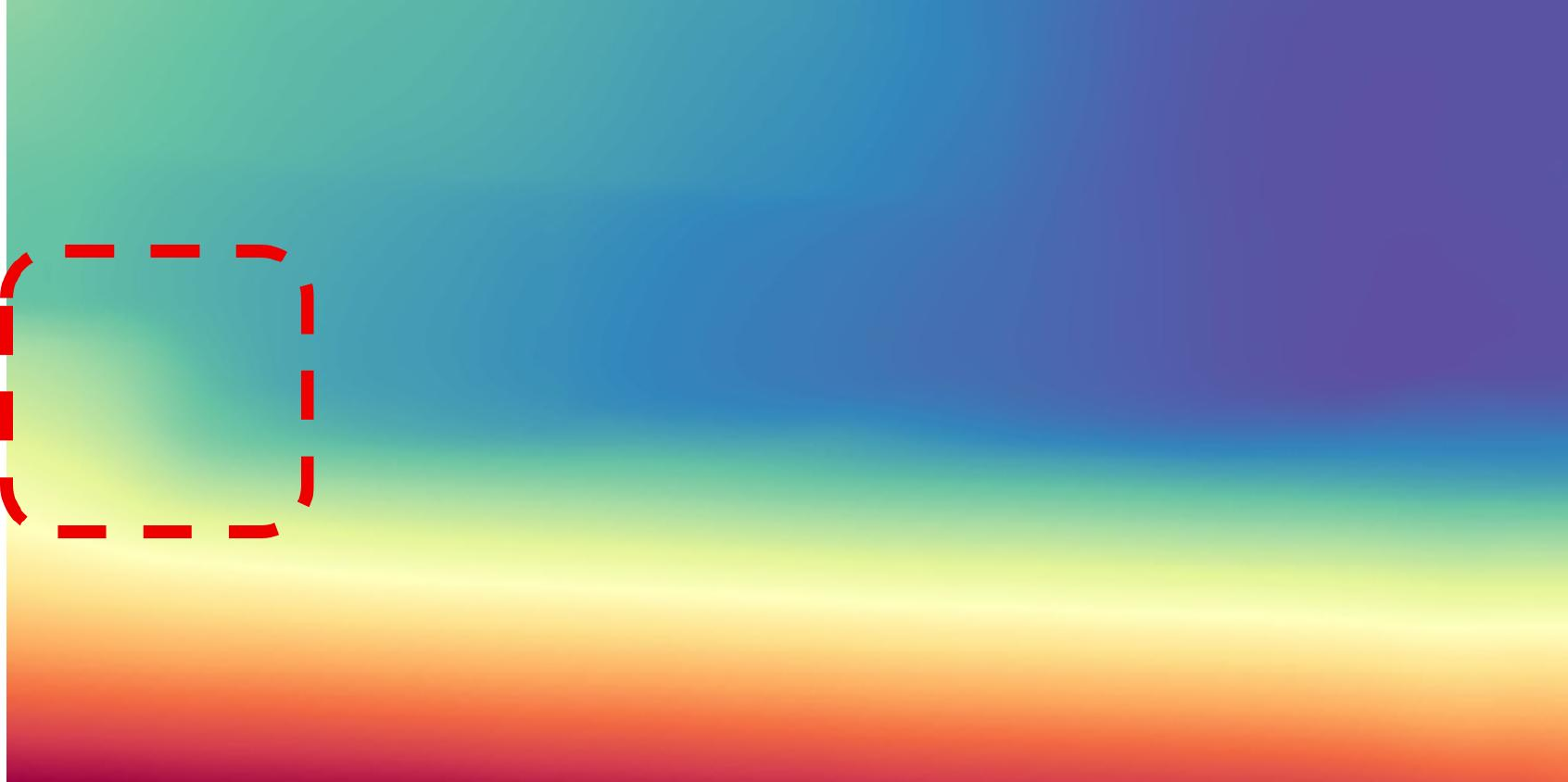} &
    \includegraphics[width=\turnheightnew]{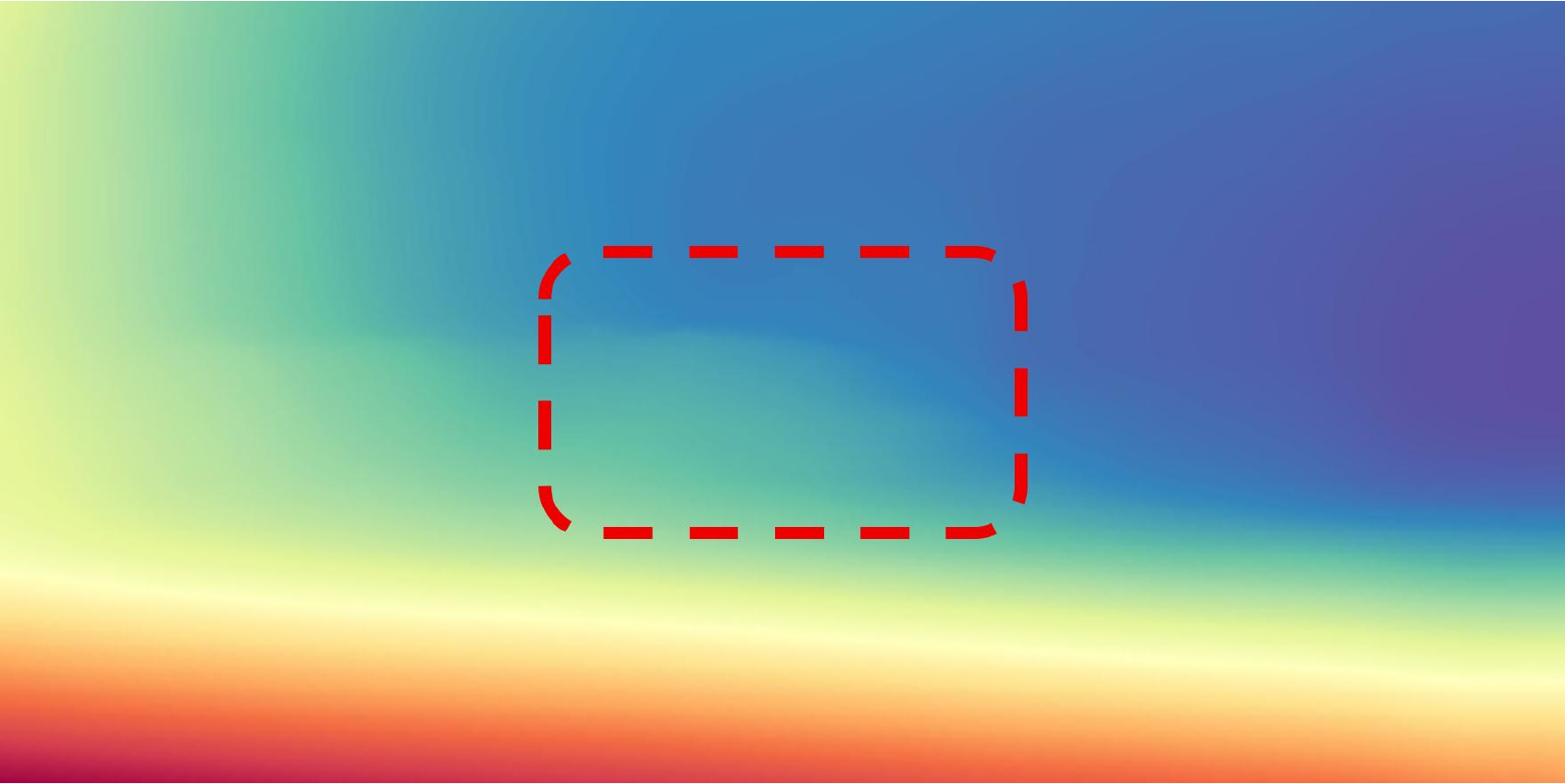} 
    \\

    \rotatebox{90}{\hspace{1mm}\fontsize{5pt}{5pt}\selectfont{Ours DepthDark}}&
    
    \includegraphics[width=\turnheightnew]{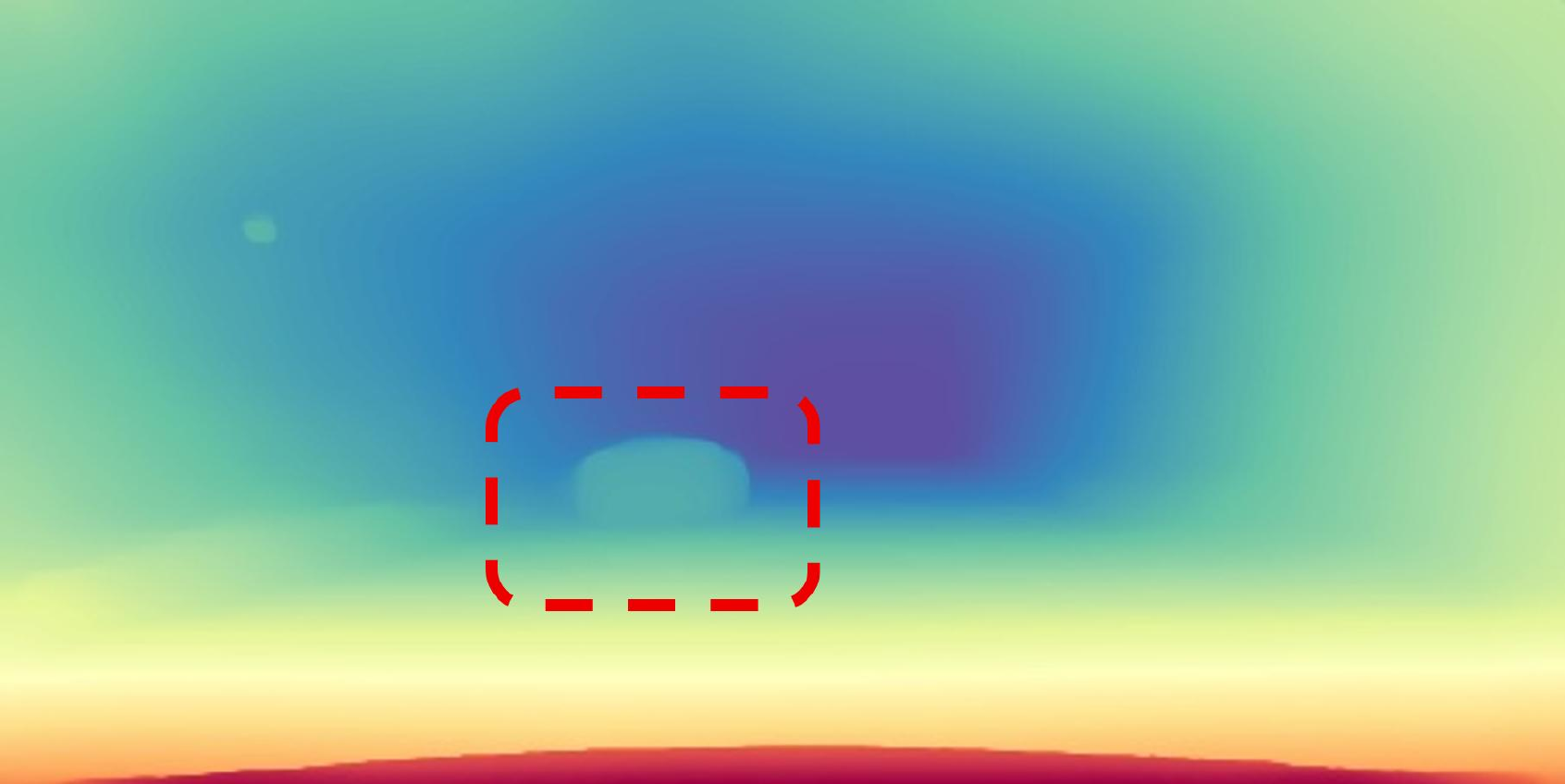} &
    \includegraphics[width=\turnheightnew]{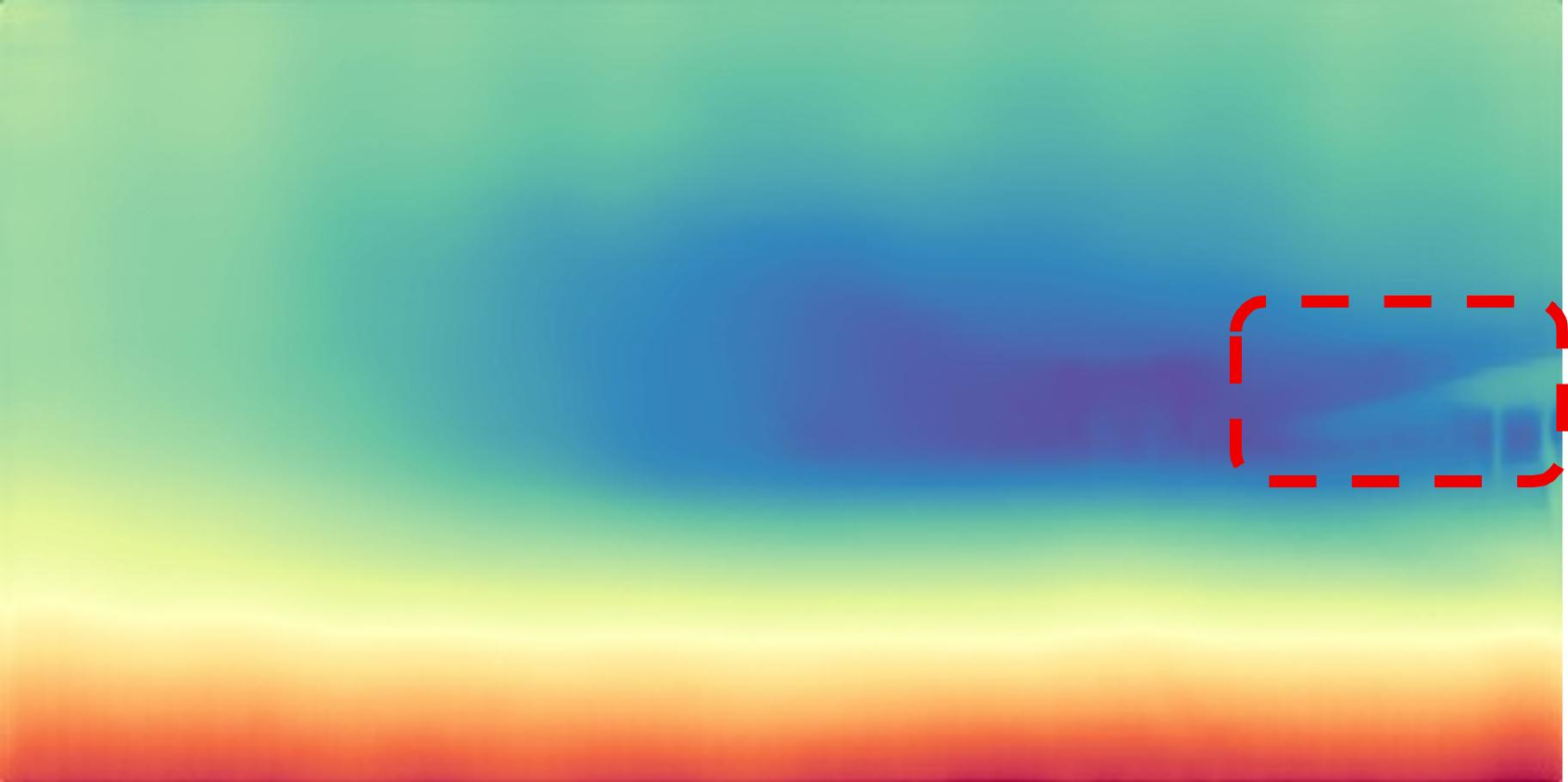} &
    \includegraphics[width=\turnheightnew]{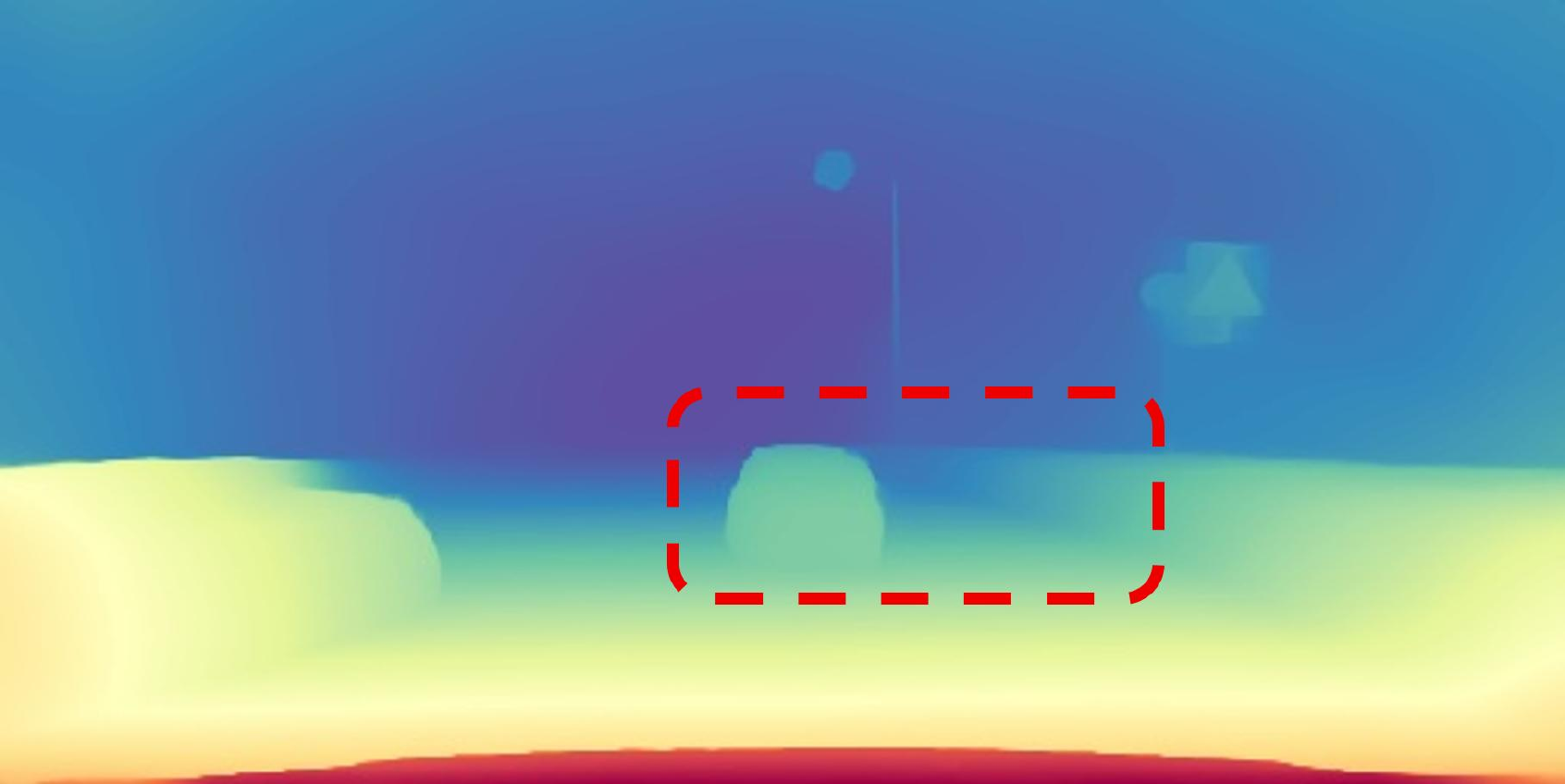} &
    
    \includegraphics[width=\turnheightnew]{img/Fig_visible/nuscenes-robot-night_pdf/nuscen_night_3_good/69_15.pdf} &
    \includegraphics[width=\turnheightnew]{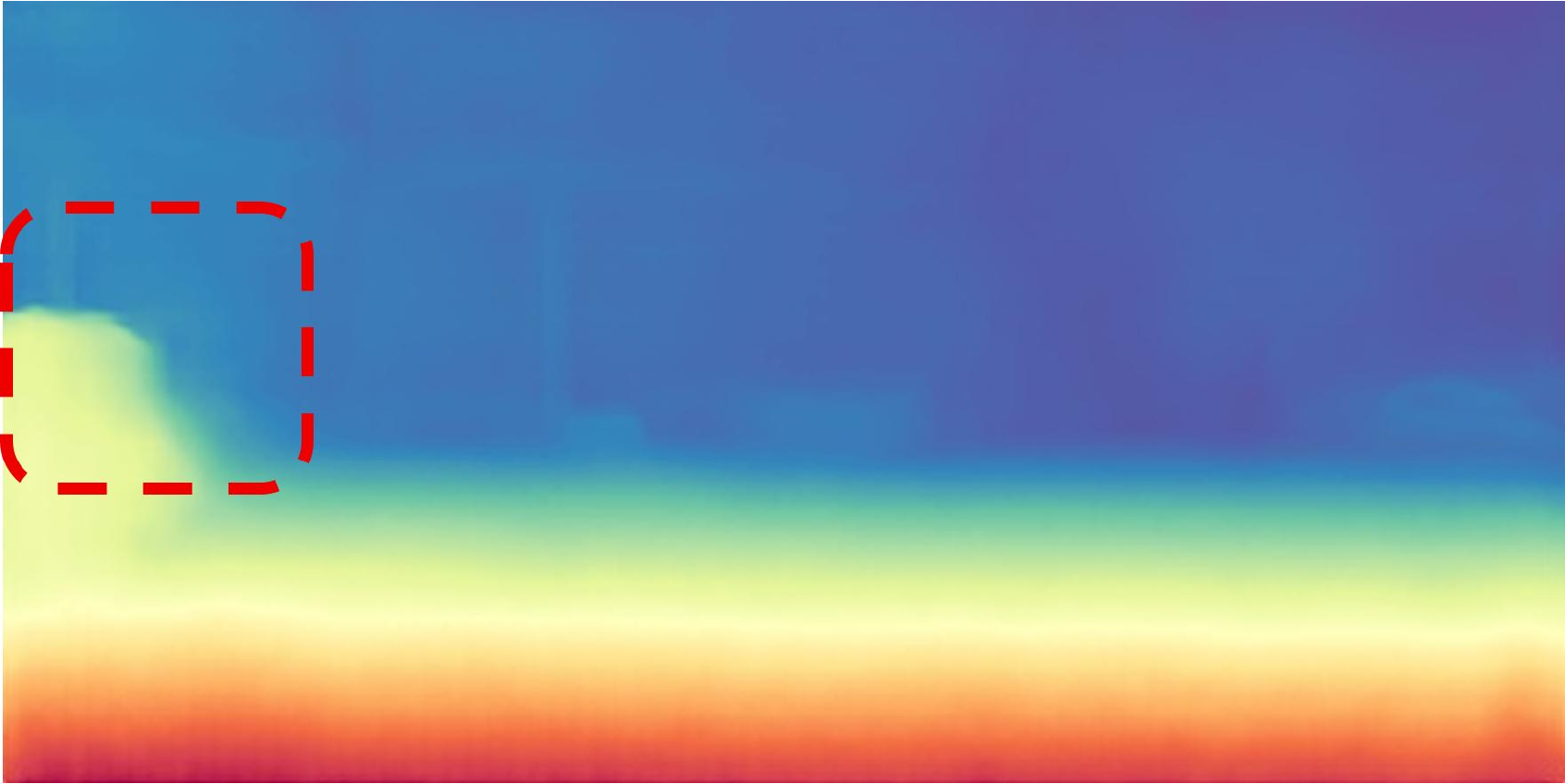} &
    \includegraphics[width=\turnheightnew]{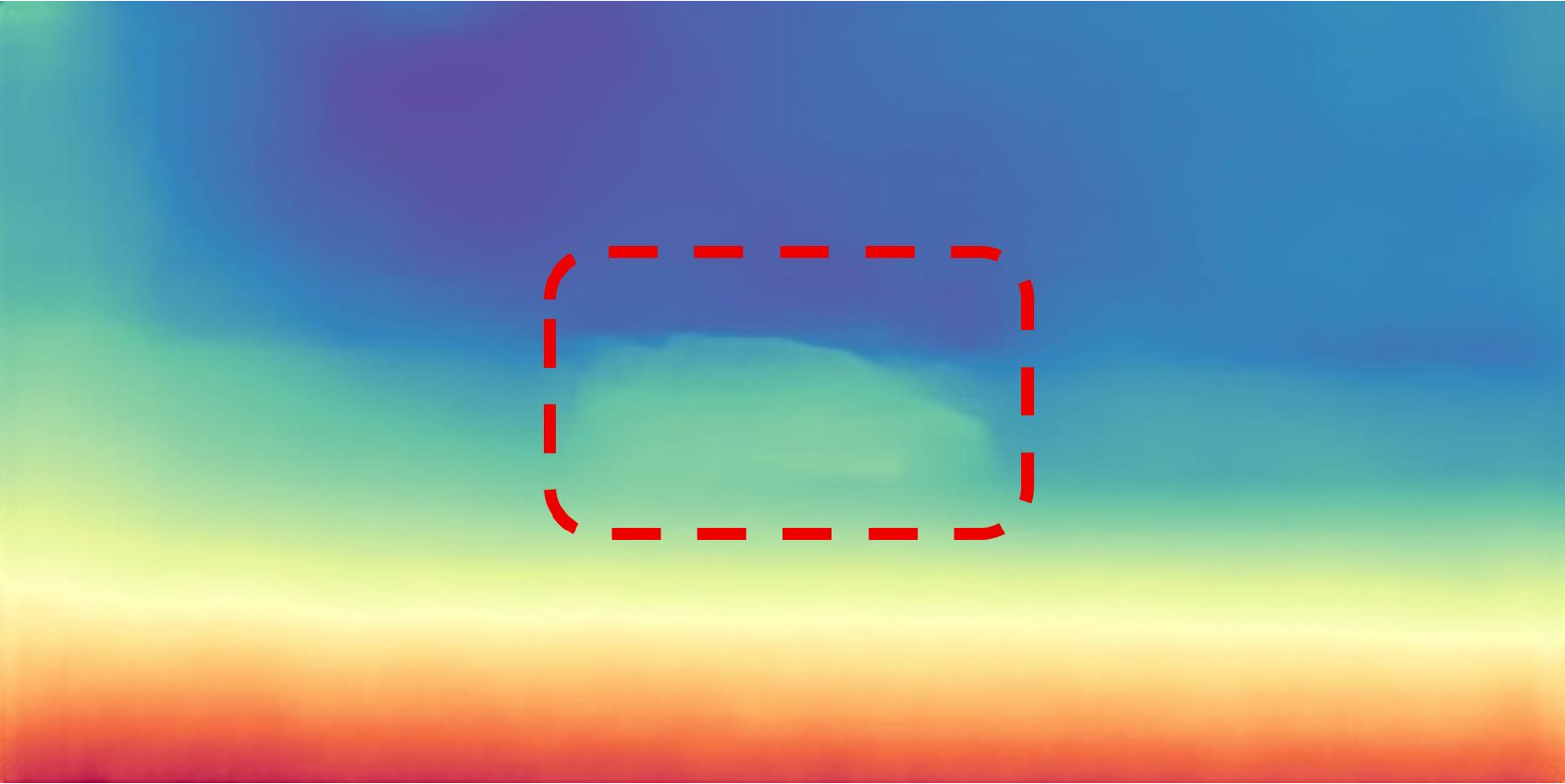}
    \\
    \end{tabular}
    \vspace{-2mm}

    \caption{Qualitative comparison results of different monocular depth estimation methods on the nuScenes-Night dataset. For conciseness, we present a visual comparison of four state-of-the-art methods with the most competitive performance, where the red dashed boxes distinctly indicate the regions where our approach demonstrates significant advantages.
    \vspace{-1mm}
    } 
    \label{fig:visual_nu}
   \vspace{-2mm}
\end{figure*}

\begin{figure*}[ht]
\newpage
    \centering
    \newcommand{\turnheightnew}{0.33\columnwidth}
    \begin{tabular}{@{\hskip 0mm}c@{\hskip 1mm}c@{\hskip 1mm}c@{\hskip 1mm}c@{\hskip 1mm}c@{\hskip 1mm}@{\hskip 0mm}c@{\hskip 1mm}c@{\hskip 1mm}c@{\hskip 1mm}c@{\hskip 1mm}c@{\hskip 1mm}c@{}}
    {
 
    \rotatebox{90}{\hspace{3mm}\fontsize{5pt}{5pt}\selectfont{N-N \& R-N}}} &
    
    \includegraphics[width=\turnheightnew]{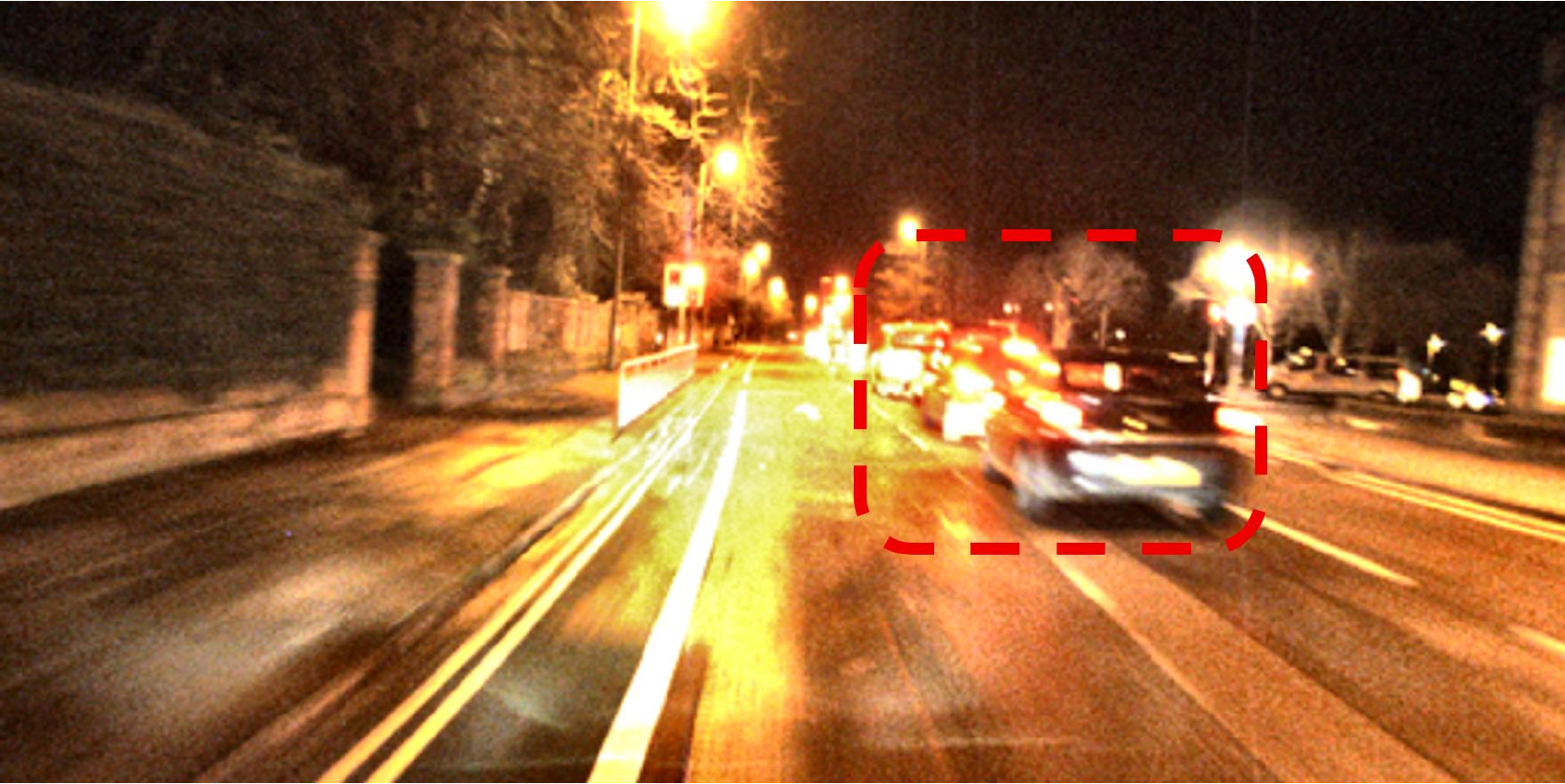} &
    \includegraphics[width=\turnheightnew]{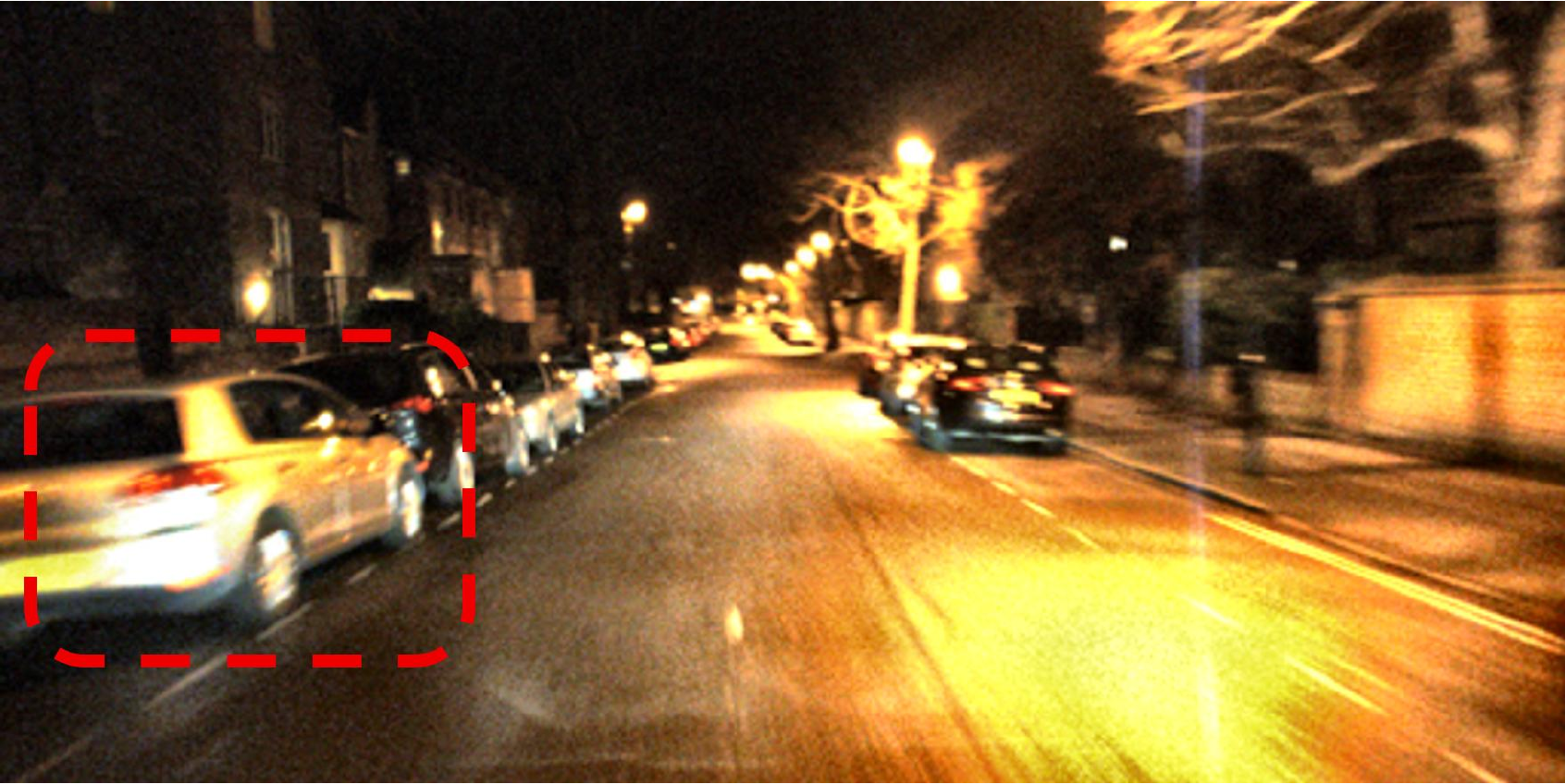} &
    \includegraphics[width=\turnheightnew]{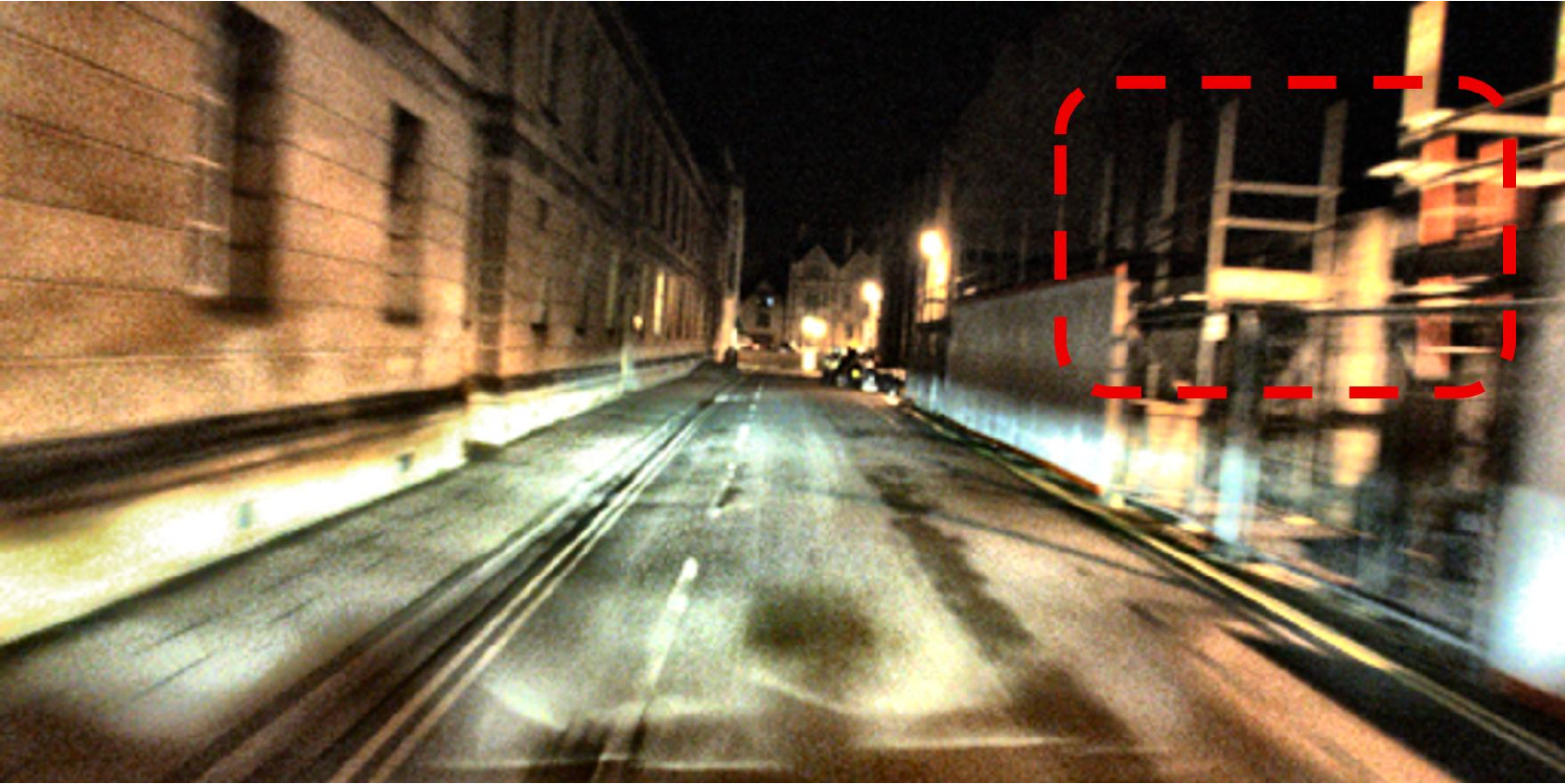} &

    \includegraphics[width=\turnheightnew]{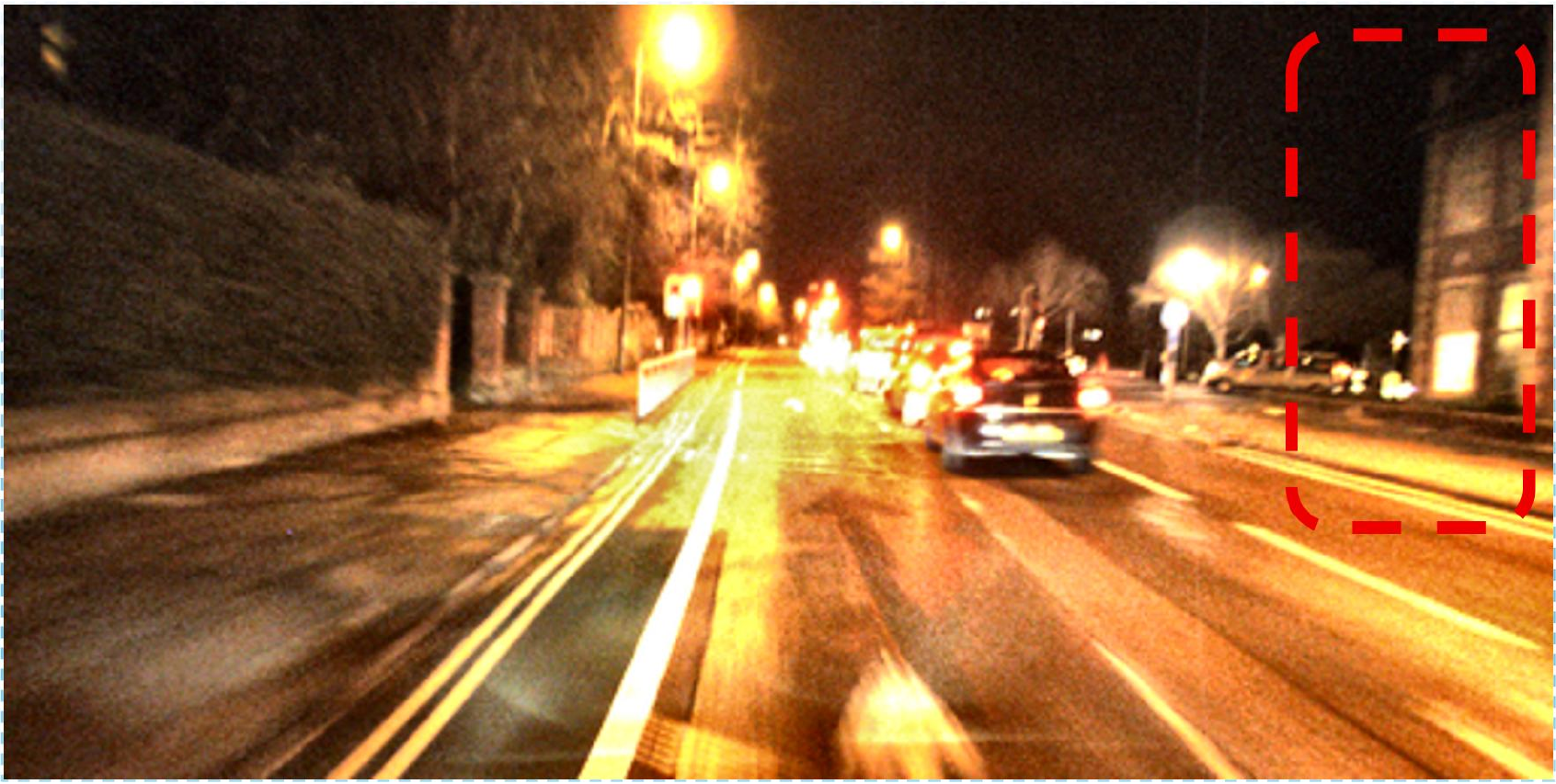} &
    \includegraphics[width=\turnheightnew]{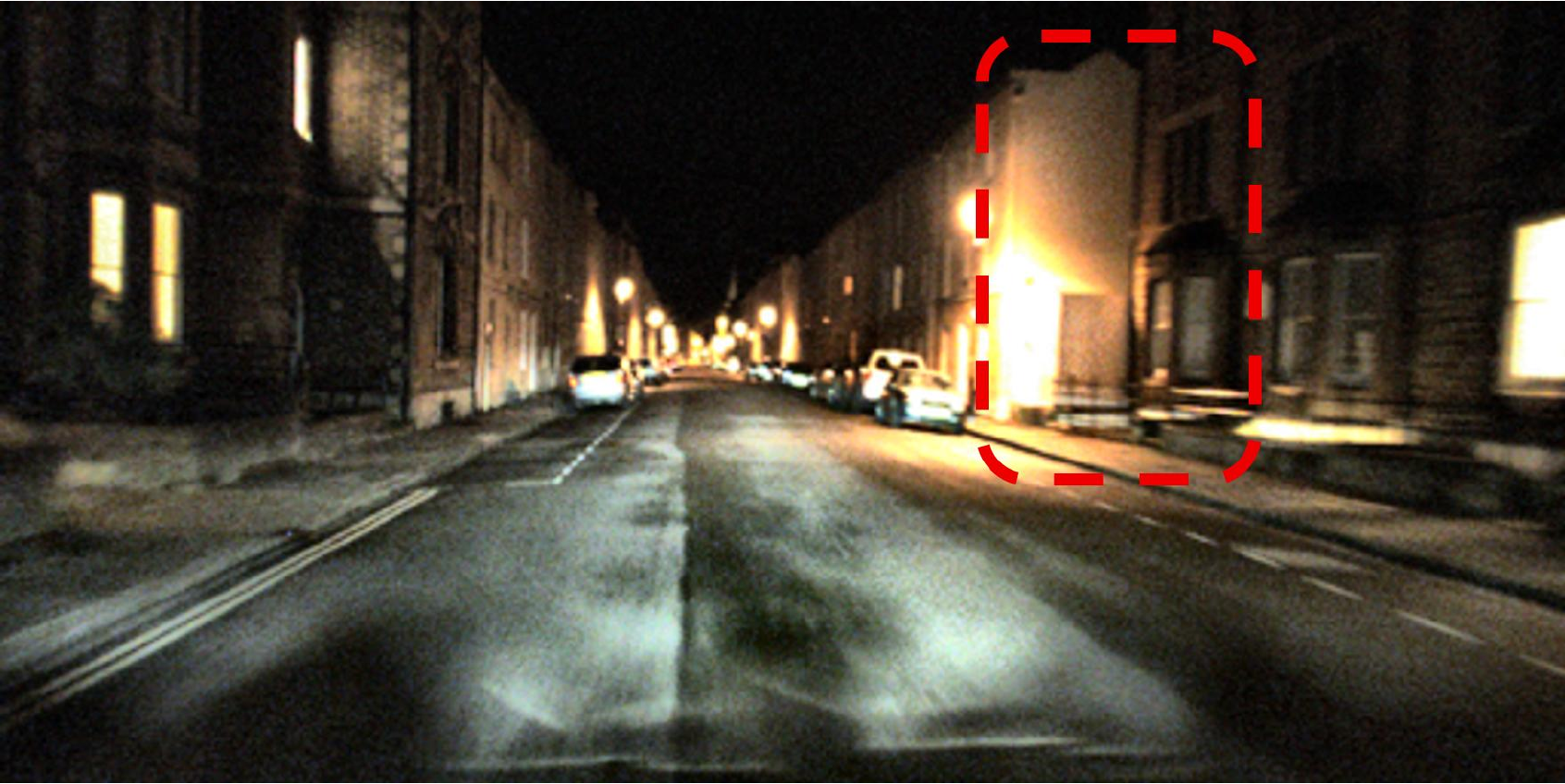} &
    \includegraphics[width=\turnheightnew]{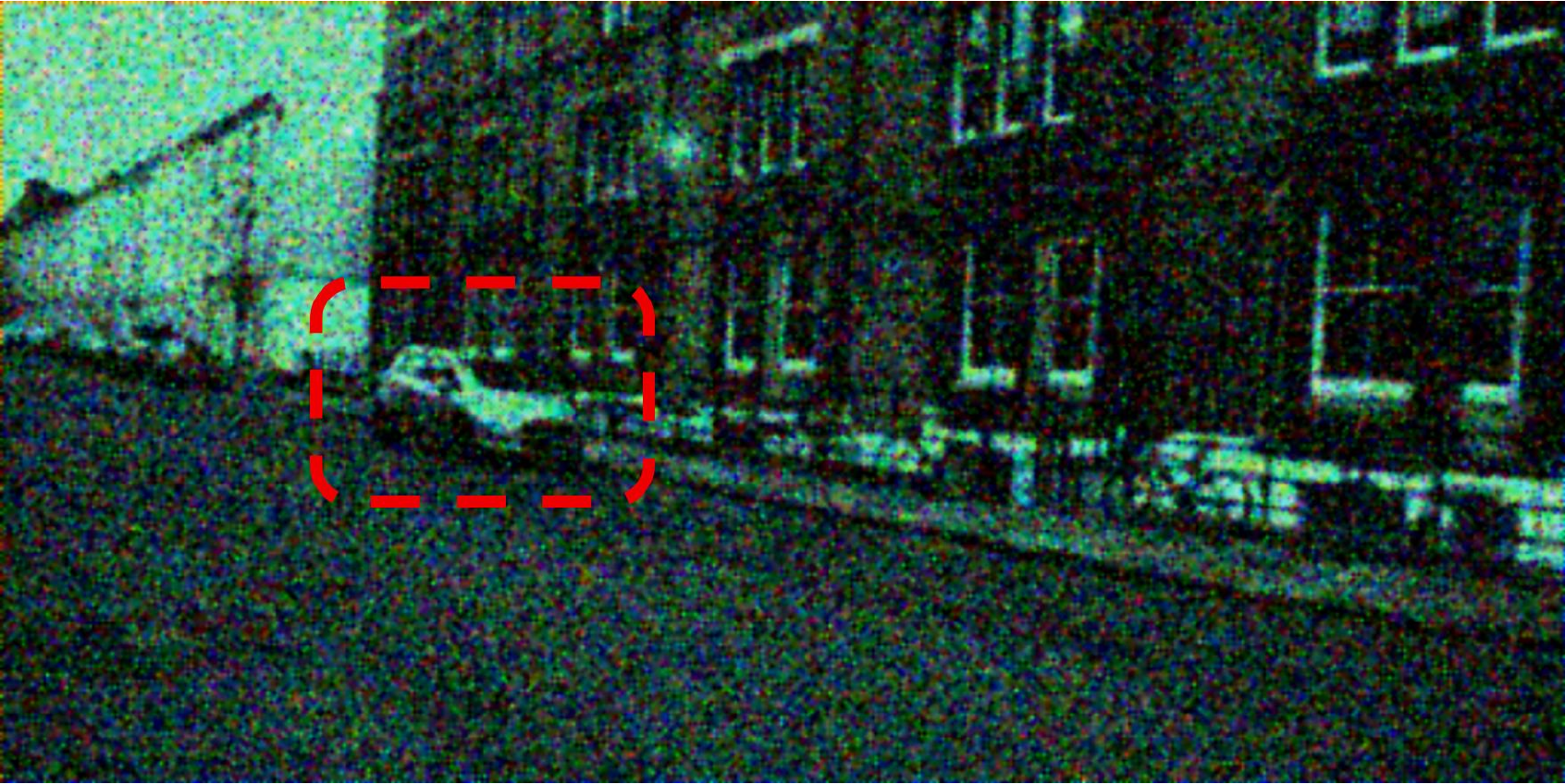} 
    \\
 
    \rotatebox{90}{\hspace{5mm}\fontsize{5pt}{5pt}\selectfont{ADDS}
    }&
    \includegraphics[width=\turnheightnew]{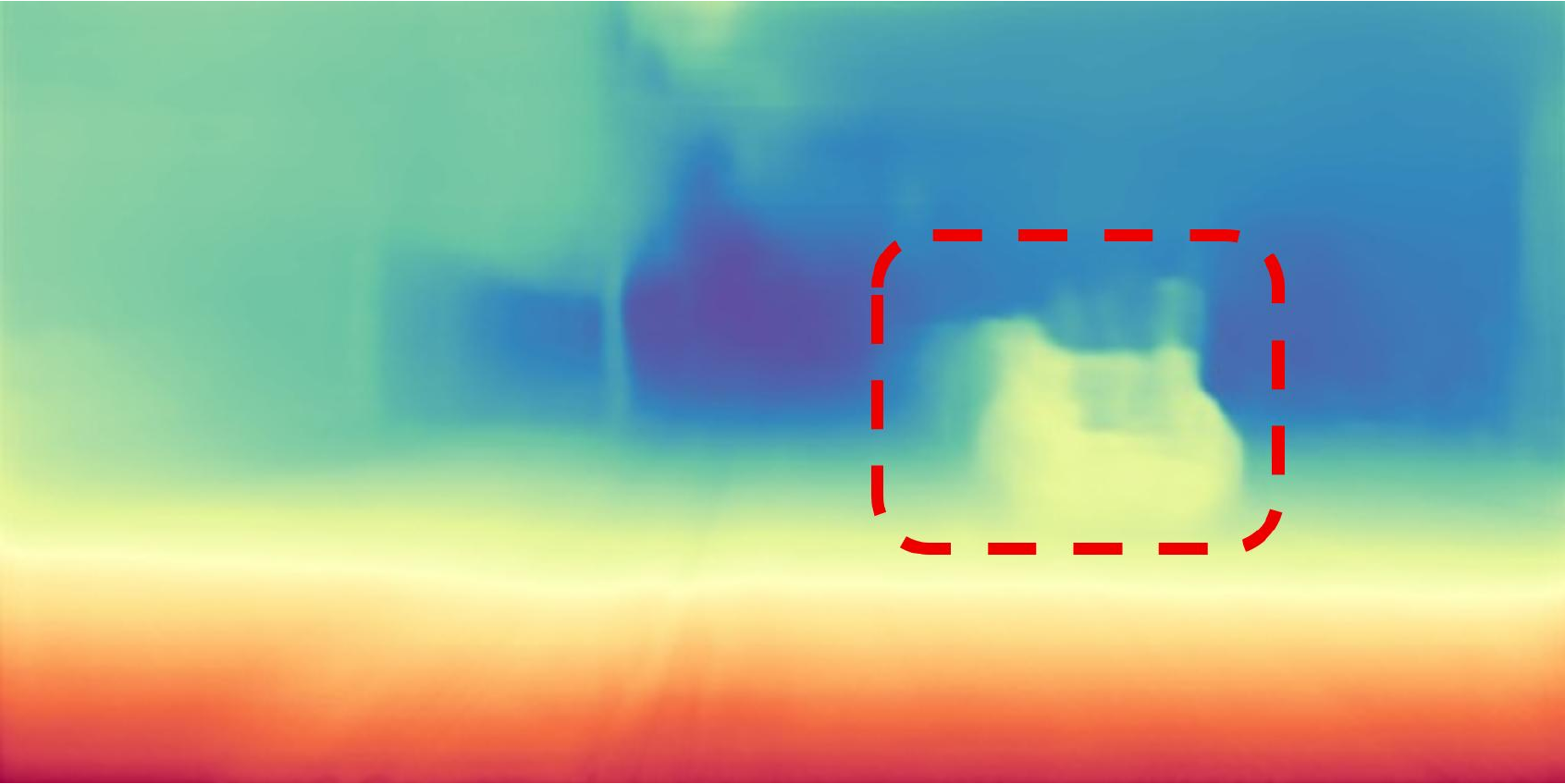} &
    \includegraphics[width=\turnheightnew]{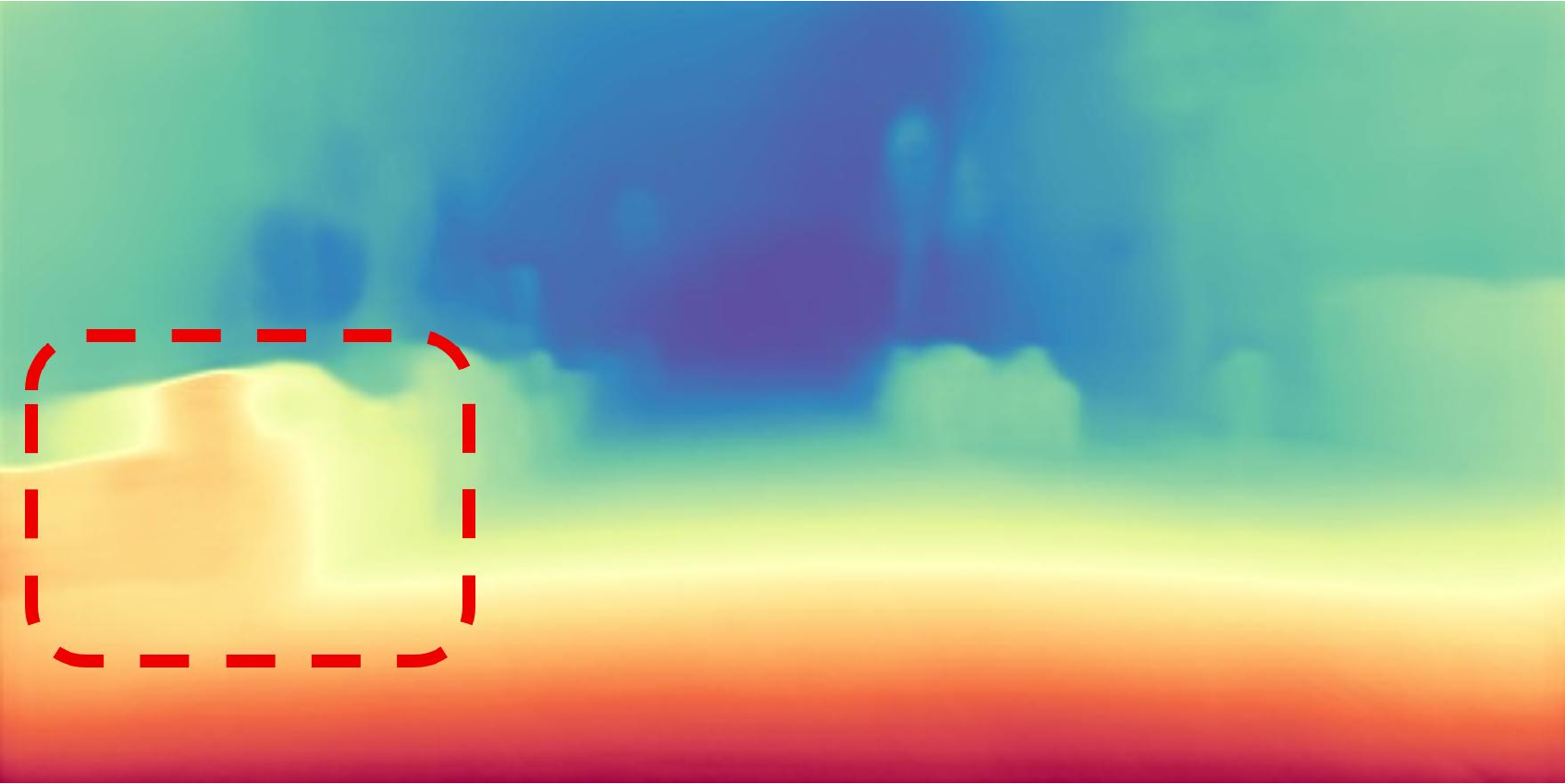} &
    \includegraphics[width=\turnheightnew]{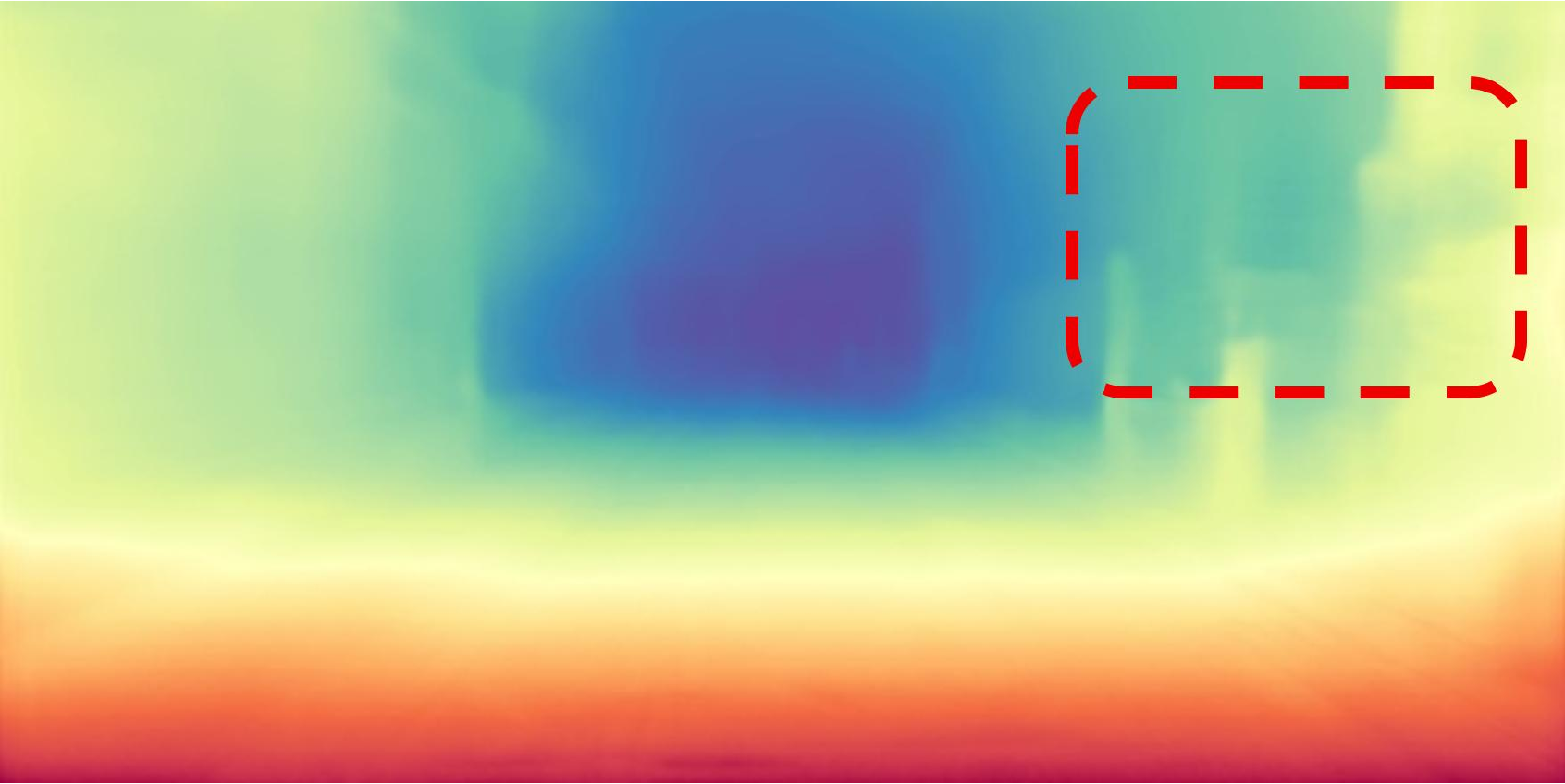} &

   \includegraphics[width=\turnheightnew]{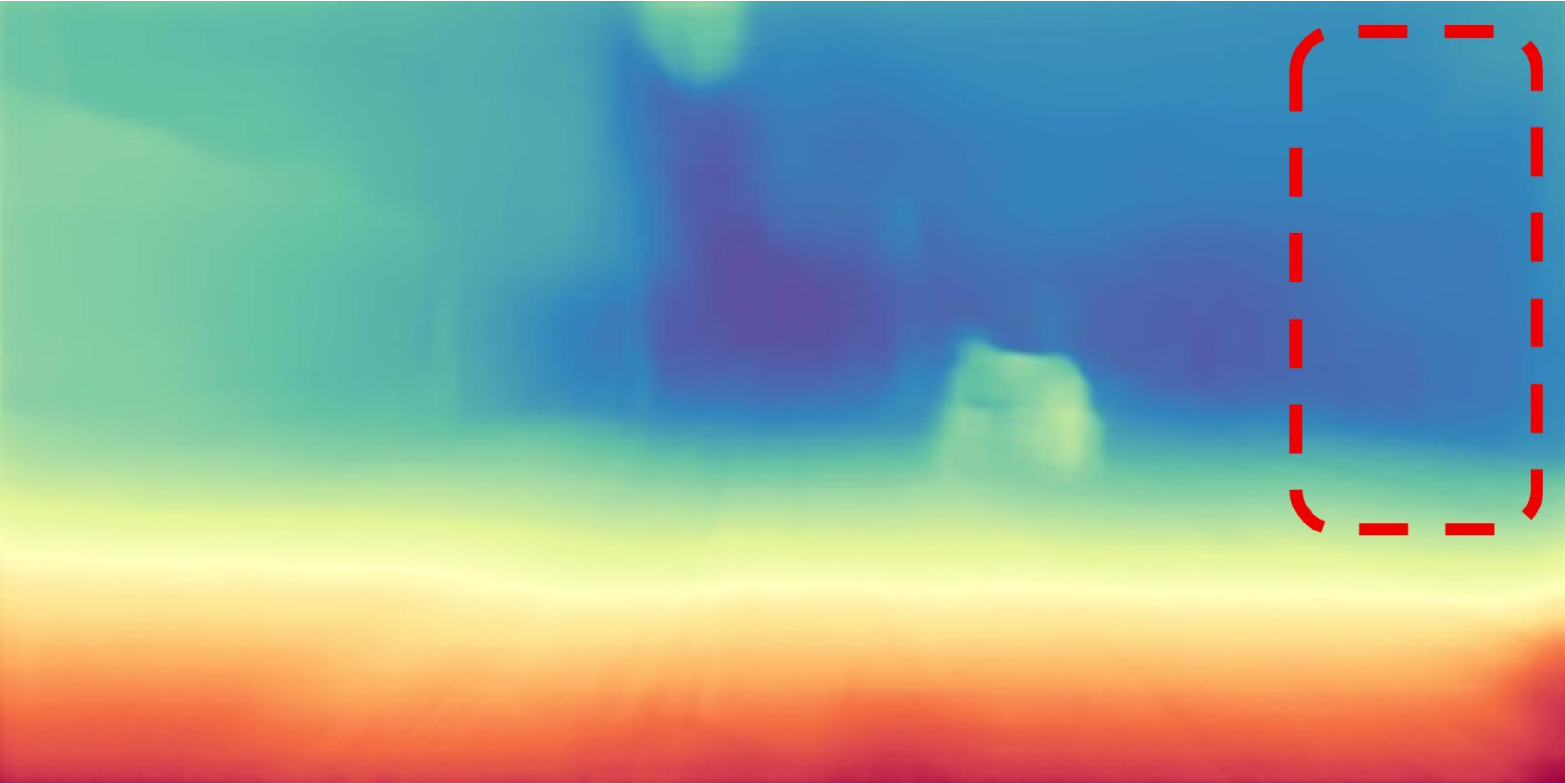} &
    \includegraphics[width=\turnheightnew]{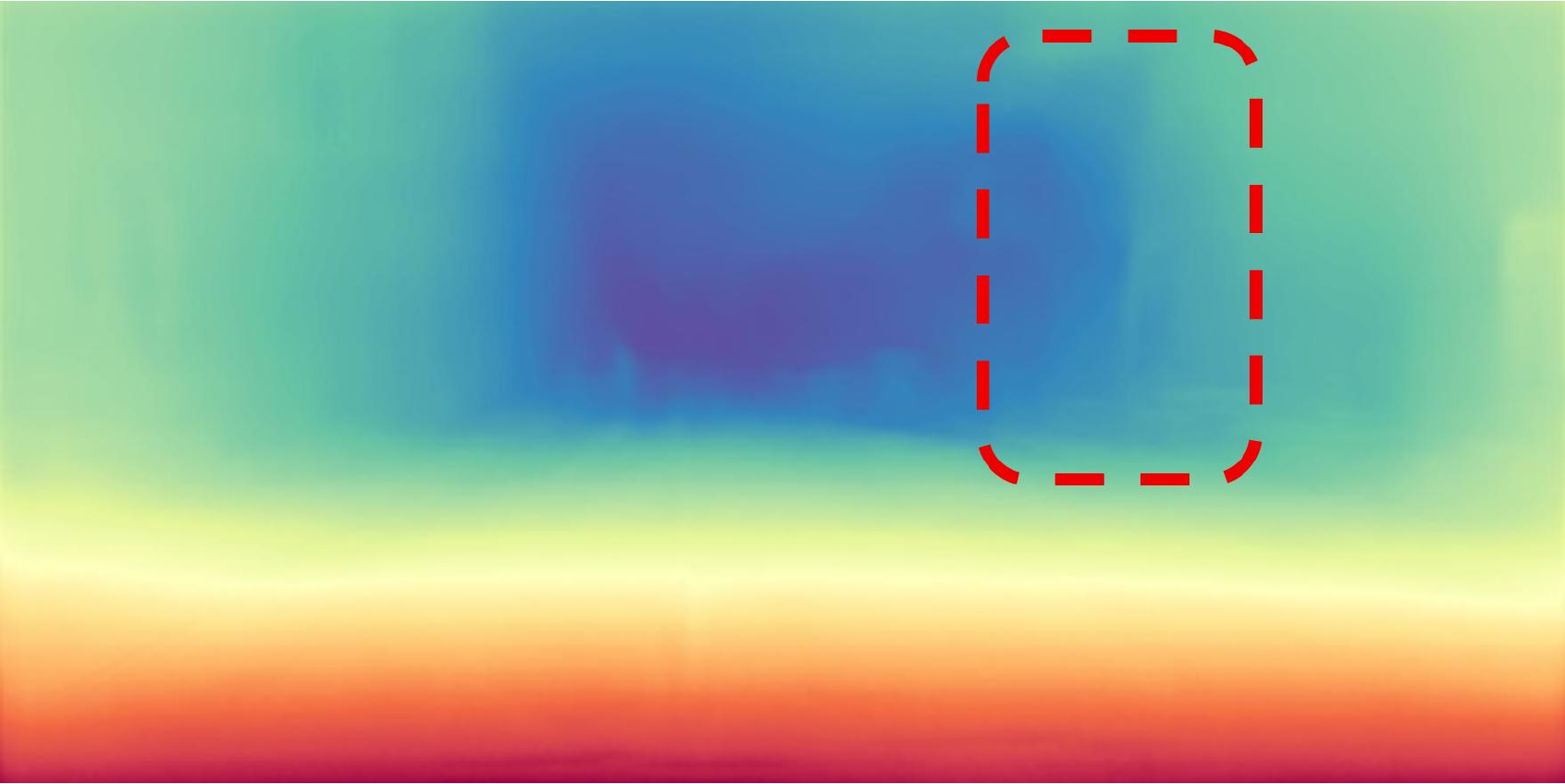} &
    \includegraphics[width=\turnheightnew]{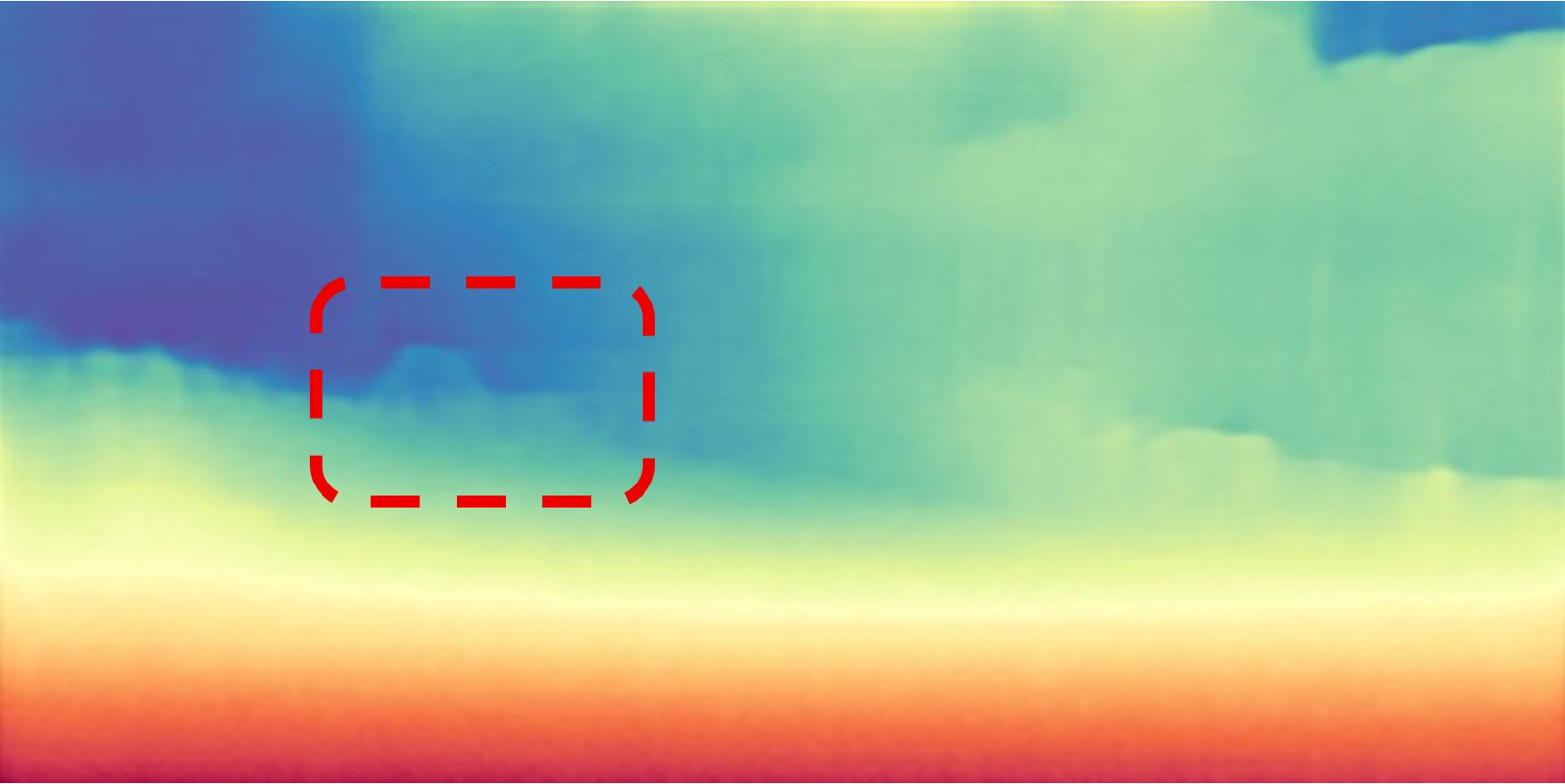}  
    \\
  
    \rotatebox{90}{\hspace{1mm}\fontsize{5pt}{5pt}\selectfont{Depth Anything}
    }&
    \includegraphics[width=\turnheightnew]{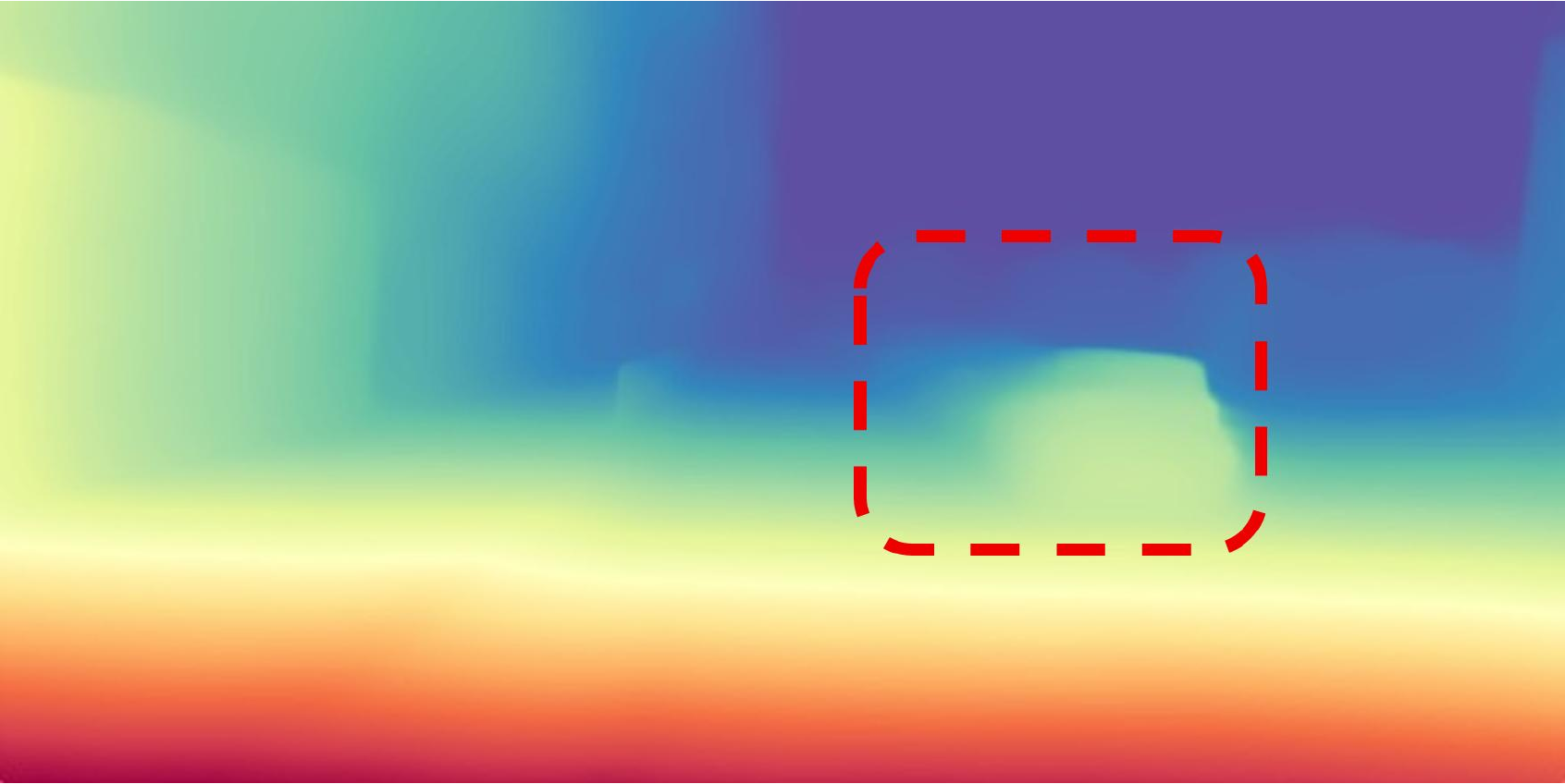} &
    \includegraphics[width=\turnheightnew]{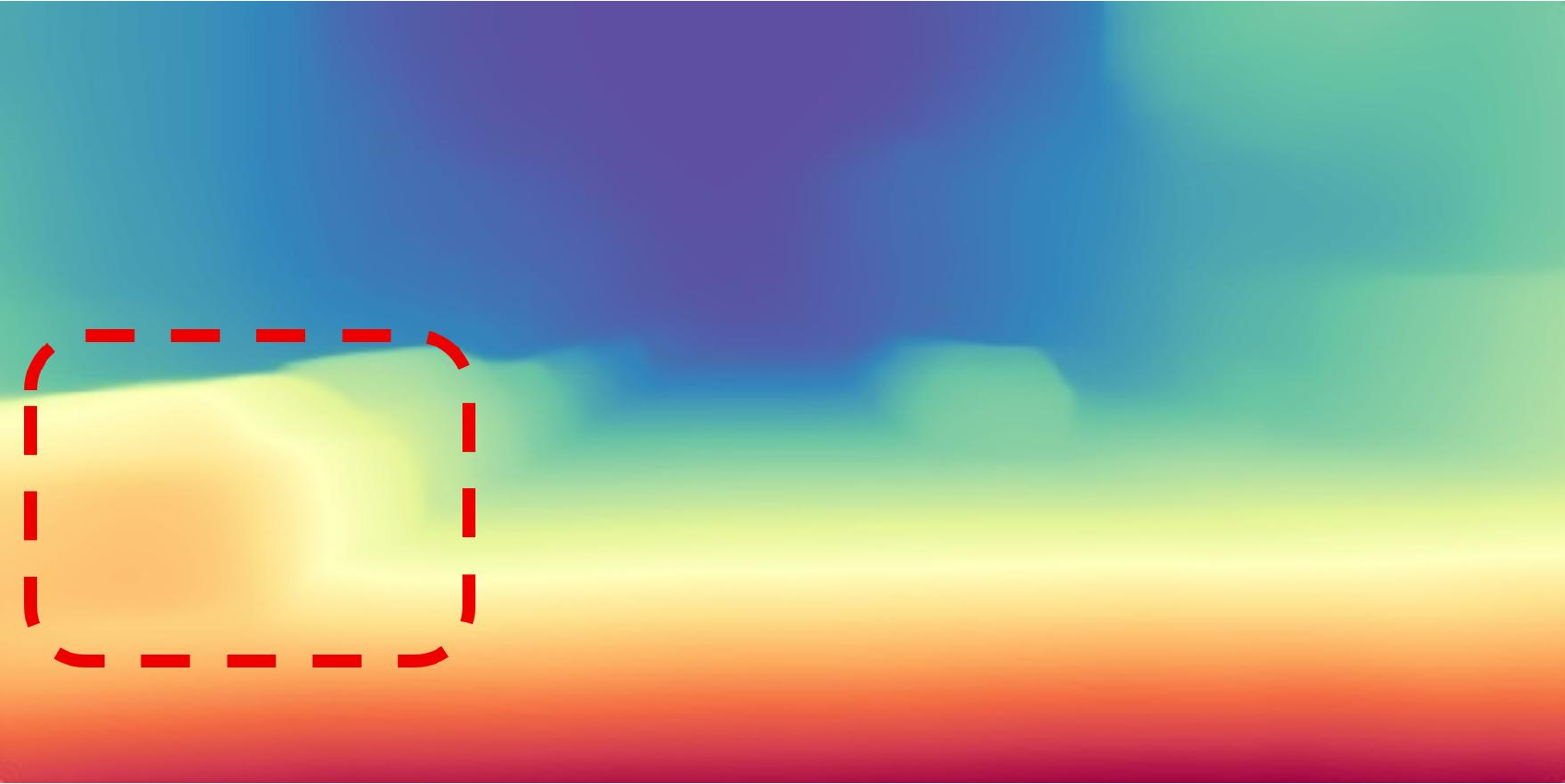} &
    \includegraphics[width=\turnheightnew]{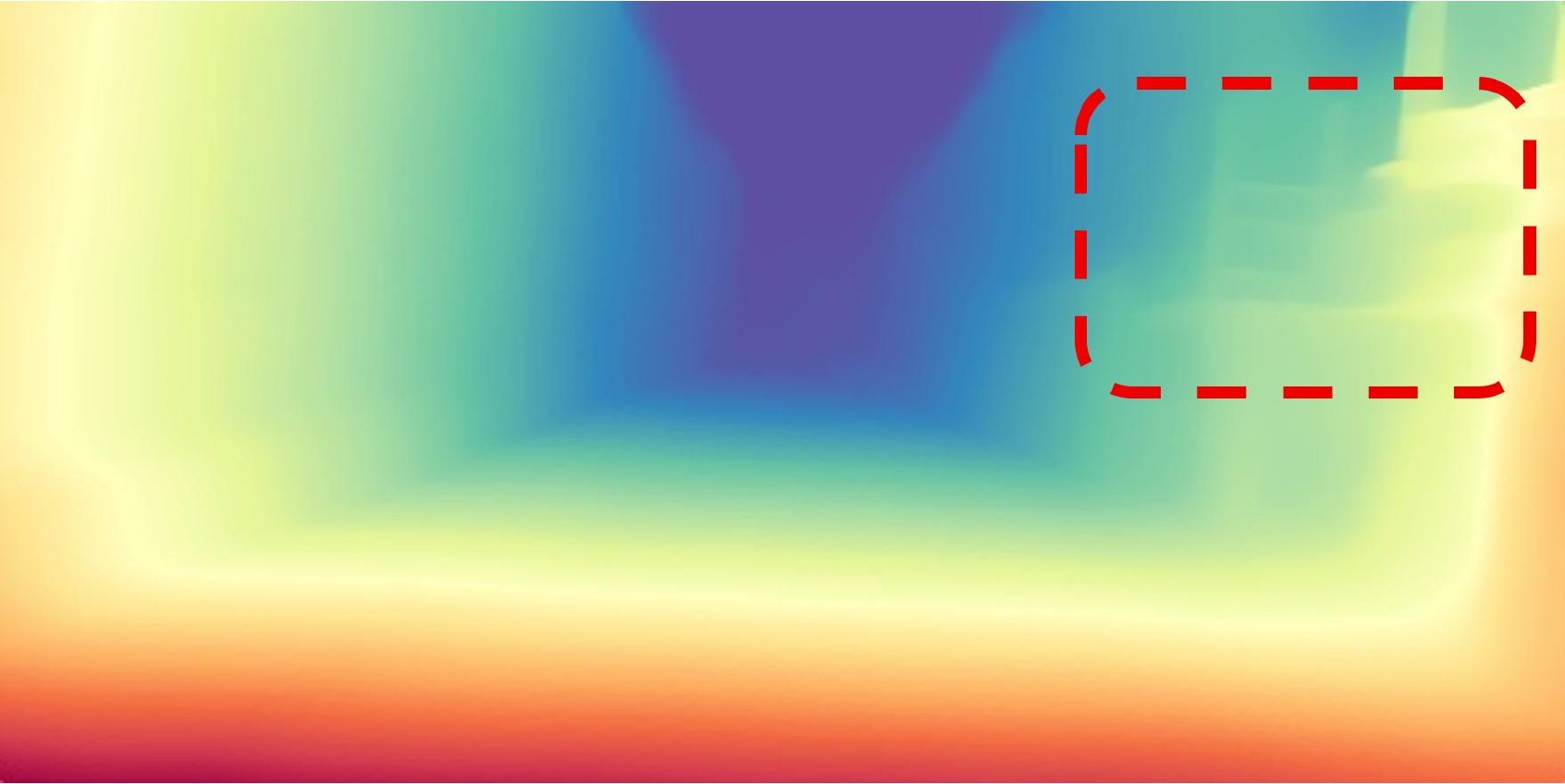} &

    \includegraphics[width=\turnheightnew]{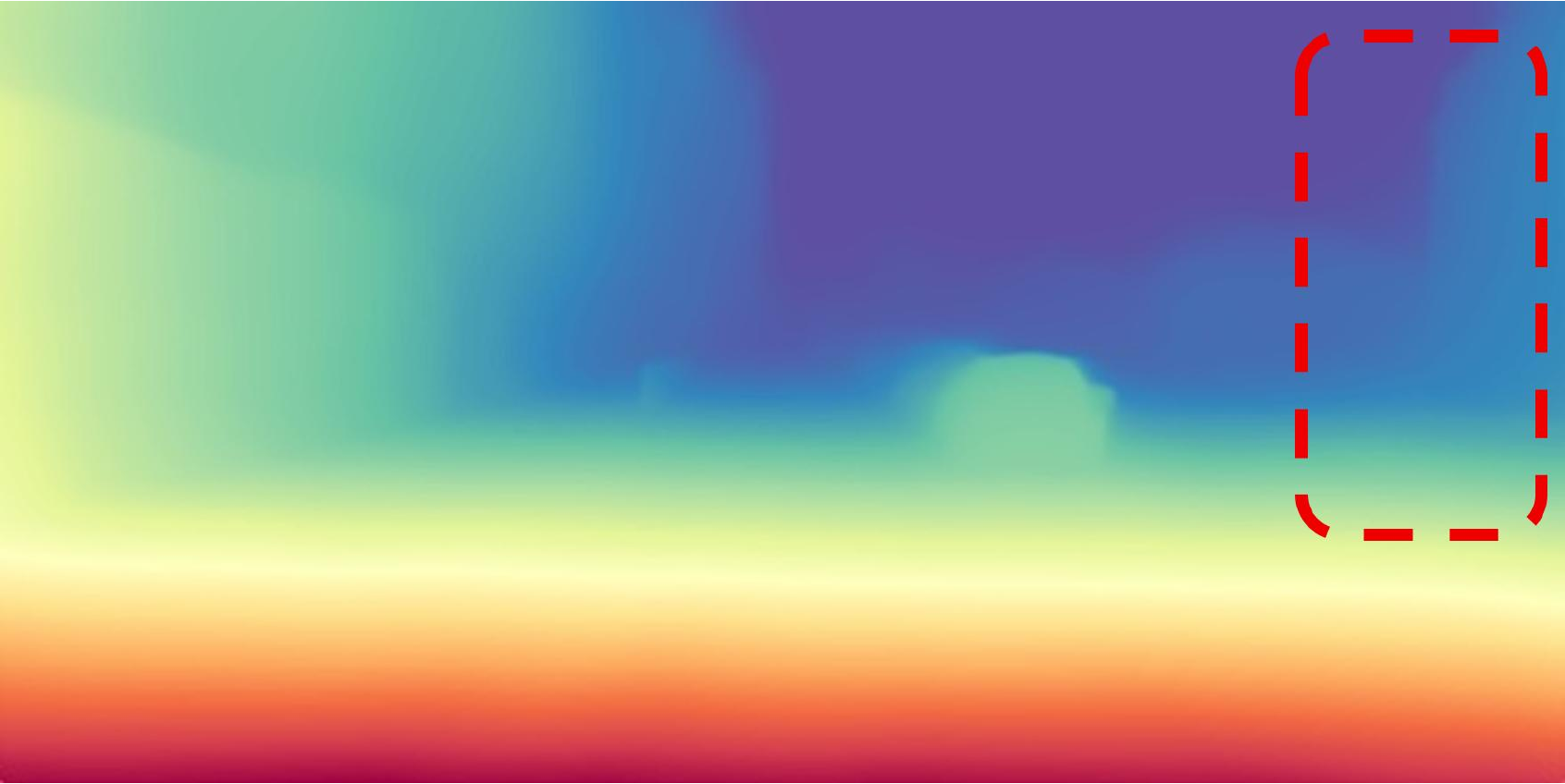} &
    \includegraphics[width=\turnheightnew]{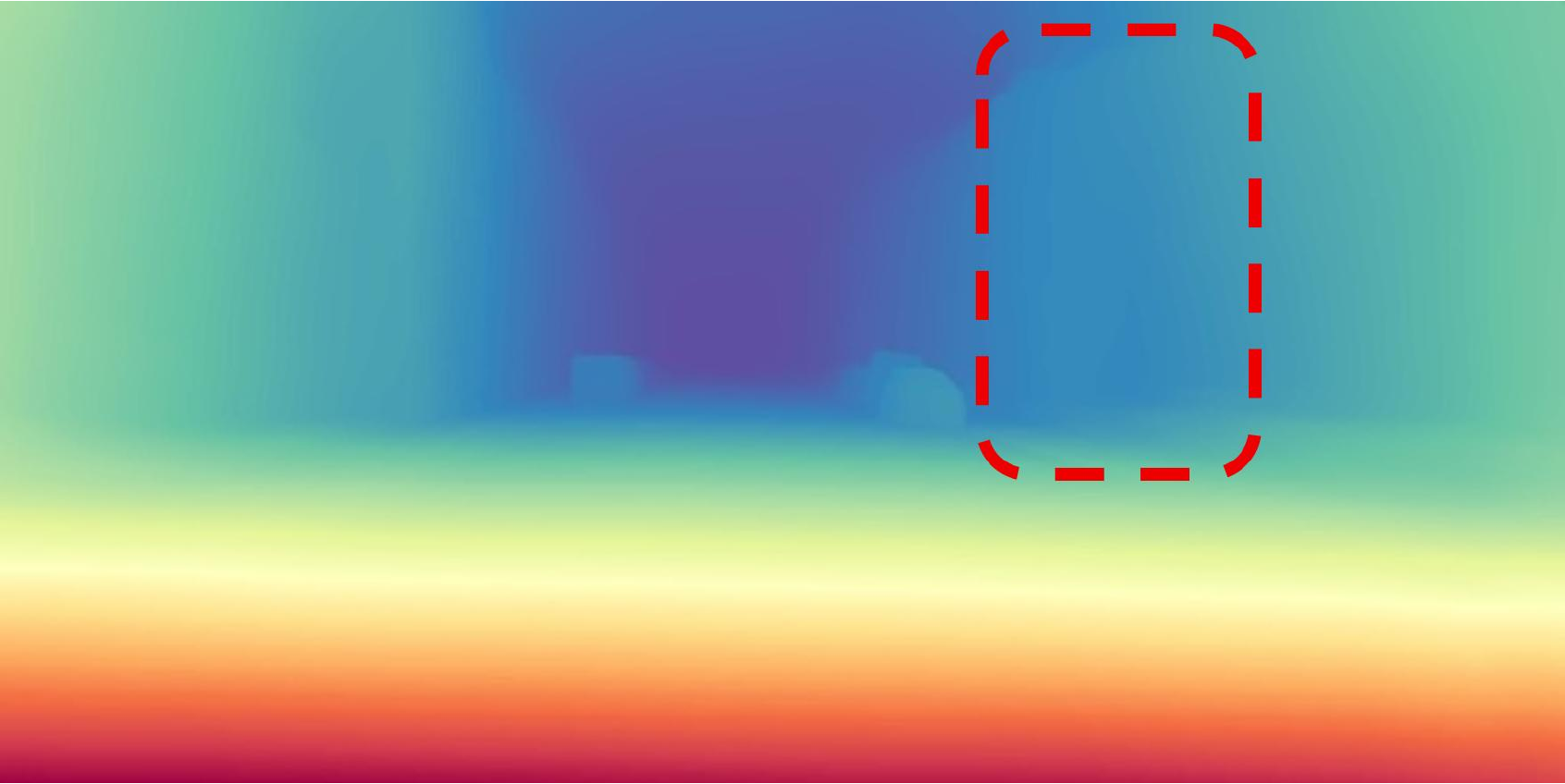} &
    \includegraphics[width=\turnheightnew]{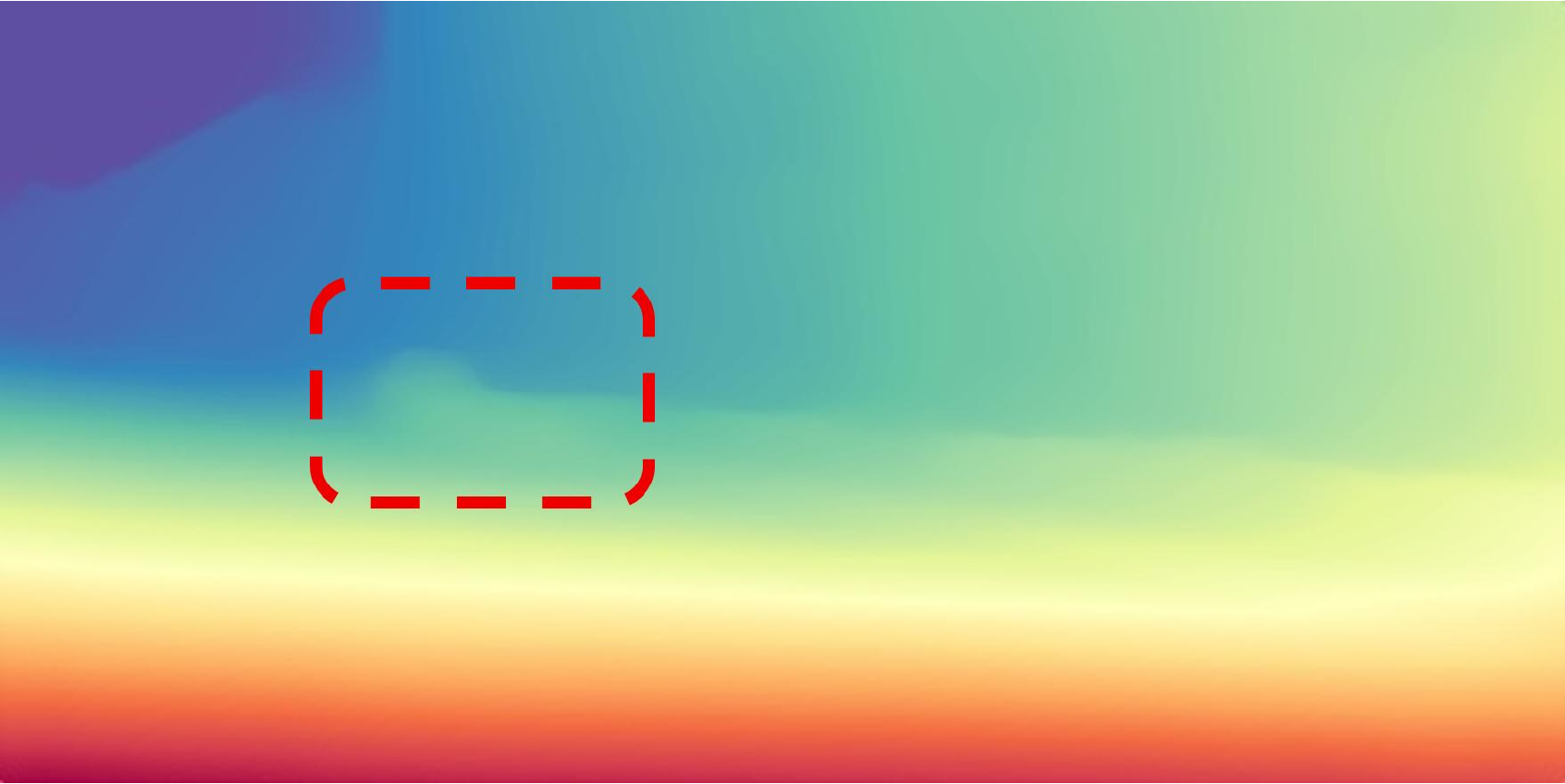}
    \\

    \rotatebox{90}{\hspace{0mm}\fontsize{5pt}{5pt}\selectfont{Depth Anything V2}
    }&
   \includegraphics[width=\turnheightnew]{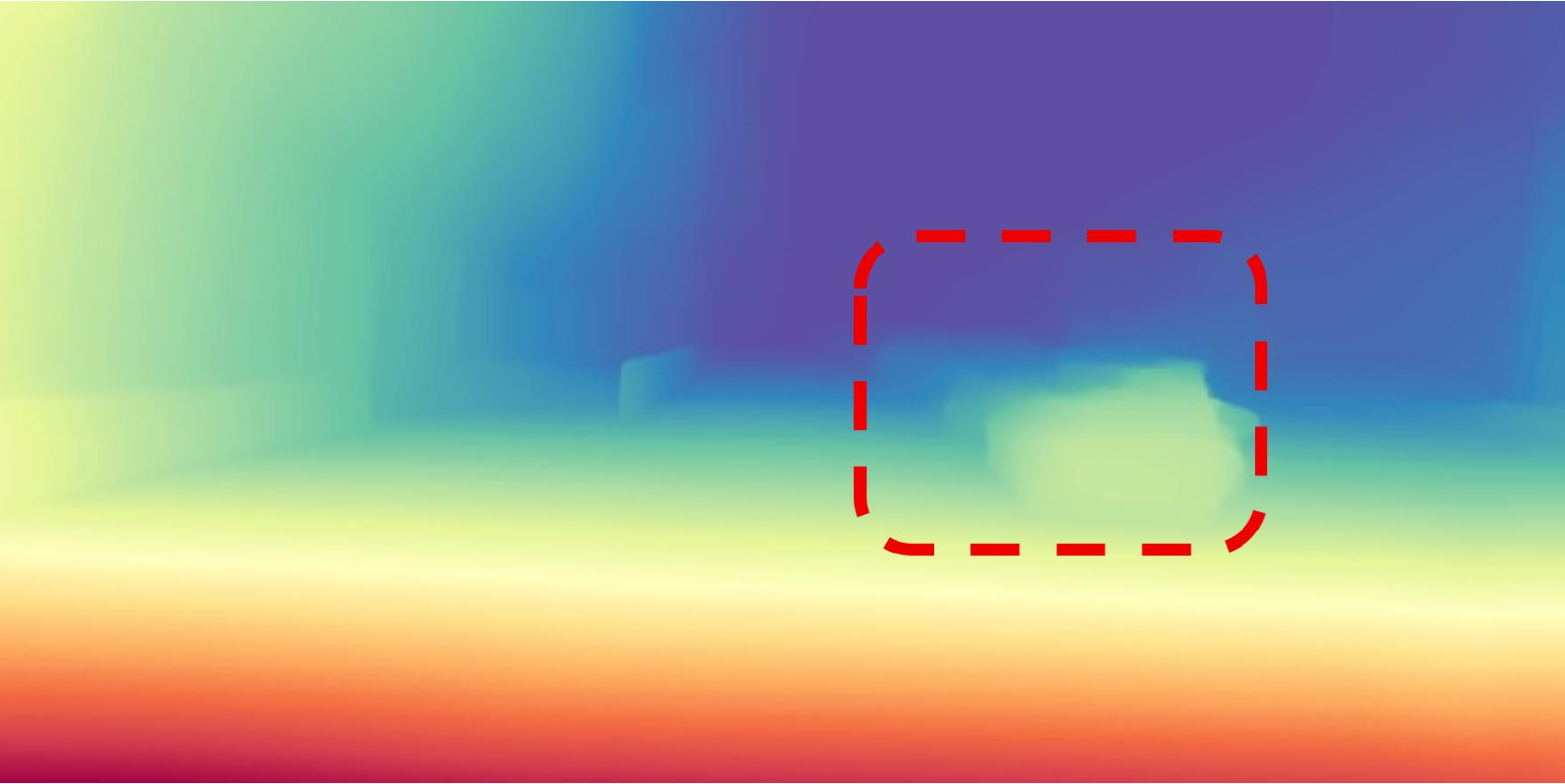} &
    \includegraphics[width=\turnheightnew]{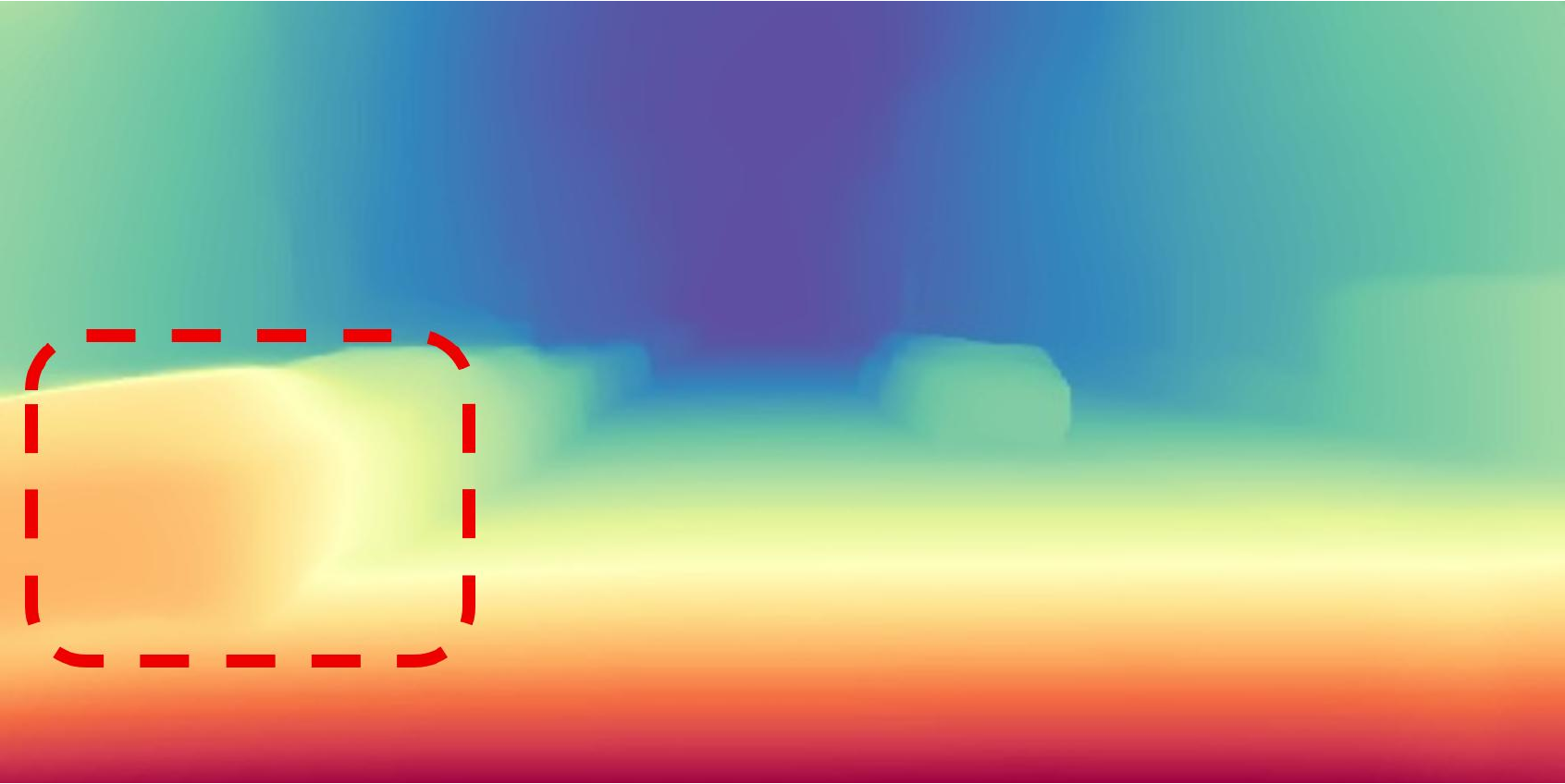} &
    \includegraphics[width=\turnheightnew]{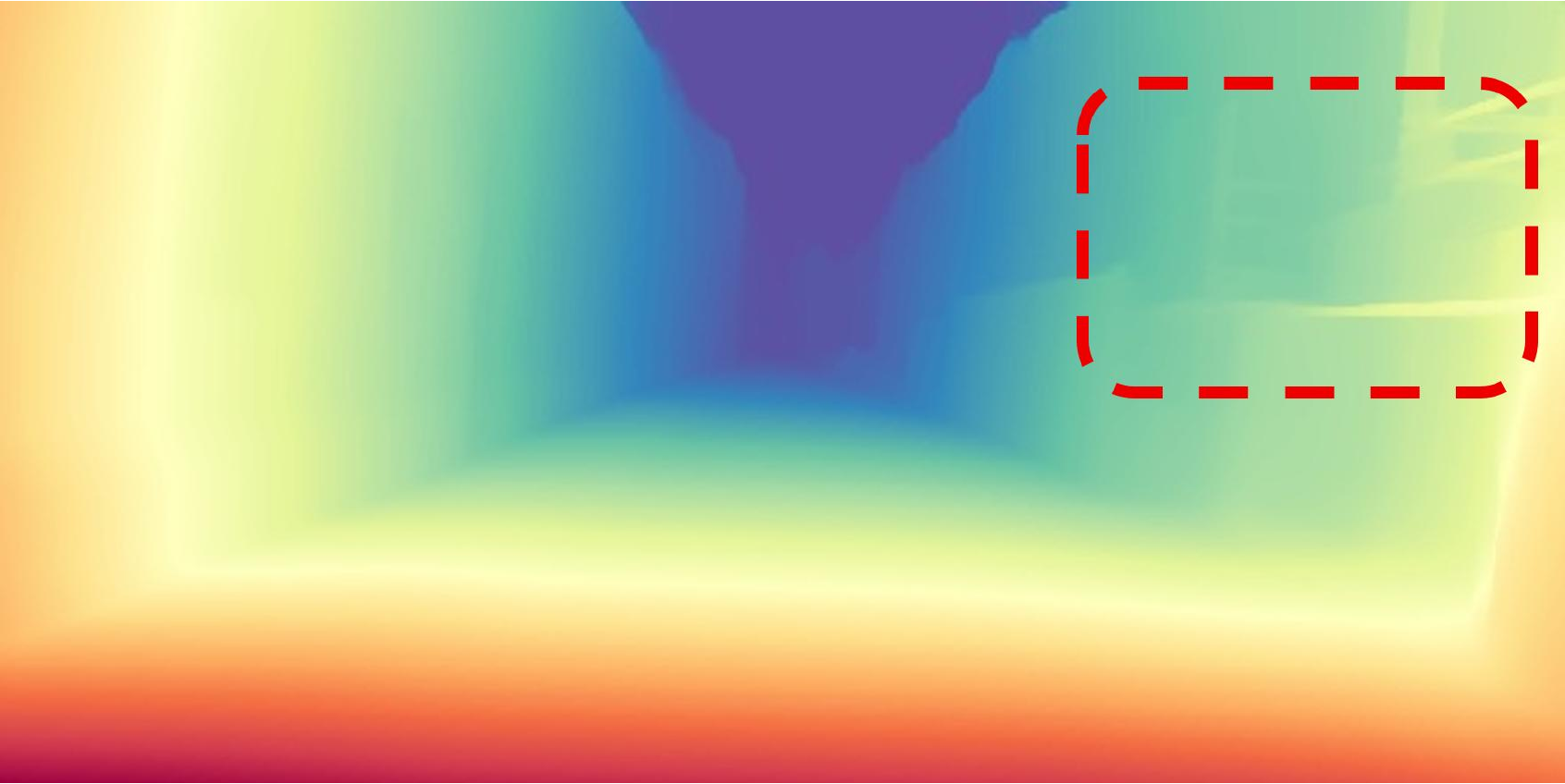} &

     \includegraphics[width=\turnheightnew]{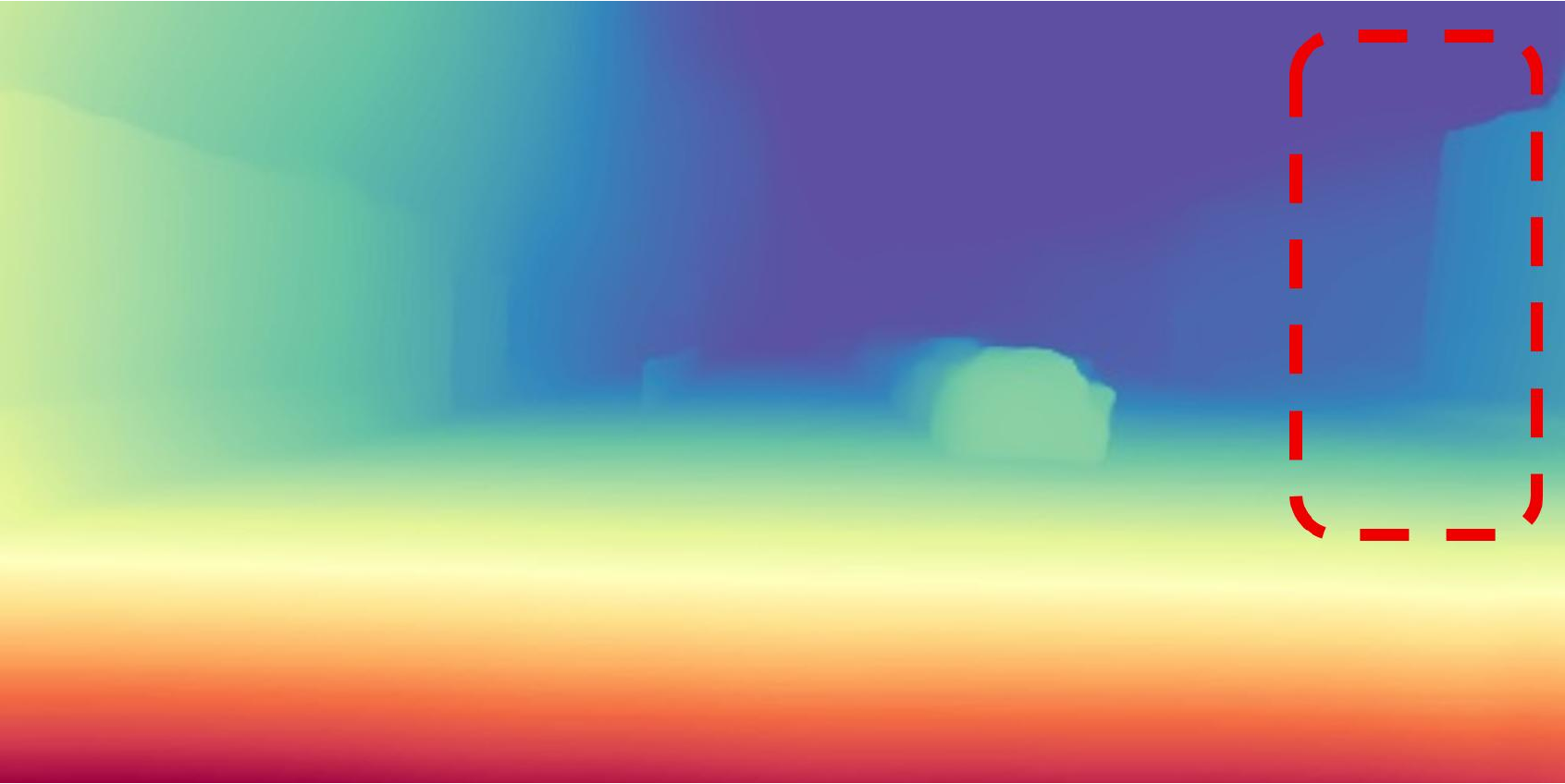} &
    \includegraphics[width=\turnheightnew]{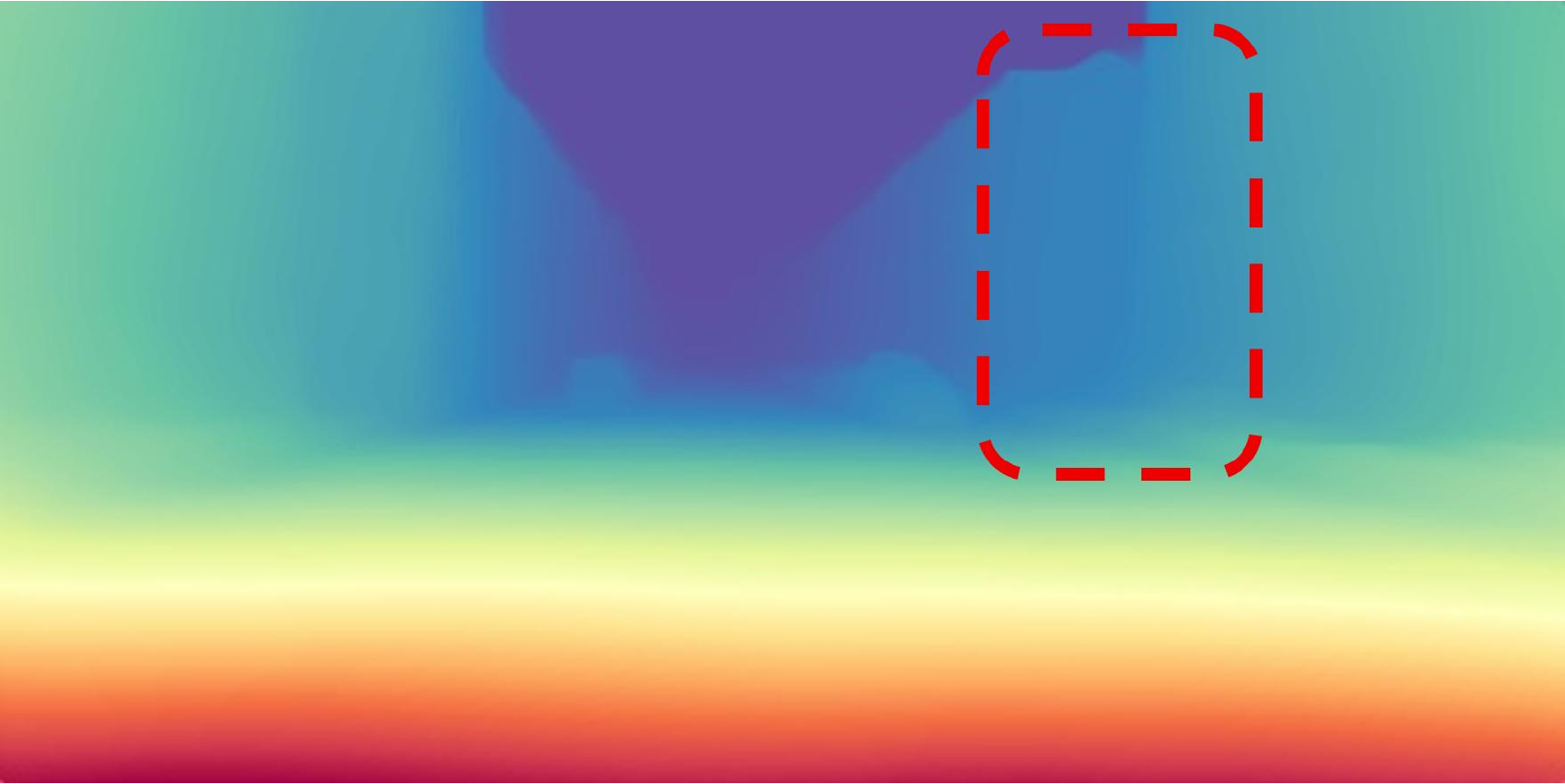} &
    \includegraphics[width=\turnheightnew]{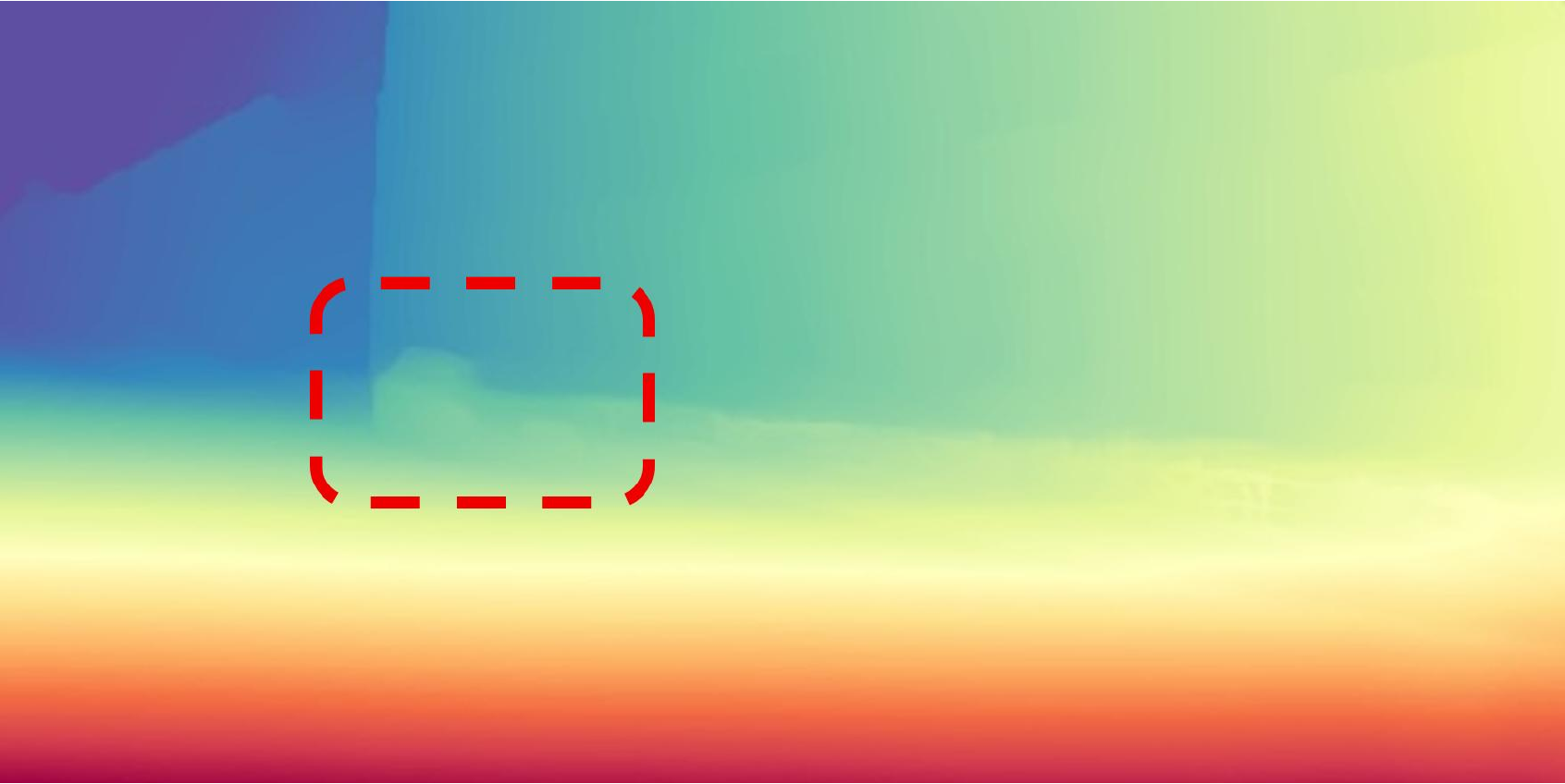}
    \\

    \rotatebox{90}{\hspace{1mm}\fontsize{5pt}{5pt}\selectfont{Ours DepthDark}}&

    \includegraphics[width=\turnheightnew]{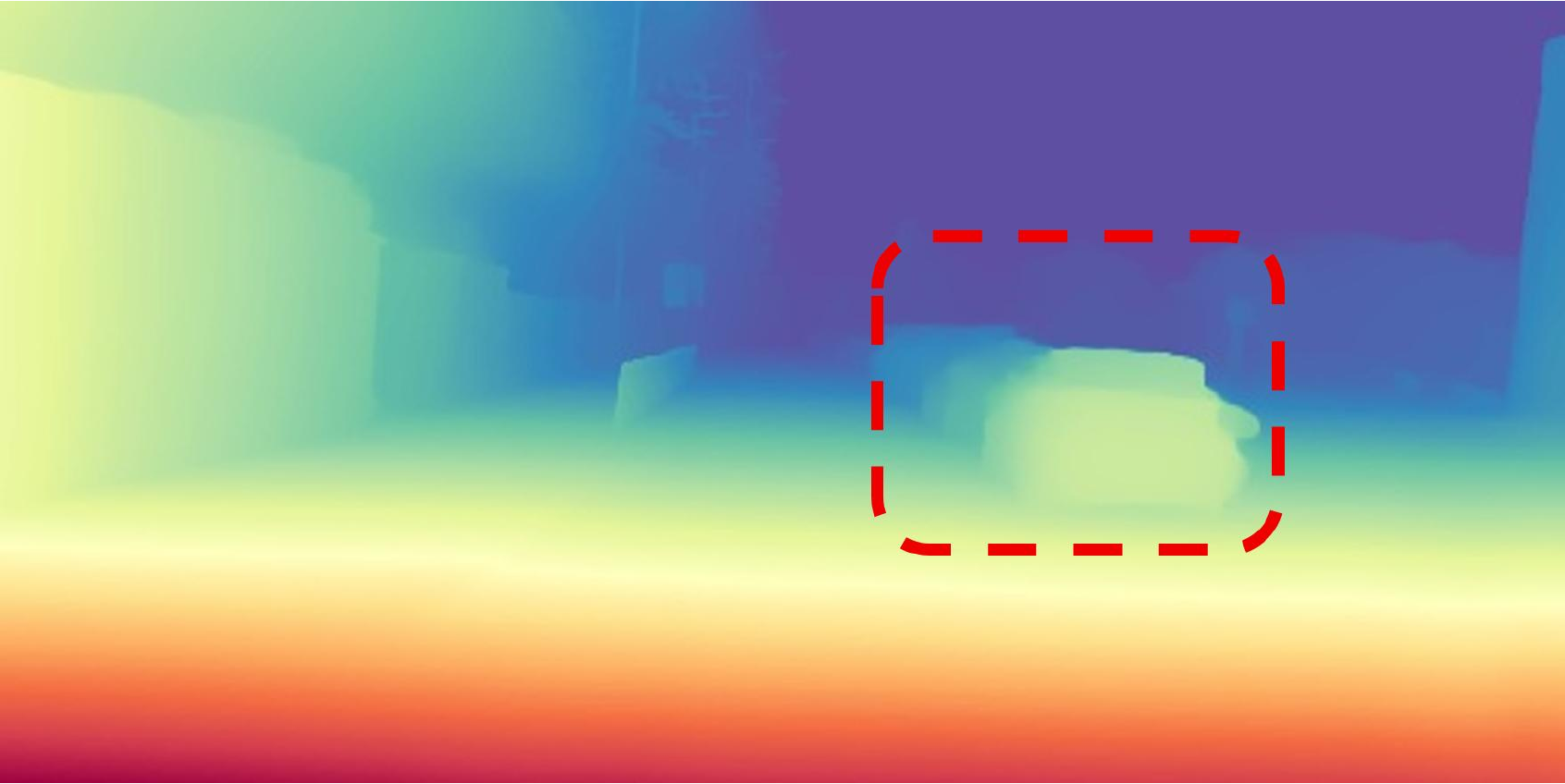} &
    \includegraphics[width=\turnheightnew]{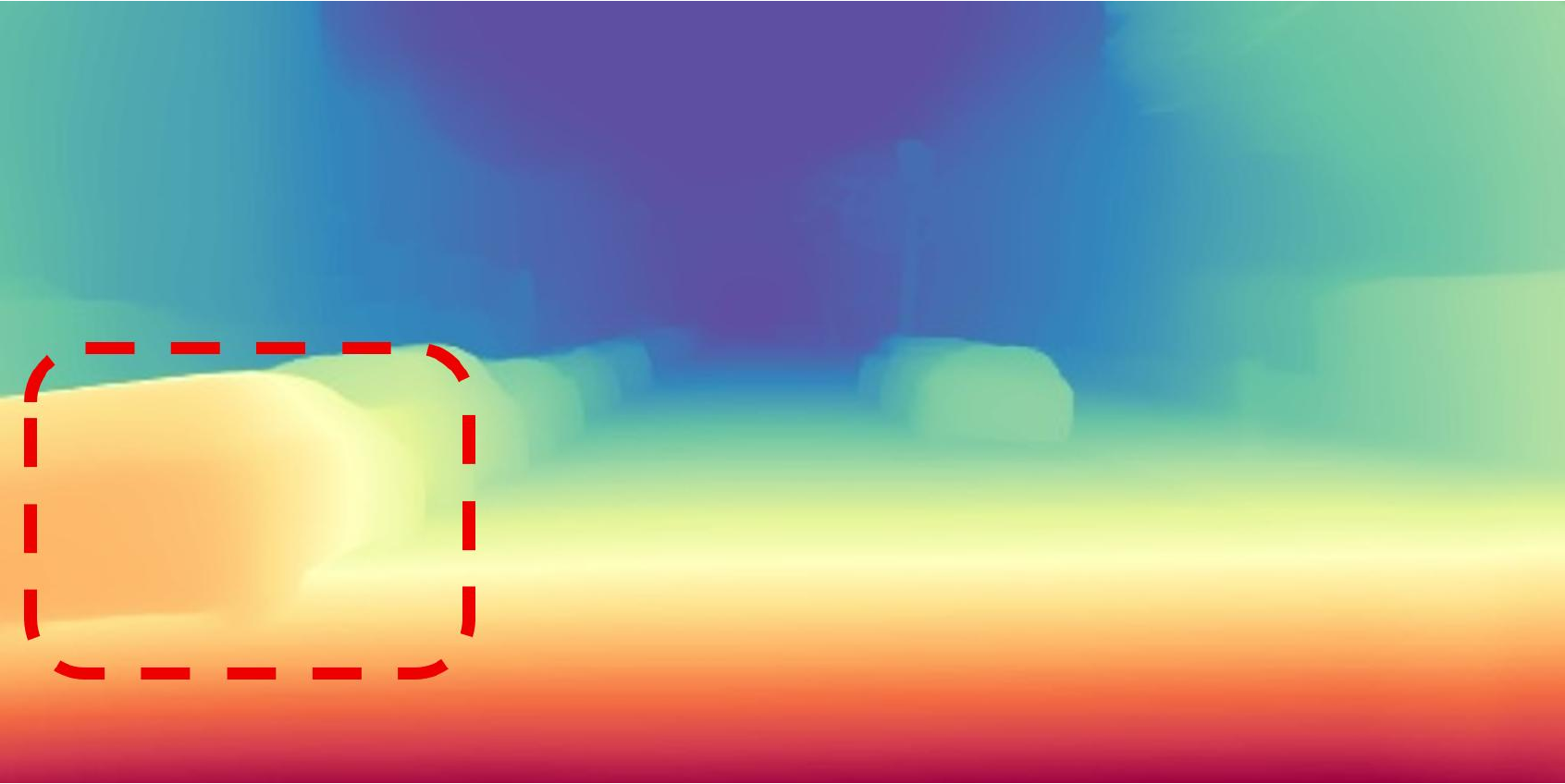} &
    \includegraphics[width=\turnheightnew]{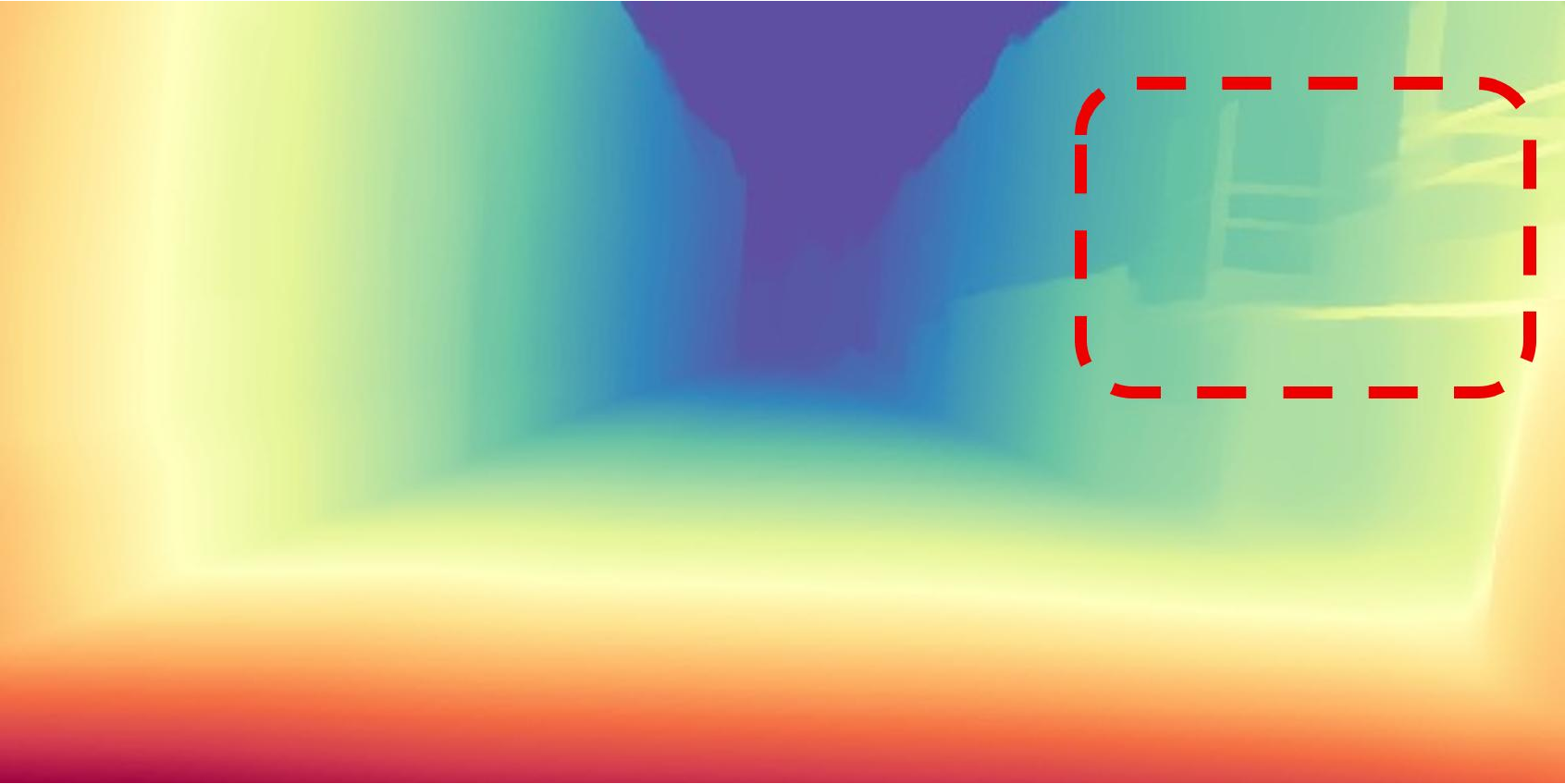} &
    
    \includegraphics[width=\turnheightnew]{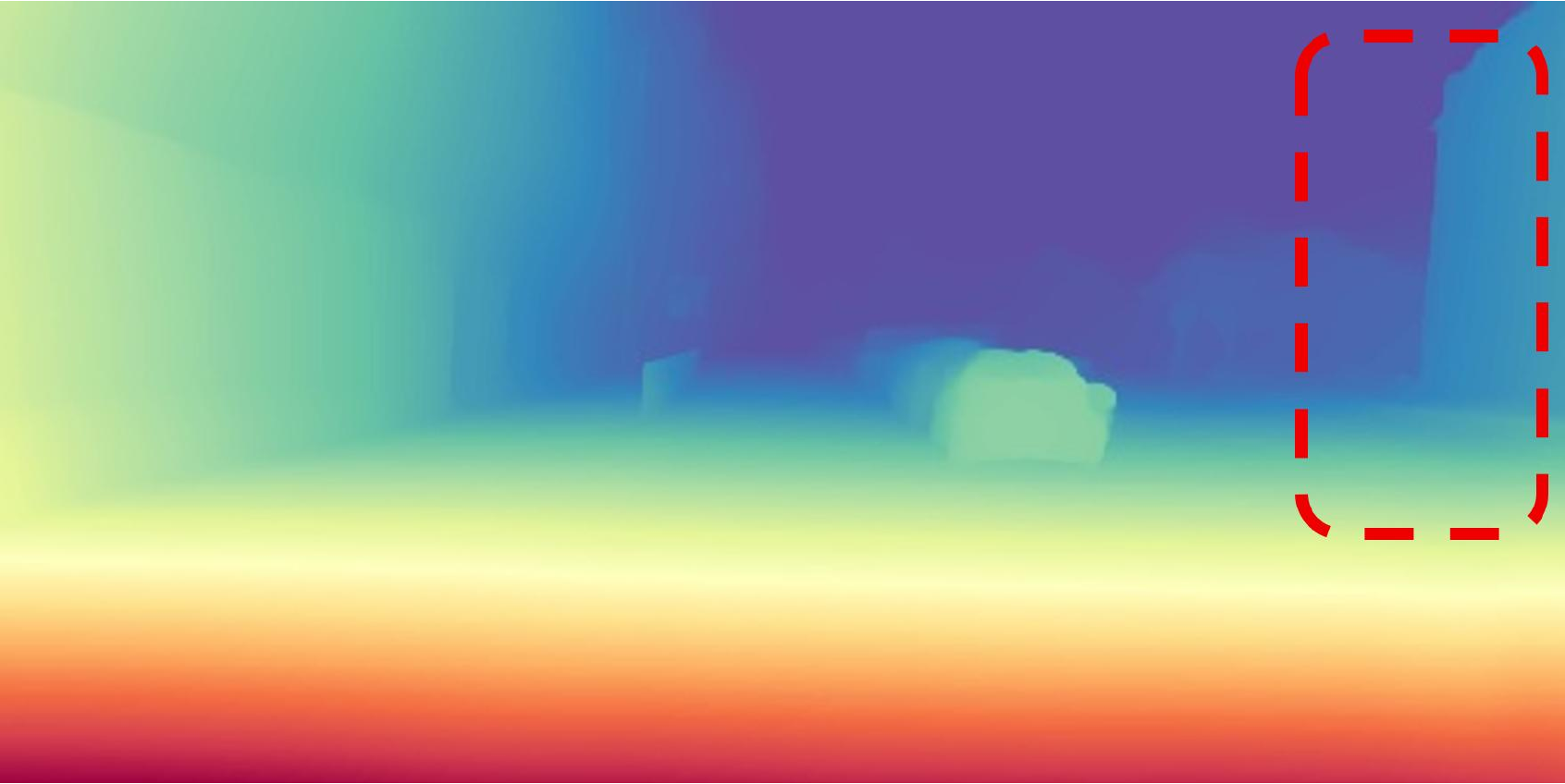} &
    
    \includegraphics[width=\turnheightnew]{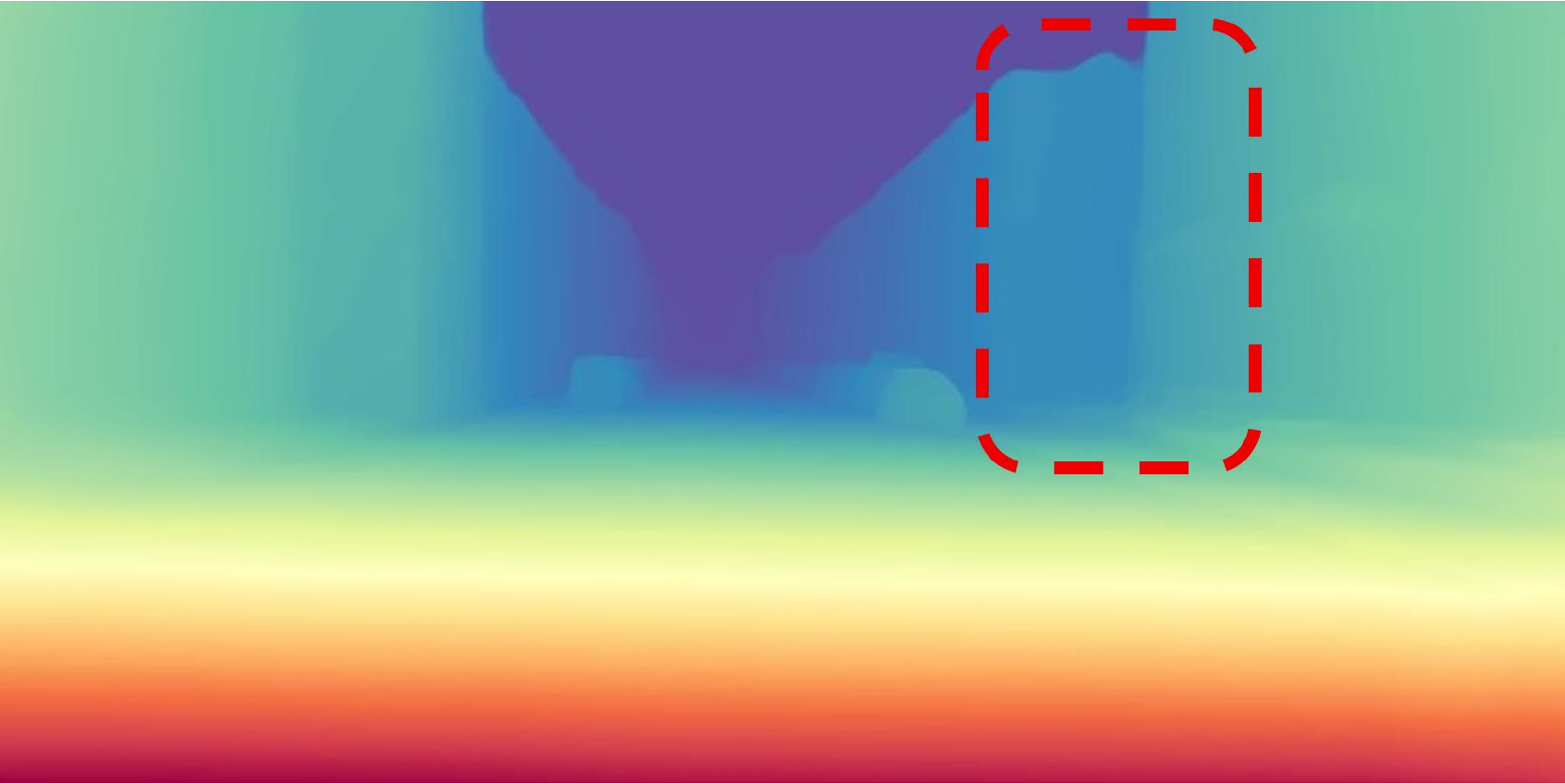} &
    \includegraphics[width=\turnheightnew]{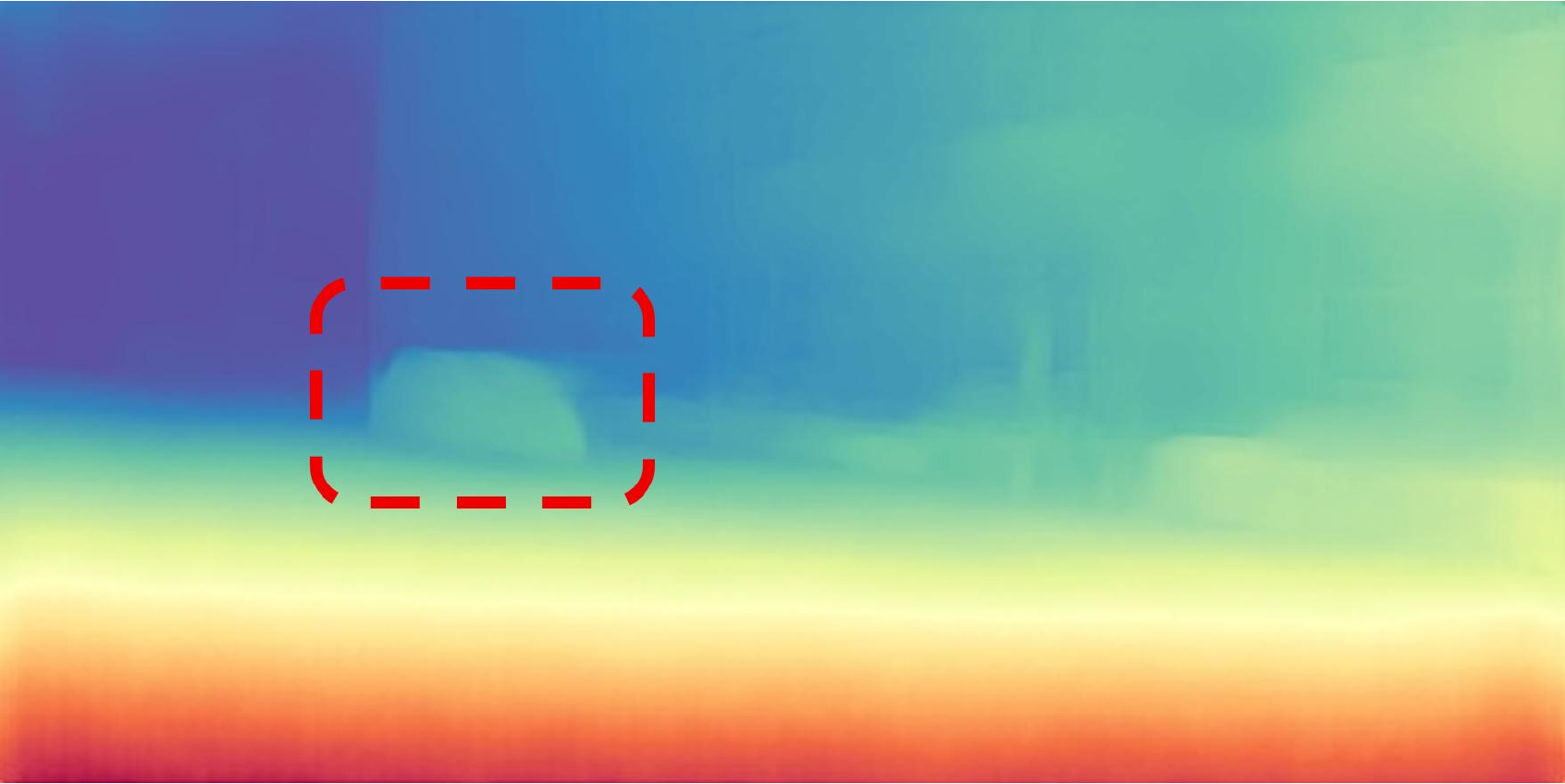} 
    \\
    
    \end{tabular}
    \vspace{-1mm}
    \caption{Qualitative comparison results of different monocular depth estimation methods on the RobotCar-Night dataset. We systematically evaluate performance across scenes with varying noise levels, where red dashed boxes clearly demarcate regions where our method demonstrates substantial performance advantages.
    \vspace{-2mm}
    }
    \label{fig:visual_ro}
    \vspace{-2mm}
\end{figure*}

Since we observed that the ground truth data in the KITTI training dataset is unreliable in certain scenarios and suffers from limited data volume and weak scene generalization, it is not ideal for training foundation models. Therefore, we selected the Hypersim \cite{Hypersim} indoor dataset and the Virtual KITTI \cite{kitti_2} synthetic outdoor dataset as our training data. As these datasets contain only daytime images, we employed the techniques described in Section \ref{sec:method_noise} to synthesize a high-quality, large-scale low-light depth-aligned dataset for training our proposed DepthDark model.

\subsubsection{Evaluation Protocol}

To evaluate our method, we selected two challenging benchmark datasets: nuScenes-Night and RobotCar-Night, which are, to the best of our knowledge, the most widely used benchmarks for low-light monocular depth estimation tasks. The evaluation settings for nuScenes-Night follow RNW \cite{RNW}, while those for RobotCar-Night follow ADDS \cite{adds}. Moreover, performing low-light image enhancement before applying these depth estimation models not only increases computational cost but also reduces the generalization capability of the models. Therefore, we primarily compare our experimental results with five methods designed for low-light monocular depth estimation. Specifically, direct training (DT) methods include MonoViT \cite{Monovit} and WSGD \cite{WSGD}, whereas domain adaptation (DA) methods consist of RNW, ADDS, and ITDFA \cite{ITDFA}. The evaluation settings are based on the protocol of TDDC \cite{TDDC}. Since the official code for TDDC is not publicly available, we use the experimental results reported in their paper. Additionally, we attempt to reproduce TDDC's approach and provide the general (G) version results for each DA method to facilitate further comparison. Our final experiments indicate that the G-version results are largely consistent with those reported in TDDC.  

For a more comprehensive comparison, we further compare our experimental results with state-of-the-art methods, including Depth Anything, Depth Anything V2, and TDDC. Evaluations on both nuScenes-Night and RobotCar-Night datasets demonstrate that our method exhibits superior robustness in low-light scenarios.

\subsection{Comparison with other methods.} 
\subsubsection{Quantitative Results.}        

We performed a quantitative evaluation of DepthDark and compared it with various state-of-the-art methods. As shown in Tab. \ref{Table:all_result}, the overall performance demonstrates that results on the nuScenes-Night dataset are generally worse compared to RobotCar-Night. This is primarily due to the strong corrective effects of the ISP in RobotCar-Night, which results in a cleaner image plane compared to nuScenes-Night. 

As demonstrated in Tab. \ref{Table:all_result}, although our training data contained no samples from the autonomous driving domain, our method achieved state-of-the-art performance on the nuScenes-Night dataset (with only the $\delta_{1}$ metric ranking second) and attained comprehensive leading results across all metrics on RobotCar-Night. This remarkable advantage stems from our innovative fine-tuning strategy, while preserving the core parameters pre-trained on large-scale synthetic datasets in Depth Anything V2, we specifically optimized the model for low-light conditions.

From the detailed results, our method outperforms all other approaches across every evaluation metric, even when tested on completely unseen datasets, which is particularly significant for depth estimation under low-light conditions. Additionally, we present experimental results on the generalized versions (G) of other methods, which exhibit a significant decline in accuracy for unseen nighttime scenes, indicating their limited generalization capability. In contrast, our method successfully overcomes this limitation, demonstrating outstanding robustness and generalization ability. 

To comprehensively evaluate the effectiveness and efficiency of our training framework and ensure a fair comparison with domain adaptation methods, we provide quantitative results of various monocular depth estimation methods on the RobotCar-Day and nuScenes-Day datasets, along with their generalized versions in the \emph{supplementary material}.

\subsubsection{Qualitative results.} 

\begin{table}[!t]
    \centering
        \resizebox{0.48\textwidth}{!}{
        \begin{tabular}{ c | c c c c | ccc}
            \hline
            Method  & \cellcolor{cell1} ABS rel $\downarrow$ & 
            \cellcolor{cell1} Sq rel $\downarrow$  & \cellcolor{cell1} RMSE $\downarrow$ & \cellcolor{cell1} RMSE log $\downarrow$  & \cellcolor{cell2} $\delta_{1} \uparrow$ & \cellcolor{cell2} $\delta_{2} \uparrow$ & \cellcolor{cell2} $\delta_{3} \uparrow$\\
            \hline
            
            \multicolumn{8}{c}{\cellcolor{cell3} Test on nuScenes-Night} \\
            
                  Depth Anything V2   &   0.272  &   \underline{ 2.551 }  &   8.576  &   0.304  &   0.518  &   0.843  &   0.971  \\
                  
            + Only LLDG    & 0.264 &2.956    	& 9.209     	& 0.371     	& 0.581     	& 0.813     	& 0.914 \\
            
             + Only LLPEFT  & \underline{ 0.255 } &   3.050  &  \underline{ 7.686 }  &   \underline{ 0.302 } &   \underline{ 0.619 }   &   \underline{ 0.836 }  &   \underline{ 0.934 }  \\

        \textbf{DepthDark}   &   \textbf{0.210}  &   \textbf{1.910} &   \textbf{7.764}  &  \textbf{0.260}   &   \textbf{0.630}  &  \textbf{0.914}   &  \textbf{0.976}   \\

             \multicolumn{8}{c}{\cellcolor{cell3} Test on RobotCar-Night} \\
           
           Depth Anything V2  &   0.235  &   2.474  &   \underline{ 6.239}  &   0.268  &  0.697   &   0.868  &   0.946  \\ 
           
            + Only LLDG &   0.183  &   1.679  &   6.769  &   0.251  &   \underline{ 0.702 }  &   0.893  &   0.972  \\ 
    
             + Only LLPEFT &   \underline{ 0.177 }  &   \underline{ 1.584 }  &   6.727  &   \underline{ 0.245 }  &   0.700  &   \underline{ 0.895 } &  \underline{ 0.973 }   \\

             \textbf{DepthDark}  &    \textbf{0.157}  &  \textbf{1.063}   &   \textbf{4.284}   & \textbf{0.202}    &  \textbf{0.760}  &  \textbf{0.941}   &  \textbf{0.985}   \\
            
            \hline
        \end{tabular}
        }

        \caption{Efficacy of Proposed Modules: Ablation studies on nuScenes-Night and RobotCar-Night demonstrate that each module enhances DepthDark's performance, with combined use significantly improving robustness.
        \vspace{-2mm}
        }
    \label{Talble:ablation_each}
    \vspace{-3mm}  
\end{table}

To further validate the effectiveness of DepthDark, Figs. \ref{fig:visual_nu} and \ref{fig:visual_ro} presents qualitative results on the nuScenes-Night and RobotCar-Night low-light datasets, comparing our method with other monocular depth estimation approaches.  

As shown in the first row of Figs. \ref{fig:visual_nu} and \ref{fig:visual_ro}, low-light conditions exhibit significant image quality variations due to uneven illumination and noise artifacts. These lead to bright spots and noise patches, severely impairing object contours and overall image fidelity, thus posing challenges for monocular depth estimation. However, such phenomena are common in real-world nighttime environments and are critical for practical applications. Therefore, we selected diverse low-light scenarios for thorough testing and analysis. To ensure an objective evaluation of DepthDark, we carefully chose comparison methods. Due to the unavailability of TDDC’s official code, we selected ADDS as it is among the most advanced and representative low-light depth estimation methods.  

As illustrated in Figs. \ref{fig:visual_nu} and \ref{fig:visual_ro}, ADDS produces depth maps that significantly deviate from the ground truth. While Depth Anything and Depth Anything V2 can roughly estimate object contours and depth information, their predictions still exhibit noticeable discrepancies, particularly in the contours of vehicles and buildings. This is because ADDS is specifically designed for night scenes with strong artificial lighting, whereas Depth Anything and Depth Anything V2 were trained on large-scale daytime image datasets.  

In contrast, our approach demonstrates superior performance across diverse low-light conditions, generating more accurate and clearer depth maps. Even in the presence of strong noise and significant photometric distortions, our method remains robust, highlighting the necessity of efficiently fine-tuning Depth Anything V2 for low-light scenarios.

\begin{table}[!t]
    \centering
        \resizebox{0.48\textwidth}{!}{
        \begin{tabular}{ c | c |c c c c | ccc}
            \hline
            Method  & Parameter  & \cellcolor{cell1} ABS rel $\downarrow$ & 
            \cellcolor{cell1} Sq rel $\downarrow$  & \cellcolor{cell1} RMSE $\downarrow$ & \cellcolor{cell1} RMSE log $\downarrow$  & \cellcolor{cell2} $\delta_{1} \uparrow$ & \cellcolor{cell2} $\delta_{2} \uparrow$ & \cellcolor{cell2} $\delta_{3} \uparrow$\\
            \hline
            
            \multicolumn{9}{c}{\cellcolor{cell3} Test on nuScenes-Night} \\
            
            Depth Anything V2 & 97.470M  & 0.264 &2.956    	& 9.209     	& 0.371     	& 0.581     	& 0.813     	& 0.914 \\

            + AMFG  & 99.229M&   0.243  &   2.351  &   5.416  &   0.267  &   0.674  &   0.862  &   0.945  \\
           + LoRA  & 99.830M 	&   \underline{ 0.233 }  &  \underline{ 2.112 }  &   \textbf{5.220}  &   \underline{0.259}  &   \textbf{0.685} &   \underline{0.865}  &  \underline{0.947}   \\
             
            \textbf{+ LLPEFT } & 97.479M   &   \textbf{0.210}  &   \textbf{1.910} &   \underline{7.764}  &  \textbf{0.260}   &   \underline{0.630}  &  \textbf{0.914}   &  \textbf{0.976}   \\
        
            \hline
             \multicolumn{9}{c}{\cellcolor{cell3} Test on RobotCar-Night} \\
        
          Depth Anything V2 & 97.470M  &   0.183  &   1.679  &   6.769  &   0.251  &   0.702  &   0.893  &   0.972  \\ 

            + AMFG  & 99.229M  &   \underline{0.168}  &   1.481  &   6.470  &   0.234  &   0.717  &   0.908  &   0.976  \\          
             + LoRA  & 99.830M 	&   0.173  &   \underline{1.260}  &   \underline{4.527}  &   \underline{0.217}  &   \underline{0.738}  &  \underline{0.919}   &   \underline{0.981}  \\
             
            \textbf{+ LLPEFT } & 97.479M &    \textbf{0.157}   &  \textbf{1.063}   &   \textbf{4.284}   &  \textbf{0.202}   &  \textbf{0.760} &  \textbf{0.941}   &  \textbf{0.985}    \\
            \hline
        \end{tabular}
        }

        \caption{We conduct comprehensive ablation studies on PEFT methods for Depth Anything V2 across both nuScenes-Night and RobotCar-Night datasets, with quantitative analysis of parameter counts for each fine-tuned variant.
        \vspace{-2mm}
        }
    \label{Talble:ablation_LLPEFT}
    \vspace{-3mm}
\end{table}

\section{Ablation Studies}
\label{sec: ablation study}
\subsection{Ablation on Each Component}

As presented in Tab. \ref{Talble:ablation_each}, this study conducts systematic ablation experiments to evaluate the performance gains introduced by the LLDG and LLPEFT modules. The experimental results demonstrate that both modules significantly enhance the depth estimation performance in low-light conditions on both the nuScenes-Night and RobotCar-Night datasets, with particularly notable improvements on RobotCar-Night. Specifically, the large-scale paired low-light dataset generated by LLDG addresses the challenge of collecting training data for low-light foundation models, making the training of such models feasible. The LLPEFT module innovatively integrates illumination guidance and multiscale feature fusion, significantly enhancing the robustness of the foundation model in extreme low-light conditions.

\subsection{Ablation on Parameter Efficient Fine Tuning} 

To thoroughly evaluate the effectiveness of LLPEFT, we conduct systematic comparisons with representative PEFT methods. As presented in Tab. \ref{Talble:ablation_LLPEFT}, using the dataset generated by LLDG, we compare various PEFT approaches to fine-tuning Depth Anything V2, including the AMFG method adopted by RobustSam for low-light SAM fine-tuning \cite{chen2024robustsam} and the classical Low-Rank Adaptation (LoRA) module \cite{hu2021lora}. Experimental results on both RobotCar-Night and nuScenes-Night datasets demonstrate that LLPEFT achieves performance comparable to other advanced methods while introducing negligible parameter overhead. This advantage primarily stems from LLPEFT's novel illumination guidance and multiscale feature fusion mechanism, which jointly optimize illumination perception and feature extraction to effectively address critical challenges including noise interference and uneven illumination distribution in low-light conditions.

\section{Conclusion}

In this paper, we introduce DepthDark, a robust foundation model for low-light monocular depth estimation. We first propose flare-simulation module and noise-simulation module techniques to accurately simulate the imaging process under nighttime conditions, thereby generating high-quality paired depth datasets for low-light environments. Additionally, we present an effective PEFT strategy that integrates illumination guidance and multiscale feature fusion, enhancing the model's robustness and adaptability in low-light conditions. Fine-tuning Depth Anything V2, our method, DepthDark, achieves state-of-the-art performance on the nuScenes-Night and RobotCar-Night datasets, surpassing existing methods in qualitative results in low-light scenarios. Ultimately, our proposed LLDG and LLPEFT modules provide a novel and viable approach for fine-tuning large-scale foundation models in low-light monocular depth estimation.

\section{Acknowledgments}
\begin{sloppypar}
This work was supported by the National Key Research and Development Program of China (U22A2047), the Key R\&D Program of Zhejiang under Grant No. (2025C03001, 2023C01044), the Fundamental Research Funds for the Provincial Universities of Zhejiang (GK259909299001-023), the National Nature Science Foundation of China (62301198).
\end{sloppypar}


\bibliographystyle{ACM-Reference-Format}
\bibliography{acm25}

\end{document}